\documentclass{article}

\PassOptionsToPackage{numbers, compress}{natbib}

\usepackage[preprint]{neurips_2023}

\usepackage[utf8]{inputenc} 
\usepackage[T1]{fontenc}    
\usepackage{hyperref}       
\usepackage{url}            
\usepackage{booktabs}       
\usepackage{amsfonts}       
\usepackage{nicefrac}       
\usepackage{microtype}      
\usepackage{xcolor}         
\usepackage{bbm}
\usepackage{bm}
\usepackage[noend]{algpseudocode}

\usepackage{algorithmicx,algorithm,amsmath}
\usepackage{amsthm}

\usepackage{appendix}
\usepackage{threeparttable}
\usepackage{multirow}
\usepackage{utfsym}
\usepackage{graphicx}
\usepackage{subcaption}
\usepackage{epstopdf}
\usepackage{wrapfig}
\usepackage{amssymb}
\allowdisplaybreaks[4]

\usepackage{color, colortbl}
\definecolor{Gray}{gray}{0.9}

\usepackage{ulem} 

\usepackage{soul}
\soulregister{\cite}7 
\soulregister{\citep}7 
\soulregister{\citet}7 
\soulregister{\ref}7 
\soulregister{\pageref}7 

\title{DFRD: Data-Free Robustness Distillation for Heterogeneous Federated Learning}

\author{%
  Kangyang Luo$^{1}$, Shuai Wang$^{1}$, Yexuan Fu$^{1}$, Xiang Li$^{1}$\thanks{corresponding author}, Yunshi Lan$^{1}$, Ming Gao$^{1, 2}$ \\
  School of Data Science \& Engineering$^{1}$\\
  KLATASDS-MOE in School of Statistics$^{2}$\\
  East China Normal University\\
  Shanghai, China \\
  \texttt{\{52205901003, 51215903058,51215903042\}@stu.ecnu.edu.cn} \\
  \texttt{\{xiangli, yslan, mgao\}@dase.ecnu.edu.cn} \\
}

\begin{document}

\maketitle

\begin{abstract}
Federated Learning~(FL) is a privacy-constrained decentralized machine learning paradigm in which clients enable collaborative training without compromising private data.
However, how to learn a robust global model in the data-heterogeneous and model-heterogeneous FL scenarios is challenging.
To address it, we resort to data-free knowledge distillation to propose a new FL method~(namely DFRD).
DFRD equips a conditional generator on the server to approximate the training space of the local models uploaded by clients, and systematically investigates its training in terms of \textit{fidelity}, \textit{transferability} and \textit{diversity}.
To overcome the catastrophic forgetting of the global model caused by the distribution shifts of the generator across communication rounds, we maintain an exponential moving average copy of the generator on the server. 
Additionally, we propose dynamic weighting and label sampling to accurately extract knowledge from local models.
Finally, our extensive experiments on various image classification tasks illustrate that DFRD achieves significant performance gains compared to SOTA baselines.
Our code is here:
\href{https://anonymous.4open.science/r/DFRD-0C83/}{https://anonymous.4open.science/r/DFRD-0C83/}.
\end{abstract}

\section{Introduction}
With the surge of data, deep learning algorithms have made significant progress in both established and emerging fields~\cite{Cordts2016cityscapes, Krizhevsky2017Imagenet, Chen2022Towards}.
However, in many real-world applications (e.g., mobile devices~\cite{Lim2020Federated}, IoT~\cite{Nguyen2021Federated}, and autonomous driving~\cite{Li2021Privacy}, etc.), data is 
generally dispersed across different clients~(i.e., data silos).
Owing to the high cost of data collection and strict privacy protection regulations, the centralized training that integrates data together is prohibited~\cite{Voigt2017eu, Kairouz2021Advances}.
Driven by this reality, 
Federated Learning~(FL)~\cite{Konecny2015Federated, Konečný2016Federated} has gained considerable attention in industry and academia as a promising distributed learning paradigm that allows multiple clients to participate in the collaborative training of a global model without access to their private data, thereby ensuring the basic privacy. 

Despite its remarkable success, the inevitable hurdle that plagues FL research is the vast heterogeneity among real-world clients~\cite{Li2020Federated1}. 
Specifically, the distribution of data among clients may be non-IID (identical and independently distributed), resulting in \textbf{data heterogeneity}~\cite{Fallah2020Personalized, Zhao2018Federated, Li2022Federated,  Kairouz2021Advances, Li2021Fedbn}. 
It has been confirmed that the vanilla FL method FedAvg~\cite{McMahan2017Communication} suffers from \textit{client drift} in this case, which leads to severe performance degradation.
To ameliorate this issue,
a plethora of modifications~\cite{Li2020Federated, Karimireddy2020Scaffold, Acar2021Federated, Kim2022Multi, Mendieta2022Local, Lee2022Preservation, Zhang2022Federated, Luo2023GradMA, Li2021Model} for FedAvg focus on regularizing the objectives of the local models to align the global optimization objective. 
All of the above methods follow the widely accepted assumption of model homogeneity, where local models have to share the same architecture as the global model. 
Nevertheless, when deploying FL systems, different clients may have distinct hardware and computing resources, and can only train the model architecture matching their own resource budgets~\cite{Ignatov2018Ai, Hong2022Efficient}, resulting in \textbf{model heterogeneity}.
In this case, to enable the FL systems with model homogeneity, on the one hand, clients with low resource budgets, which may be critical for enhancing the FL systems, will be discarded at the expense of training bias~\cite{Pan2010A, Kairouz2021Advances, Makhija2022Architecture}.
On the other hand, keeping a low complexity for the global model to accommodate all clients leads to performance drop due to the limited model capacity~\cite{Caldas2018Expanding}.
Therefore, the primary challenge of model-heterogeneous FL is how to conduct model aggregation of heterogeneous architectures among clients to enhance the inclusiveness of federated systems.
To solve this challenge, existing efforts fall broadly into two categories: knowledge distillation~(KD)-based methods~\cite{Lin2020Ensemble, He2020Group, Li2019Fedmd, Afonin2021Towards, Cho2022Heterogeneous, Fang2022Robust} and partial training~(PT)-based methods~\cite{Caldas2018Expanding, Horvath2021Fjord, Diao2020HeteroFL, Alam2022FedRolex}, yet each of them has its own limitations.
Concretely, KD-based methods require additional public data to align the logits outputs between
the global model~(student) and local models~(teachers). 
But the desired public data is not always available in practice and the performance may decrease dramatically if the apparent disparity in distributions exists between public data and clients' private data~\cite{Tan2022Fedproto}.
PT-based methods send width-based sub-models to clients, which are extracted by the server from a larger global model according to each client's resource budget, and then aggregate these trained sub-models to update the global model.
PT-based methods can be considered as an extension of FedAvg to model-heterogeneous scenarios, which means they are implementation-friendly and computationally efficient, but they also suffer from the same adverse effects from data heterogeneity as FedAvg or even more severely.
In a word, how to learn a robust global model in FL with both data and model heterogeneity is a highly meaningful and urgent problem. 

\begin{figure}[htbp]
  \centering
  \includegraphics[scale=0.6]{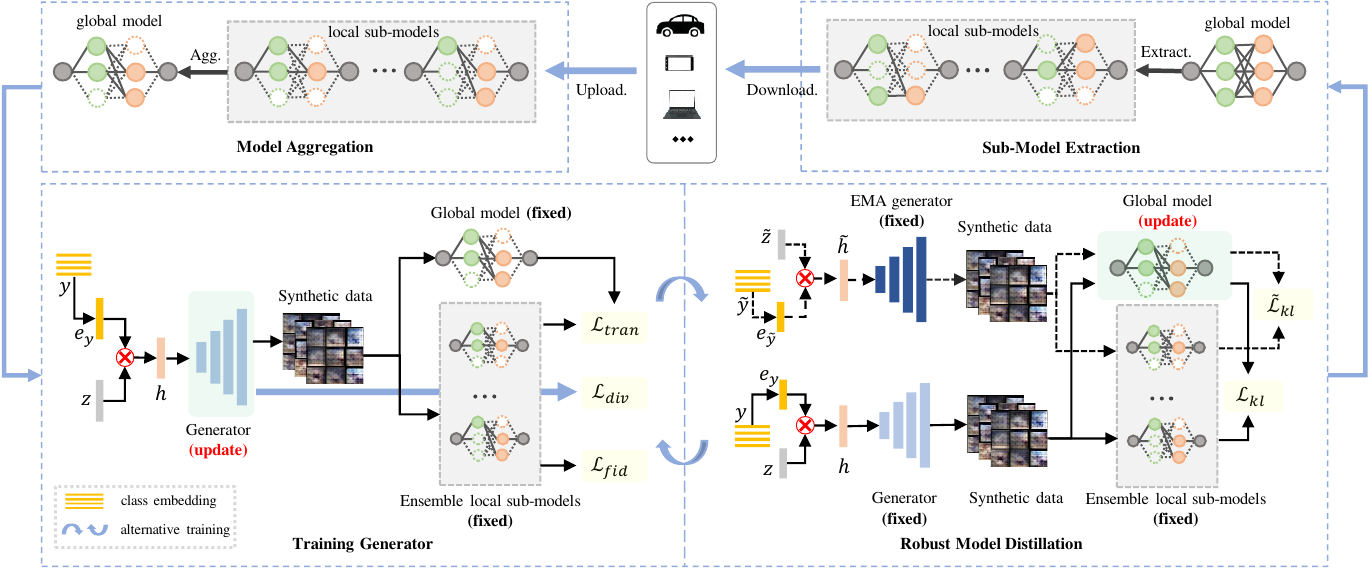}
  \caption{The full pipeline for DFRD combined with a PT-based method. DFRD works on the server and contains two phases, {\it training generator} and {\it robust model distillation}, where $\mathcal{L}_{tran},\mathcal{L}_{div},\mathcal{L}_{fid}$ and $\mathcal{L}_{kl},\widetilde{\mathcal{L}}_{kl}$ are the loss objectives of the conditional generator and the global model, respectively.}
  \label{Framework:}
\end{figure}

To this end, we systematically investigate the training of a robust global model in FL with both data and model heterogeneity with the aid of data-free knowledge distillation~(DFKD)~\cite{Chen2019Data, Micaelli2019Zero, Do2022Momentum, Yoo2019Knowledge}.
See the related work in Appendix~\ref{Related_work:} for more DFKD methods.
Note that the strategy of integrating DFKD to FL is not unique to us.
Recently, FedFTG~\cite{Zhang2022Fine} leverages DFKD to fine-tune the global model in model-homogeneous FL to overcome data heterogeneity, and DENSE~\cite{Zhang2022DENSE} aggregates knowledge from heterogeneous local models based on DFKD to train a global model for one-shot FL.
They all equip the server, which possesses powerful hardware and computing resources, with a generator to approximate the training space of the local models~(teachers), and train the generator and the global model~(student) in an adversarial manner.
However, the local models uploaded per communication round are not only architecturally heterogeneous but also trained on non-IID distributed private data in the situation of both data and model heterogeneity.
{\it In this case, the generator tends to deviate from the real data distribution. Also, its output distribution may undergo large shifts~(i.e., distribution shifts) across communication rounds, causing the global model to catastrophically forget useful knowledge learned in previous rounds and suffer from performance degradation.}
To confront the mentioned issues, we propose a novel \underline{\textbf{D}}ata-\underline{\textbf{F}}ree \underline{\textbf{R}}obust \underline{\textbf{D}}istillation FL method called DFRD, which utilizes a conditional generator to generate synthetic data and thoroughly studies how to effectively and accurately simulate the local models' training space in terms of \textit{fidelity}, \textit{transferability} and \textit{diversity}~\cite{Zhang2022Fine, Zhang2022DENSE}.
To mitigate catastrophic forgetting of the global model, an exponential moving average~(EMA) copy of the conditional generator is maintained on the server to store previous knowledge learned from the local models. 
The EMA generator, along with the current generator, provides training data for updates of the global model.
Also, we propose dynamic weighting and label sampling to aggregate the logits outputs of the local models and sample labels respectively, thereby properly exploring the knowledge of the local models.
We revisit FedFTG and DENSE, and argue that DFRD as a fine-tuning method~(similar to FedFTG) can significantly enhance the global model. 
So, we readily associate the PT-based methods in model-heterogeneous FL with the ability to rapidly provide a preliminary global model, which will be fine-tuned by DFRD. 
We illustrate the schematic for DFRD as a fine-tuning method based on a PT-based method in Fig.~\ref{Framework:}. 
Although FedFTG and DENSE can also be applied to fine-tune the global model from the PT-based methods after simple extensions, we empirically find that they do not perform as well, and the performance of the global model is even inferior to that of local models tailored to clients' resource budgets.

Our main contributions of this work are summarized as follows. 
First, we propose a new FL method termed DFRD that enables a robust global model in both data and model heterogeneity settings with the help of DFKD.
Second, we systematically study the training of the conditional generator w.r.t. \textit{fidelity}, \textit{transferability} and \textit{diversity} to ensure the generation of high-quality synthetic data. 
Additionally, we maintain an EMA generator on the server to overcome the global model's catastrophic forgetting caused by the distribution shifts of the generator.
Third, we propose dynamic weighting and label sampling to accurately extract the knowledge of local models.
At last, our extensive experiments on six real-world image classification datasets verify the superiority of DFRD.

\section{Notations and Preliminaries}
\label{Notations_Problem_Setting:}

\textbf{Notations.} 
In this paper, we focus on the centralized setup that consists of a central server and $N$ clients owning private labeled datasets  $\{(\bm{X}_i, \bm{Y}_i)\}_{i=1}^N$, where $\bm{X}_i=\{\bm{x}_i^b\}_{b=1}^{n_i}$ follows the data distribution $\mathcal{D}_i $ over feature space $\mathcal{X}_i$, i.e., $\bm{x}_i^b \sim \mathcal{D}_i$, and $\bm{Y}_i=\{y_i^b\}_{b=1}^{n_i} \subset [C]:=\{1, \cdots, C\}$ denotes the ground-truth labels of $\bm{X}_i$. 
And $C$ refers to the total number of labels.
Remarkably, the heterogeneity for FL in our focus includes both data heterogeneity and model heterogeneity. 
\textbf{For the former}, we consider the same feature space, yet the data distribution may be different among clients, that is, label distribution skewness in clients~(i.e., $\mathcal{X}_i=\mathcal{X}_j$ and $\mathcal{D}_i \neq \mathcal{D}_j, \forall i\neq j, i,j \in [N]$). 
\textbf{For the latter}, each client $i$ holds an on-demand local model $f_i$ parameterized by $\bm{\theta}_i$.  
Due to the difference in resource budgets, the model capacity of each client may vary, i.e., $|\bm{\theta}_i| \neq |\bm{\theta}_j|, \exists i\neq j, i,j \in [N]$. 
In PT-based methods, we define a confined width capability $R_i \in (0, 1]$ according to the resource budget of client $i$, which is the proportion of nodes extracted from each layer in the global model $f$.
Note that $f$ is parameterized by $\bm{\theta}$, and $|\bm{a}|$ denotes the number of elements in vector $\bm{a}$.

\textbf{PT-based method} is a solution for model-heterogeneous FL, which strives to extract a matching width-based slimmed-down sub-model from the global model as a local model according to each client's budget.
As with FedAvg, it requires the server to periodically communicate with the clients.
In each round, there are two phases: local training and server aggregation. 
In local training, each client trains the sub-model received from the server utilizing the local optimizer. 
In server aggregation, the server collects the heterogeneous sub-models and aggregates them by straightforward selective averaging to update the global model, as follows~\cite{Horvath2021Fjord, Diao2020HeteroFL, Caldas2018Expanding, Alam2022FedRolex}:
\begin{equation}
    \label{partial_avg:}
     \bm{\theta}_{[l,k]}^{t}=\frac{1}{\sum_{j \in \mathcal{S}_t}p_j} \sum_{i \in \mathcal{S}_t} p_i \bm{\theta}_{i, [l,k]}^{t},
\end{equation}
where $\mathcal{S}_t$ is a subset sampled from $[N]$ and $p_i$ is the weight of client $i$, which generally indicates the size of data held by client $i$. At round $t$, $\bm{\theta}_{[l,k]}^{t}$ denotes the $k^{th}$ parameter of layer $l$ of the global model and $\bm{\theta}_{i,[l,k]}^{t}$ denotes the parameter $\bm{\theta}_{[l,k]}^{t}$ updated by client $i$. 
We can clearly see that Eq.~(\ref{partial_avg:}) independently calculates the average of each parameter for the global model according to how many clients update that parameter in round $t$.
Instead, the parameter remains unchanged if no clients update it.
Notably, if $|\bm{\theta}_{i}^{t}|=|\bm{\theta}^{t}|$ for any $i\in[N]$, PT-based method becomes FedAvg.
The key to PT-based method is to select $\bm{\theta}_{i}^{t}$ from the global model $\bm{\theta}^{t}$ when given $R_i$.
And existing sub-model extraction schemes fall into three categories: \textit{static}~\cite{Horvath2021Fjord, Diao2020HeteroFL}, \textit{random}~\cite{Caldas2018Expanding} and \textit{rolling}~\cite{Alam2022FedRolex}.


\section{Proposed Method} 
In this section, we detail the proposed method DFRD.
We mainly work on considering DFRD as a fine-tuning method to enhance the PT-based methods, thus enabling a robust global model in FL with both data and model heterogeneity.
Fig.~\ref{Framework:} visualizes the training procedure of DFRD combined with a PT-based method, consisting of four stages on the server side: \textit{training generator}, \textit{robust model distillation}, \textit{sub-model extraction} and \textit{model aggregation}.
Note that \textit{sub-model extraction} and \textit{model aggregation} are consistent with that in the PT-based methods, so we detail the other two stages. 
Moreover, we present pseudocode for DFRD in Appendix~\ref{sec:pseudo}.

\subsection{Training Generator}
\label{Train_Gen:}
At this stage, we aim to train a well-behaved generator to capture the training space of local models uploaded from active clients.
Specifically,
we consider a conditional generator $G(\cdot)$ parameterized by $\bm{w}$. 
It takes as input a random noise $\bm{z} \in \mathbbm{R}^d$ sampled from standard normal distribution $\mathcal{N} (\bm{0}, \bm{I})$,
and a random label $y \in [C]$ sampled from label distribution $p(y)$, i.e., the probability of sampling $y$,
thus generating the synthetic data $\bm{s}=G(\bm{h}=o(\bm{z}, y), \bm{w})$.
Note that $o(\bm{z}, y)$ represents the merge operator of $\bm{z}$ and $y$.
To the best of our knowledge, synthetic data generated by a well-trained generator should satisfy several key characteristics: \textit{fidelity}, \textit{transferability}, and \textit{diversity}~\cite{Zhang2022Fine, Zhang2022DENSE}. 
Therefore, in this section, we construct the loss objective from the referred
aspects to ensure the quality and utility of $G(\cdot)$.

\textbf{Fidelity.} 
To commence, we study the fidelity of the synthetic data. 
Specifically, we expect 
$G(\cdot)$
to simulate the training space of the local models to generate the synthetic dataset with a similar distribution to the original dataset. 
To put it differently, we want the synthetic data $\bm{s}$ to approximate the training data with label $y$ without access to  clients' training data. 
To achieve it, we form the fidelity loss $\mathcal{L}_{fid}$ at logits level:  
\begin{align}
    \label{L_fidelity:}
     \mathcal{L}_{fid}=CE(\sum_{i \in \mathcal{S}_t}\tau_{i, y} f_i(\bm{s}, \bm{\theta}_i), y),
\end{align} 
where $CE$ denotes the cross-entropy function, $f_i(\bm{s}, \bm{\theta}_i)$ is the logits output of the local model from client $i$ when $\bm{s}$ is given, $\tau_{i, y}$ dominates the weight of logits from different clients $\{i | i \in \mathcal{S}_t\}$ when $y$ is given. 
And $\mathcal{L}_{fid}$ is the cross-entropy loss between the weighted average logits $\sum_{i \in \mathcal{S}_t}\tau_{i, y} f_i(\bm{s}, \bm{\theta}_i)$ and the label $y$.
By minimizing $\mathcal{L}_{fid}$, $\bm{s}$ is enforced to be classified into label $y$ with a high probability, thus facilitating the fidelity of $\bm{s}$. 

\begin{wrapfigure}{r}{0cm}
  \vspace*{-6ex}
  \centering
  \includegraphics[scale=0.3]{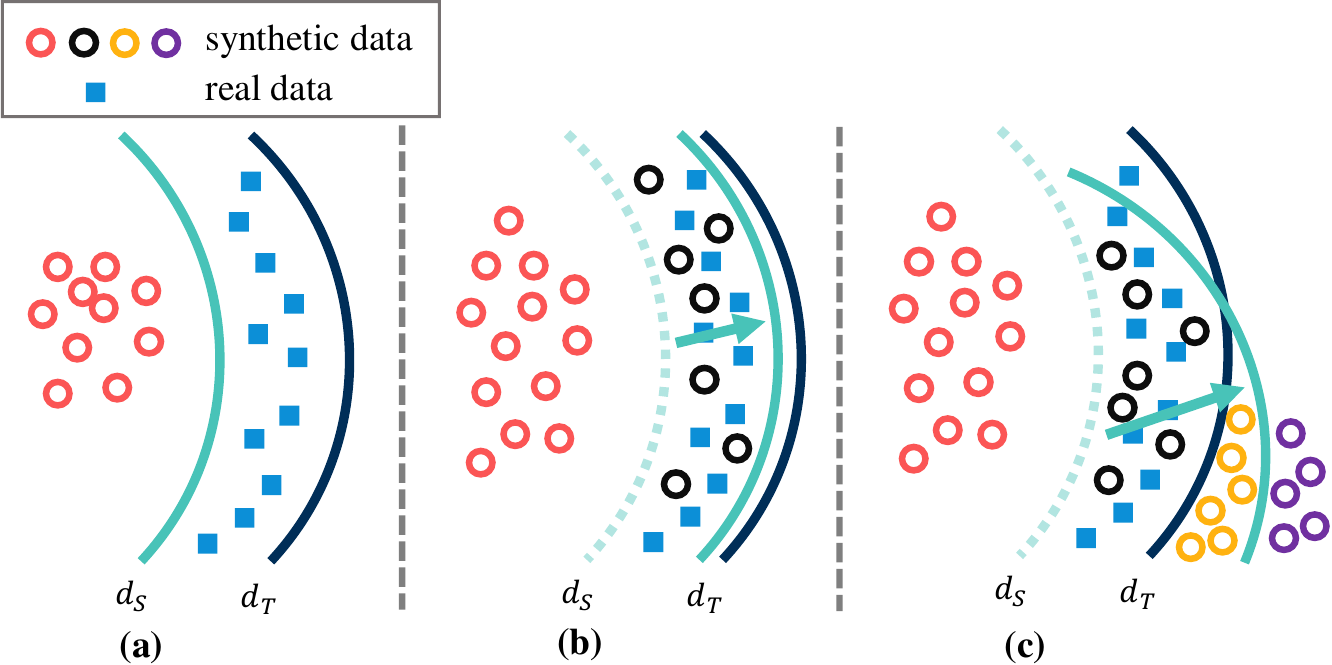}
  \caption{The visualization of synthetic data and decision boundaries of global model~(student) and ensemble model~(teacher). \textit{Left panel}: synthetic data~(red circles) are far away from the decision boundary $d_T$. \textit{Middle panel}: synthetic data~(black circles) near the decision boundaries $d_T$.
  \textit{Right panel}: synthetic data~(yellow and purple circles) cross over the decision boundary $d_T$.}
  \label{pic_Dis_dia:}
  \vspace*{-3ex}
\end{wrapfigure}

In reality, the conditional generator $G(\cdot)$ easily generates synthetic data with low classification errors~(i.e. $\mathcal{L}_{fid}$ close to $0$) as the training proceeds.
This may cause the synthetic data to fall into a space far from the decision boundary of the ensemble model~(i.e., $\sum_{i \in \mathcal{S}_t}\tau_{i, y} f_i(\cdot, \bm{\theta}_i)$) if only $\mathcal{L}_{fid}$ is optimized, as shown in the synthetic data represented by red circles in Fig.~\ref{pic_Dis_dia:} (a).
Note that $d_S$ and $d_T$ denote the decision boundaries of the global model~(student) and the ensemble model~(teacher), respectively.
An obvious observation is that the red circles are correctly classified on the same side of the two decision boundaries~(i.e., $d_S$ and $d_T$), making it difficult to transfer teacher's knowledge to student.
We next explore how to augment the transferability of the synthetic data to ameliorate this pitfall.

\textbf{Transferability} 
is intended to guide $G(\cdot)$ in generating
synthetic data that moves the decision boundary of the global model towards that of the ensemble model, such as synthetic data with black circles in Fig.~\ref{pic_Dis_dia:}~(b). 
However, during the training of $d_S$ to approach $d_T$, we find that $G(\cdot)$ can easily generate two other types of synthetic data, the yellow and purple circles in Fig.~\ref{pic_Dis_dia:}~(c).  Both of them are misclassified by the ensemble model~($d_T$), while the yellow circles are correctly classified and the purple circles are misclassified by the global model~($d_S$).
For the conditional generator $G(\cdot)$ that takes label information as one of the inputs, 
yellow and purple circles can mislead the generator, thereby leading to $d_S$ approximating $d_T$ with a large deviation, as shown in Fig.~\ref{pic_Dis_dia:}~(c).
Based on the above observation, we reckon that the synthetic data $\bm{s}=G(\bm{h}=o(\bm{z}, y), \bm{w})$  is useful if it is classified as $y$ by the ensemble model but classified not as $y$ by the global model.
To realize it, we maximize the logits discrepancy between the global model and the ensemble model on synthetic data with black circles by leveraging Kullback-Leibler divergence loss, which takes the form:
\begin{align}
     \label{L_tran:}
     \mathcal{L}_{tran}=- \varepsilon \cdot KL(\sum_{i \in \mathcal{S}_t}\tau_{i, y} f_i(\bm{s}, \bm{\theta}_i), f(\bm{s}, \bm{\theta})),
\end{align}
where $KL$ is Kullback-Leibler divergence function and $f(\bm{s}, \bm{\theta})$ denotes the logits output of the global model on $\bm{s}$ with label $y$. Note that
$\varepsilon = 1$ if $\arg \max f(\bm{s}, \bm{\theta}) \neq y$ and $\arg \max \sum_{i \in \mathcal{S}_t}\tau_{i, y} f_i(\bm{s}, \bm{\theta}_i) = y$ hold, otherwise $\varepsilon = 0$.~($\Diamond$)

We would like to point out that the existing works~\cite{Zhang2022Fine} and~\cite{Zhang2022DENSE} are in line with our research perspective on the transferability of generator, which aims to generate more synthetic data with black circles.
However, they do not carefully frame learning objective for enhancing the transferability of generator. 
Concretely, \cite{Zhang2022Fine} does not consider the type of synthetic data, i.e., $\varepsilon = 1$ always holds, thus inducing the generation of synthetic data with yellow and purple circles.~($\bigtriangleup$)
\cite{Zhang2022DENSE} focuses on synthetic data satisfying $\arg \max f(\bm{s}, \bm{\theta}) \neq \arg \max \sum_{i \in \mathcal{S}_t}\tau_{i, y} f_i(\bm{s}, \bm{\theta}_i)$, but enables the generation of synthetic data with purple circles yet.~($\bigtriangledown$) 
\footnote{
Note that $\bigtriangleup$, $\bigtriangledown$ and $\Diamond$ denote the strategies for the transferability of generator  in~\cite{Zhang2022Fine}, in~\cite{Zhang2022DENSE} and in this paper, respectively.
}

\textbf{Diversity.} 
Although we enable $G(\cdot)$ to generate synthetic data that falls around the real data by optimizing $\mathcal{L}_{fid}$ and $\mathcal{L}_{tran}$, the diversity of synthetic data is insufficient.
Due to the fact that the generator may get stuck in \textit{local equilibria} as the training proceeds, model collapse occurs~\cite{Odena2017Conditional, Kodali2017On}.
In this case, the generator may produce similar data points for each class with little diversity. 
Also, the synthetic data points may not differ significantly among classes.
This causes the empirical distribution estimated by $G(\cdot)$ to cover only a small manifold in the real data space, and thus only partial knowledge of the ensemble model is extracted.
To alleviate this issue, we introduce a diversity loss $\mathcal{L}_{div}$ with label information to increase the diversity of synthetic data as follows:
\begin{align}
     \label{L_div:}
     \mathcal{L}_{div}= \exp{(-\sum\limits_{j,k\in [B]}\|\bm{s}_j-\bm{s}_k\|_2*\|\bm{h}_j-\bm{h}_k\|_2/B^2)},
\end{align}
where $B$ denotes the batch size and $\bm{s}_{j/k}=G(\bm{h}_{j/k}=o(\bm{z}_{j/k}, y_{j/k}), \bm{w})$. 
Intuitively, $\mathcal{L}_{div}$ takes  
$\|\bm{h}_j-\bm{h}_k\|_2$ as a weight, and then multiplies it by the corresponding 
$\|\bm{s}_j-\bm{s}_k\|_2$ in each batch $B$, 
thus imposing a larger weight on the synthetic data points pair~($\bm{s}_j$ and $\bm{s}_k$) at the more distant input pair~($\bm{h}_j$ and $\bm{h}_k$).
Notably, we merge the random noise $\bm{z}$ with label $y$ as the input of $G(\cdot)$ to overcome spurious solutions~\cite{Do2022Momentum}.
Further, we propose a multiplicative merge operator, i.e., $o(\bm{z}, y)= \bm{z} \times \mathcal{E}(y)$, where $\mathcal{E}$ is a trainable embedding and $\times$ means vector element-wise product. 
We find that our merge operator enables synthetic data with more diversity compared to others, possibly because the label information is effectively absorbed into the stochasticity of $\bm{z}$ by multiplying them when updating $\mathcal{E}$.
See Section \ref{Ablation_study:} for more details and empirical justification.

Combining $\mathcal{L}_{fid}$, $\mathcal{L}_{tran}$ and $\mathcal{L}_{div}$, the overall objective of the generator can be formalized as follows:
\begin{align}
     \label{L_div:}
     \mathcal{L}_{gen}=\mathcal{L}_{fid} + \beta_{tran} \cdot \mathcal{L}_{tran} + \beta_{div} \cdot \mathcal{L}_{div},
\end{align}
where $\beta_{tran}$ and $\beta_{div}$ are tunable hyper-parameters. Of note, the synthetic data generated by a well-trained generator should be visually distinct from the real data for privacy protection, while it can capture the common knowledge of the local models to ensure similarity to the real data distribution for utility. 
More privacy protection is discussed in Appendices~\ref{Vis_syn_samp:} and~\ref{app_discussion:}.

\subsection{Robust Model Distillation} 
Now we update the global model.  
Normally, the global model attempts to learn as much as possible logits outputs of the ensemble model on the synthetic data generated by the generator based on knowledge distillation~\cite{Lin2020Ensemble, Zhang2020The, Zhang2022Fine, Zhang2022DENSE}.
The updated global model and the ensemble model are then served to train $G(\cdot)$ with the goal of generating synthetic data that maximizes the mismatch between them in terms of logits outputs~(see \textit{transferability} discussed in the previous section).
This adversarial game enables the generator to rapidly explore the training space of the local models to help knowledge transfer from them to the global model.
However, it also leads to dramatic shifts in the output distribution of $G(\cdot)$ across communication rounds under heterogeneous FL scenario~(i.e., distribution shifts), causing the global model to catastrophically forget useful knowledge gained in previous rounds.
To tackle the deficiency, we propose to equip the server with a generator $\widetilde{G}(\cdot)$ parameterized by $\widetilde{\bm{w}}$ that is an exponential moving average~(EMA) copy of $G(\cdot)$. 
Its parameters at the $t^{th}$ communication round are computed by
\begin{equation}
    \label{G_tilde:}
     \widetilde{\bm{w}}^{t}=\lambda \cdot \widetilde{\bm{w}}^{t-1}+ (1-\lambda) \cdot \bm{w}^{t},
\end{equation}
where $\lambda \in (0,1)$ is the momentum. 
We can easily see that the parameters of $\widetilde{G}(\cdot)$ vary very little compared to those of $G(\cdot)$ over communication rounds, if $\lambda$ is close to $1$.
We further utilize synthetic data from $\widetilde{G}(\cdot)$ as additional training data for the global model outside of $G(\cdot)$,  mitigating the huge exploratory distribution shift induced by the large update of $G(\cdot)$ and achieving stable updates of the global model.
Particularly, we compute the Kullback-Leibler divergence between logits of the ensemble model and the global model on the synthetic data points $\bm{s}=G(\bm{h}=o(\bm{z}, y), \bm{w})$ and $\widetilde{\bm{s}}=\widetilde{G}(\widetilde{\bm{h}}=o(\widetilde{\bm{z}}, \widetilde{y}), \widetilde{\bm{w}})$ respectively, 
which is formulated as follows:
\begin{align}
    \label{L_kl:}
     \mathcal{L}_{md}=\mathcal{L}_{kl}+\alpha  \widetilde{\mathcal{L}}_{kl} =KL(f(\bm{s}, \bm{\theta}), \sum_{i \in \mathcal{S}_t}\tau_{i, y} f_i(\bm{s}, \bm{\theta}_i)) + \alpha \cdot KL(f(\widetilde{\bm{s}}, \bm{\theta}), \sum_{i \in \mathcal{S}_t}\tau_{i, \widetilde{y}} f_i(\widetilde{\bm{s}}, \bm{\theta}_i)),
\end{align}
where $\alpha$ is a tunable hyper-parameter for balancing different loss items.

\textbf{Dynamic Weighting and Label Sampling.} So far, how to determine $\tau_{i, y}$ and $p(y)$ is unclear. 
The appropriate $\tau_{i, y}$ and $p(y)$ are essential for effective extraction of knowledge from local models.
For clarity, we propose dynamic weighting and label sampling, i.e., $\tau_{i, y} = n_{i,t}^{y} / n_{\mathcal{S}_t, t}^{y}$ and $p(y)=n_{\mathcal{S}_t, t}^{y} / \sum_{y\in[C]}n_{\mathcal{S}_t, t}^{y}$, where $n_{\mathcal{S}_t, t}^{y} = \sum_{ j\in [\mathcal{S}_t]}n_{j,t}^{y}$ and  $n_{i,t}^{y}$ denotes the number of data with label $y$ involved in training on client $i$ at round $t$.
Due to space limitations, see Appendix~\ref{dy_wei_La_sam:} for their detail study and experimental justification.

\section{Experiments}
\label{experiment:}
\subsection{Experimental Settings}
\textbf{Datasets.} In this paper, we evaluate different methods with six real-world image classification task-related datasets, namely Fashion-MNIST~\cite{xiao2017fashion}~(FMNIST in short), SVHN~\cite{Netzer2011Reading}, CIFAR-10, CIFAR-100~\cite{Krizhevsky2009Learning}, Tiny-imageNet\footnote{http://cs231n.stanford.edu/tiny-imagenet-200.zip} and Food101~\cite{Bossard2014Food}.
We detail the six datasets in Appendix~\ref{dataset_app:}.
To simulate data heterogeneity across clients, as in previous works~\cite{Luo2023GradMA, Yurochkin2019Bayesian, Wang2020Federated}, we use Dirichlet process $Dir(\omega)$ to partition the training set for each dataset, thereby allocating local training data for each client.  
It is worth noting that $\omega$ is the concentration parameter and smaller $\omega$ corresponds to stronger data heterogeneity. 

\textbf{Baselines.} 
We compare DFRD to FedFTG~\cite{Zhang2022Fine} and DENSE~\cite{Zhang2022DENSE}, which are the most relevant methods to our work.
To verify the superiority of DFRD, on the one hand, DFRD, FedFTG and DENSE are directly adopted on the server to transfer the knowledge of the local models to a randomly initialized global model. 
We call them collectively \textbf{data-free methods}.
On the other hand, they are utilized as \textbf{fine-tuning methods} to improve the global model's performance after 
computing weighted average per communication round. 
In this case, in each communication round, the preliminary global model is obtained using FedAvg~\cite{McMahan2017Communication} in FL with homogeneous models, whereas in FL with heterogeneous models, the PT-based methods, including HeteroFL~\cite{Diao2020HeteroFL}, Federated Dropout~\cite{Caldas2018Expanding}~(FedDp for short) and FedRolex~\cite{Alam2022FedRolex}, are employed to get the preliminary global model.
\footnote{As an example, DFRD+FedRelox indicates that DFRD is used as a fine-tuning method to improve the performance of FedRelox, and others are similar. See Tables~\ref{table_com_omega:} and~\ref{table_com_rho:} for details.}

\textbf{Configurations.} 
Unless otherwise stated, all experiments are performed on a centralized network with $N=10$ active clients.
We set $\omega \in \{0.01, 0.1, 1.0\}$ to mimic different data heterogeneity scenarios.
To simulate model-heterogeneous scenarios, we formulate exponentially distributed budgets for a given $N$: $R_i = [\frac{1}{2}]^{\min\{\sigma, \lfloor\frac{\rho \cdot i}{N}\rfloor\}} (i \in [N])$, where $\sigma$ and $\rho$ are both positive integers. 
We fix $\sigma=4$ and consider $\rho\in \{5, 10, 40\}$.
See Appendix~\ref{Budget_Distribution_app:} for more details.
Unless otherwise specified, we set $\beta_{tran}$ and $\beta_{div}$ both to $1$ in \textit{training generator}, while in \textit{robust model distillation}, we set $\lambda=0.5$ and $\alpha=0.5$.
And all baselines leverage the same setting as ours.
Due to space limitations, see Appendix~\ref{Com_Exp_Setup:} for the full experimental setup.

\textbf{Evaluation Metrics.} 
We evaluate the performance of different FL methods by local and global test accuracy. 
To be specific, for local test accuracy~(\textit{L.acc} for short), we randomly and evenly distribute the test set to each client and harness the test set on each client to verify the performance of local models.
In terms of global test accuracy~(\textit{G.acc} for short), we employ the global model on the server to evaluate the global performance of different FL methods via utilizing the original test set.
Note that \textit{L.acc} is reported in \textbf{round brackets}.
To ensure reliability, we report the average for each experiment over $3$ different random seeds.

\begin{table}[htbp]
  \centering
  \caption{Top test accuracy~(\%) of distinct methods across $\omega \in \{0.01, 0.1, 1.0\}$ on different datasets.}
    \resizebox{1.0\columnwidth}{!}{
    \begin{tabular}{cp{5.625em}p{5.375em}p{6.315em}|p{5.315em}p{5.875em}p{5.25em}|p{5.565em}p{5.375em}p{6em}|p{5.44em}p{5.25em}p{5.565em}}
    \toprule
    \multirow{2}[4]{*}{Alg.s} & \multicolumn{3}{c|}{FMNIST} & \multicolumn{3}{c|}{SVHN} & \multicolumn{3}{c|}{CIFAR-10} & \multicolumn{3}{c}{CIFAR-100} \\
\cmidrule{2-13}          & \multicolumn{1}{c}{$\omega=1.0$} & \multicolumn{1}{c}{$\omega=0.1$} & \multicolumn{1}{c|}{$\omega=0.01$} & \multicolumn{1}{c}{$\omega=1.0$} & \multicolumn{1}{c}{$\omega=0.1$} & \multicolumn{1}{c|}{$\omega=0.01$} & \multicolumn{1}{c}{$\omega=1.0$} & \multicolumn{1}{c}{$\omega=0.1$} & \multicolumn{1}{c|}{$\omega=0.01$} & \multicolumn{1}{c}{$\omega=1.0$} & \multicolumn{1}{c}{$\omega=0.1$} & \multicolumn{1}{c}{$\omega=0.01$} \\
    \midrule
    DENSE & 13.99{\scriptsize$\pm$4.39} (74.26{\scriptsize$\pm$2.64}) & 13.96{\scriptsize$\pm$1.33} (30.27{\scriptsize$\pm$1.10}) & 13.83{\scriptsize$\pm$2.88} (14.82{\scriptsize$\pm$3.55}) & 16.12{\scriptsize$\pm$2.08} (52.67{\scriptsize$\pm$1.68}) & 14.53{\scriptsize$\pm$3.56} (25.95{\scriptsize$\pm$0.42}) & 13.06{\scriptsize$\pm$5.40} (13.09{\scriptsize$\pm$1.64}) & 10.47{\scriptsize$\pm$1.21} (46.89{\scriptsize$\pm$0.97}) & 10.28{\scriptsize$\pm$0.60} (22.67{\scriptsize$\pm$0.66}) & 10.96{\scriptsize$\pm$1.93} (13.03{\scriptsize$\pm$1.56}) & 1.22{\scriptsize$\pm$0.10} (21.68{\scriptsize$\pm$0.23}) & 1.11{\scriptsize$\pm$0.03} (13.61{\scriptsize$\pm$0.30}) & 1.11{\scriptsize$\pm$0.02} (8.73{\scriptsize$\pm$0.09}) \\
    \rowcolor{Gray}
    FedFTG & 48.10{\scriptsize$\pm$1.61} (73.96{\scriptsize$\pm$2.51}) & 44.44{\scriptsize$\pm$1.30} (30.33{\scriptsize$\pm$1.43}) & 26.96{\scriptsize$\pm$8.33} (14.79{\scriptsize$\pm$3.61}) & 18.58{\scriptsize$\pm$1.41} (54.75{\scriptsize$\pm$0.63}) & 17.42{\scriptsize$\pm$2.03} (26.39{\scriptsize$\pm$0.38}) & 15.44{\scriptsize$\pm$4.12} (\textbf{13.15{\scriptsize$\pm$1.63}}) & 28.73{\scriptsize$\pm$1.03} (47.67{\scriptsize$\pm$0.92}) & 21.97{\scriptsize$\pm$1.18} (22.95{\scriptsize$\pm$0.64}) & 15.18{\scriptsize$\pm$0.72} (13.06{\scriptsize$\pm$1.54}) & 1.05{\scriptsize$\pm$0.09} (22.67{\scriptsize$\pm$0.44}) & 1.06{\scriptsize$\pm$0.10} (14.01{\scriptsize$\pm$0.49}) & 1.11{\scriptsize$\pm$0.09} (8.82{\scriptsize$\pm$0.21}) \\
        DFRD & \textbf{65.38{\scriptsize$\pm$0.72}} (\textbf{74.44{\scriptsize$\pm$2.47}}) & \textbf{52.55{\scriptsize$\pm$1.61}} (\textbf{30.44{\scriptsize$\pm$1.42}}) & \textbf{36.63{\scriptsize$\pm$6.52}} (\textbf{14.87{\scriptsize$\pm$3.64}}) & \textbf{24.13{\scriptsize$\pm$0.96}} (\textbf{55.48{\scriptsize$\pm$1.56}}) & \textbf{18.53{\scriptsize$\pm$0.42}} (\textbf{26.87{\scriptsize$\pm$0.11}}) & \textbf{18.18{\scriptsize$\pm$0.56}} (12.83{\scriptsize$\pm$1.90}) & \textbf{33.82{\scriptsize$\pm$1.81}} (\textbf{48.33{\scriptsize$\pm$1.17}}) & \textbf{24.58{\scriptsize$\pm$0.59}} (\textbf{23.21{\scriptsize$\pm$0.70}}) & \textbf{20.59{\scriptsize$\pm$2.93}} (\textbf{14.18{\scriptsize$\pm$1.53}}) & \textbf{3.21{\scriptsize$\pm$0.36}} (\textbf{23.31{\scriptsize$\pm$0.39}}) & \textbf{1.55{\scriptsize$\pm$0.13}} (\textbf{14.15{\scriptsize$\pm$0.32}}) & \textbf{1.36{\scriptsize$\pm$0.06}} (\textbf{8.93{\scriptsize$\pm$0.13}}) \\
    \midrule
    FedAvg & 89.71{\scriptsize$\pm$0.10} (84.57{\scriptsize$\pm$0.63}) & 83.24{\scriptsize$\pm$1.41} (62.48{\scriptsize$\pm$5.86}) & 60.89{\scriptsize$\pm$5.24} (29.81{\scriptsize$\pm$13.98}) & 70.63{\scriptsize$\pm$1.95} (54.21{\scriptsize$\pm$2.54}) & 58.12{\scriptsize$\pm$1.26} (24.66{\scriptsize$\pm$0.42}) & 32.22{\scriptsize$\pm$1.16} (12.57{\scriptsize$\pm$1.81}) & 67.16{\scriptsize$\pm$0.98} (49.82{\scriptsize$\pm$1.47}) & 56.36{\scriptsize$\pm$2.09} (21.92{\scriptsize$\pm$0.83}) & 32.92{\scriptsize$\pm$7.40} (12.37{\scriptsize$\pm$1.71}) & 57.31{\scriptsize$\pm$0.09} (47.01{\scriptsize$\pm$0.99}) & 49.50{\scriptsize$\pm$0.51} (25.33{\scriptsize$\pm$0.88}) & 39.22{\scriptsize$\pm$1.08} (11.25{\scriptsize$\pm$0.52}) \\
    \rowcolor{Gray}
    +DENSE & 90.13{\scriptsize$\pm$0.14} (84.81{\scriptsize$\pm$0.60}) & 84.10{\scriptsize$\pm$1.55} (62.80{\scriptsize$\pm$5.76}) & 
    63.36{\scriptsize$\pm$6.24}
    (29.37{\scriptsize$\pm$13.83}) & 74.48{\scriptsize$\pm$1.48} (58.02{\scriptsize$\pm$2.65}) & 62.41{\scriptsize$\pm$2.36} (26.23{\scriptsize$\pm$0.33}) & 32.84{\scriptsize$\pm$12.37} (12.72{\scriptsize$\pm$1.82}) & 69.73{\scriptsize$\pm$0.60} (53.46{\scriptsize$\pm$1.58}) & 57.48{\scriptsize$\pm$3.13} (22.54{\scriptsize$\pm$0.76}) & 35.85{\scriptsize$\pm$7.22} (12.41{\scriptsize$\pm$1.72}) & 58.43{\scriptsize$\pm$0.05} (48.00{\scriptsize$\pm$0.78}) & 50.84{\scriptsize$\pm$0.60} (26.19{\scriptsize$\pm$0.90}) & 40.24{\scriptsize$\pm$1.04} (11.87{\scriptsize$\pm$0.58}) \\
    +FedFTG & 91.15{\scriptsize$\pm$0.23} (85.88{\scriptsize$\pm$0.70}) & 84.93{\scriptsize$\pm$1.40} (63.14{\scriptsize$\pm$4.64}) & 64.80{\scriptsize$\pm$6.56} (\textbf{29.89{\scriptsize$\pm$14.08}}) & 73.91{\scriptsize$\pm$1.78} (56.99{\scriptsize$\pm$2.65}) & 61.12{\scriptsize$\pm$1.69} (25.82{\scriptsize$\pm$0.28}) & 37.07{\scriptsize$\pm$6.55} (12.69{\scriptsize$\pm$1.80}) & 68.85{\scriptsize$\pm$0.96} (51.05{\scriptsize$\pm$1.59}) & 58.17{\scriptsize$\pm$2.02} (22.39{\scriptsize$\pm$0.66}) & 33.93{\scriptsize$\pm$7.69} (12.39{\scriptsize$\pm$1.70}) & 58.81{\scriptsize$\pm$1.39} (48.06{\scriptsize$\pm$1.79}) & 50.70{\scriptsize$\pm$1.11} (26.40{\scriptsize$\pm$0.47}) & 40.50{\scriptsize$\pm$1.17} (12.15{\scriptsize$\pm$0.61}) \\
    \rowcolor{Gray}
    +DFRD & \textbf{91.65{\scriptsize$\pm$0.09}} (\textbf{86.16{\scriptsize$\pm$0.53}}) & \textbf{85.66{\scriptsize$\pm$1.21}} (\textbf{63.79{\scriptsize$\pm$4.29}}) & \textbf{65.23{\scriptsize$\pm$7.17}} (29.58{\scriptsize$\pm$13.45}) & \textbf{77.84{\scriptsize$\pm$1.39}} (\textbf{60.31{\scriptsize$\pm$2.51}}) & \textbf{65.79{\scriptsize$\pm$2.27}} (\textbf{27.26{\scriptsize$\pm$0.20}}) & \textbf{39.25{\scriptsize$\pm$11.71}} (\textbf{13.96{\scriptsize$\pm$1.48}}) & \textbf{71.54{\scriptsize$\pm$0.66}} (\textbf{53.59{\scriptsize$\pm$1.19}}) & \textbf{61.34{\scriptsize$\pm$1.21}} (\textbf{23.00{\scriptsize$\pm$0.68}}) & \textbf{40.77{\scriptsize$\pm$7.47}} (\textbf{13.49{\scriptsize$\pm$1.57}}) & \textbf{61.49{\scriptsize$\pm$0.28}} (\textbf{49.97{\scriptsize$\pm$0.58}}) & \textbf{53.95{\scriptsize$\pm$0.67}} (\textbf{29.10{\scriptsize$\pm$0.87}}) & \textbf{43.47{\scriptsize$\pm$1.05}} (\textbf{14.09{\scriptsize$\pm$1.09}}) \\
    \bottomrule
    \end{tabular}
    } %
  \label{table_com_omega:}%
\end{table}%

\subsection{Results Comparison}

\begin{figure*}[h]\captionsetup[subfigure]{font=scriptsize}
    \centering
    \begin{subfigure}{0.329\linewidth}
        \centering
        \includegraphics[width=1.0\linewidth]{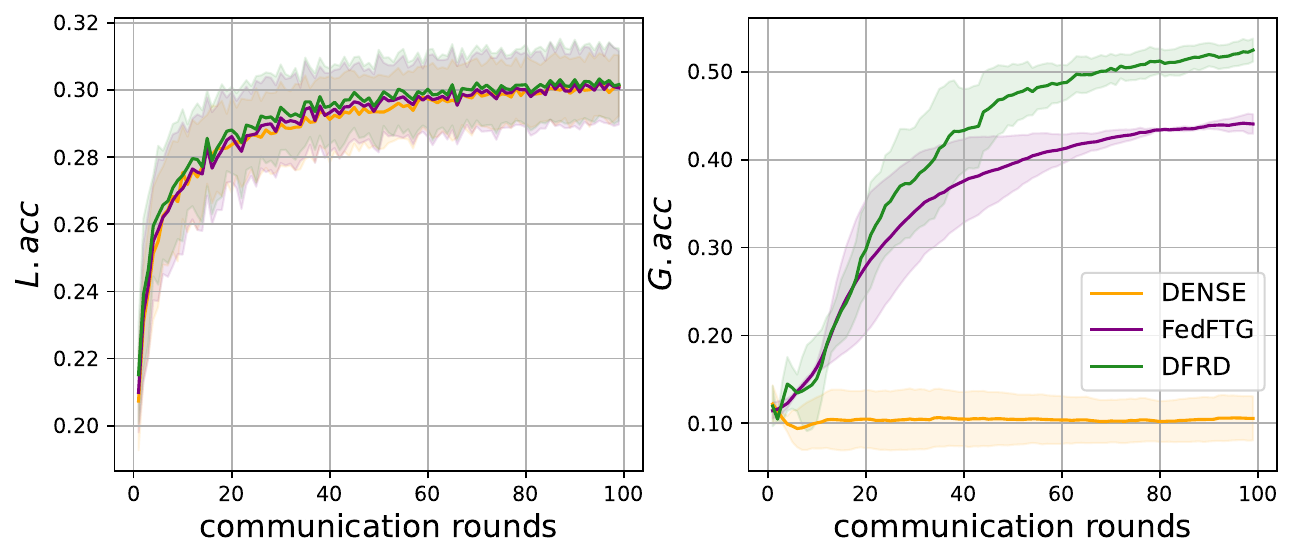}
        \caption{FMNIST, $\omega=0.1$}
    \end{subfigure}
    \centering
    \begin{subfigure}{0.329\linewidth}
        \centering
        \includegraphics[width=1.0\linewidth]{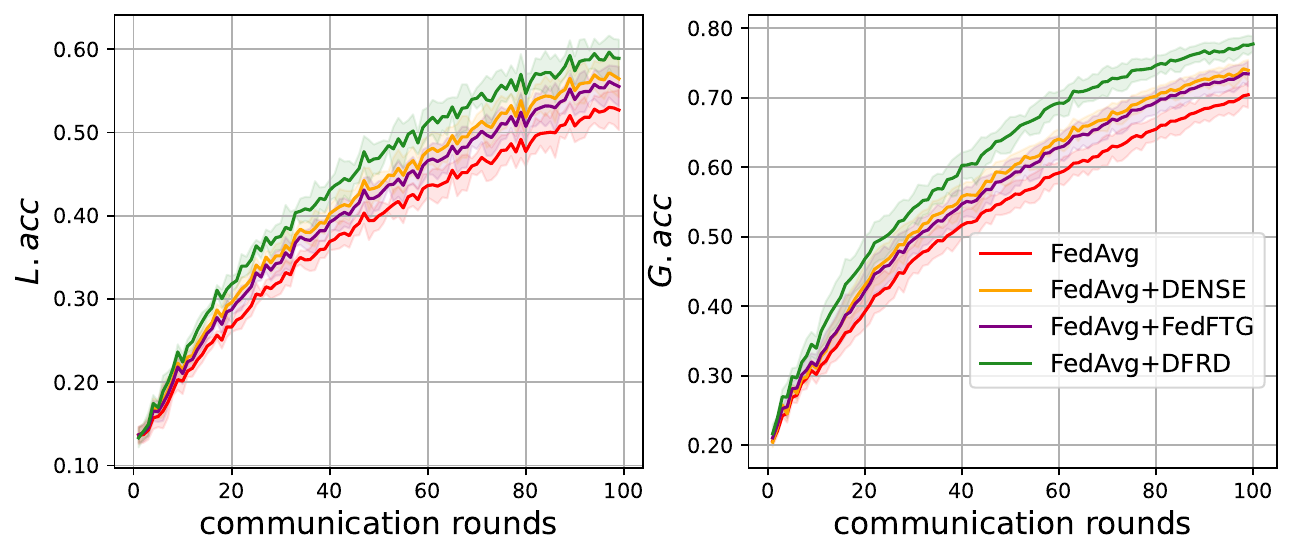}
        \caption{SVHN, $\omega=1.0$}
    \end{subfigure}
    \centering
    \begin{subfigure}{0.329\linewidth}
        \centering
        \includegraphics[width=1.0\linewidth]{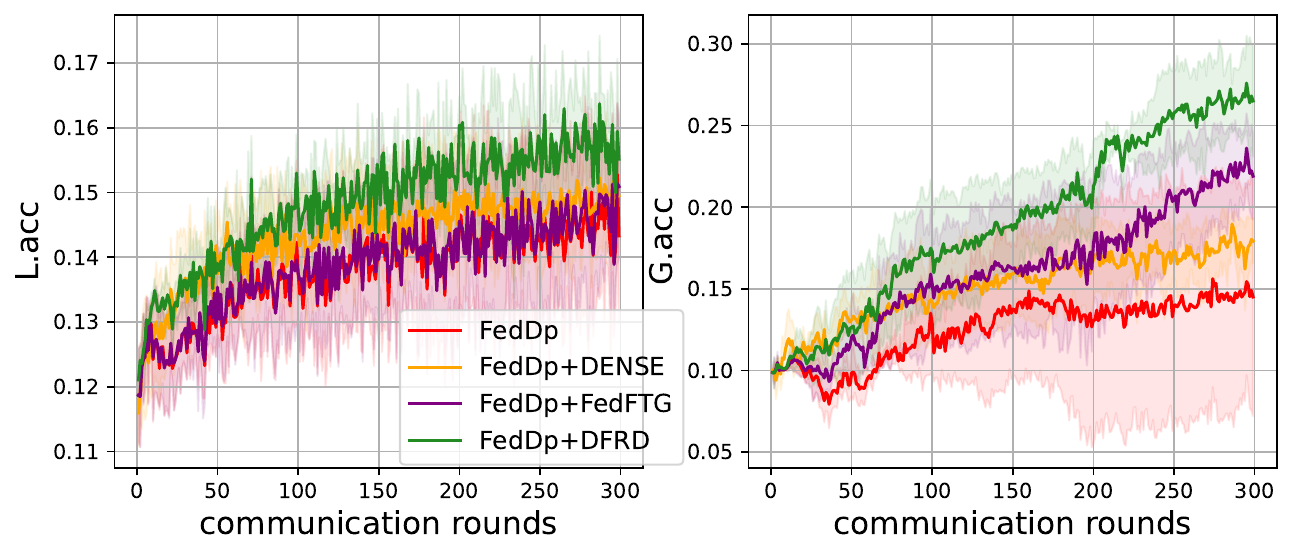}
        \caption{CIFAR-10, $\rho=5$}
    \end{subfigure}
    \caption{Accuracy curves selected of DFRD and baselines on FMNIST, SVHN and CIFAR-10.} 
    \label{sel_acc_curve:}
\end{figure*}

\textbf{Impacts of varying $\omega$.}
We study the performance of different methods at different levels of data heterogeneity on FMNIST, SVHN, CIFAR-10 and CIFAR-100, as shown in Table~\ref{table_com_omega:}.
One can see that the performance of all methods degrades severely as $\omega$ decreases, with DFRD being the only method that is robust while consistently leading other baselines with an overwhelming margin w.r.t. \textit{G.acc}.
Also, Fig.~\ref{sel_acc_curve:}~(a)-(b) show that the learning efficiency of DFRD consistently beats other baselines~(see Fig.~\ref{model_homo_data_free:}-\ref{model_homo_fine_tune:} in Appendix~\ref{full_learn_curve:} for complete curves).
Notably, DFRD, FedFTG and DENSE as fine-tuning methods uniformly surpass FedAvg w.r.t. \textit{G.acc} and \textit{L.acc}.
However, their global test accuracies suffer from dramatic deterioration or even substantially worse than that of FedAvg when they act as data-free methods.
We conjecture that FedAvg aggregates the knowledge of local models more effectively than data-free methods.
Also, when DFRD is used to fine-tune FedAvg, it can significantly enhance the global model, yet improve the performance of local models to a less extent.

\textbf{Impacts of different $\rho$.}
We explore the impacts of different model heterogeneity distributions on different methods with SVHN, CIFAR-10, Tiny-ImageNet and FOOD101. 
A higher $\rho$ means more clients with $\frac{1}{16}$-width capacity w.r.t. the global model.
From Table~\ref{table_com_rho:}, we can clearly see that the performance of all methods improves uniformly with decreasing $\rho$, where DFRD consistently and overwhelmingly dominates other baselines in terms of $G.acc$.
Specifically, DFRD improves $G.acc$ by an average of $11.07\%$ and $7.54\%$ on SVHN and CIFAR-10 respectively, compared to PT-based methods~(including HeteroFL, FedDp and FedRolex).
Meanwhile, DFRD uniformly and significantly outstrips FedFTG and DENSE w.r.t. $G.acc$.
The selected learning curve shown in Fig.~\ref{sel_acc_curve:} (c) also verifies the above statement (see Fig.~\ref{model_static_heter:}-\ref{model_roll_heter:} in Appendix~\ref{full_learn_curve:} for more results). 
The above empirical results show that DFRD not only is robust to varying $\rho$, but also has significantly intensified effects on the global model for different PT-based methods.
However, the results on Tiny-ImageNet and FOOD101 indicate that PT-based methods suffer from inferior test accuracy.
Although DFRD improves their test accuracy, the improvement is slight.
Notably, DFRD improves negligibly over PT-based methods when all clients exemplify $\frac{1}{16}$-width capability.
We thus argue that weak clients performing complex image classification tasks learn little useful local knowledge, resulting in the inability to provide effective information for the global model.

\begin{table}[htbp]
  \centering
  \caption{Top test accuracy~(\%) of distinct methods across $\rho \in \{5, 10, 40\}$ on different datasets.}
    \resizebox{1.0\columnwidth}{!}{
    \begin{tabular}{cp{5.565em}p{5.25em}p{5.565em}|p{5.75em}p{5.815em}p{5.75em}|p{4.815em}p{5.065em}p{5.815em}|p{4.94em}p{5.94em}p{5.69em}}
    \toprule
    \multirow{2}[4]{*}{Alg.s} & \multicolumn{3}{c|}{SVHN} & \multicolumn{3}{c|}{CIFAR-10} & \multicolumn{3}{c|}{Tiny-ImageNet} & \multicolumn{3}{c}{FOOD101} \\
\cmidrule{2-13}    & \multicolumn{1}{c}{$\rho=5$} & \multicolumn{1}{c}{$\rho=10$} & \multicolumn{1}{c|}{$\rho=40$} & \multicolumn{1}{c}{$\rho=5$} & \multicolumn{1}{c}{$\rho=10$} & \multicolumn{1}{c|}{$\rho=40$} & \multicolumn{1}{c}{$\rho=5$} & \multicolumn{1}{c}{$\rho=10$} & \multicolumn{1}{c|}{$\rho=40$} & \multicolumn{1}{c}{$\rho=5$} & \multicolumn{1}{c}{$\rho=10$} & \multicolumn{1}{c}{$\rho=40$} \\
    \midrule
    HeteroFL & 29.56{\scriptsize$\pm$17.60} (27.63{\scriptsize$\pm$5.35}) & 23.07{\scriptsize$\pm$12.92} (30.90{\scriptsize$\pm$5.70}) & 17.37{\scriptsize$\pm$9.50} (25.28{\scriptsize$\pm$1.13}) & 21.00{\scriptsize$\pm$3.28} (23.94{\scriptsize$\pm$2.63}) & 13.89{\scriptsize$\pm$3.73} (\textbf{24.70{\scriptsize$\pm$0.80}}) & 11.16{\scriptsize$\pm$1.95} (26.55{\scriptsize$\pm$1.29}) & 8.27{\scriptsize$\pm$0.21} (6.72{\scriptsize$\pm$0.13}) & 1.03{\scriptsize$\pm$0.20} (5.46{\scriptsize$\pm$0.33}) & 0.73{\scriptsize$\pm$0.11} (6.90{\scriptsize$\pm$0.23}) & 9.61{\scriptsize$\pm$0.97} (7.79{\scriptsize$\pm$0.49}) & 1.52{\scriptsize$\pm$0.17} (9.12{\scriptsize$\pm$0.66}) & 1.16{\scriptsize$\pm$0.07} (12.05{\scriptsize$\pm$0.37}) \\
    \rowcolor{Gray}
    +DENSE & 31.16{\scriptsize$\pm$15.65} (28.62{\scriptsize$\pm$4.62}) & 27.48{\scriptsize$\pm$9.61} (31.51{\scriptsize$\pm$4.27}) & 20.47{\scriptsize$\pm$6.24} (24.88{\scriptsize$\pm$1.82}) & 21.63{\scriptsize$\pm$2.83} (24.05{\scriptsize$\pm$2.52}) & 16.64{\scriptsize$\pm$2.86} (24.34{\scriptsize$\pm$0.73}) & 12.79{\scriptsize$\pm$0.28} (\textbf{26.80{\scriptsize$\pm$1.17}}) & 8.44{\scriptsize$\pm$0.32} (6.74{\scriptsize$\pm$0.28}) & 1.07{\scriptsize$\pm$0.16} (5.86{\scriptsize$\pm$0.80}) & 0.75{\scriptsize$\pm$0.08} (7.15{\scriptsize$\pm$0.35}) & 9.78{\scriptsize$\pm$0.85} (7.98{\scriptsize$\pm$0.43}) & 1.90{\scriptsize$\pm$0.39} (8.99{\scriptsize$\pm$0.66}) & 1.24{\scriptsize$\pm$0.11} (11.99{\scriptsize$\pm$0.33}) \\
    +FedFTG & 32.22{\scriptsize$\pm$15.66} (29.10{\scriptsize$\pm$5.00}) & 26.92{\scriptsize$\pm$8.86} (\textbf{31.56{\scriptsize$\pm$4.45}}) & 19.02{\scriptsize$\pm$4.51} (25.33{\scriptsize$\pm$1.30}) & 22.29{\scriptsize$\pm$4.01} (24.34{\scriptsize$\pm$2.64}) & 18.79{\scriptsize$\pm$4.58} (24.47{\scriptsize$\pm$0.85}) & 15.69{\scriptsize$\pm$2.91} (26.46{\scriptsize$\pm$1.12}) & 8.38{\scriptsize$\pm$0.20} (6.63{\scriptsize$\pm$0.32}) & 1.08{\scriptsize$\pm$0.17} (6.01{\scriptsize$\pm$0.69}) & 0.74{\scriptsize$\pm$0.10} (7.11{\scriptsize$\pm$0.51}) & 9.85{\scriptsize$\pm$1.07} (7.88{\scriptsize$\pm$0.47}) & 1.94{\scriptsize$\pm$0.44} (8.74{\scriptsize$\pm$0.56}) & 1.26{\scriptsize$\pm$0.17} (11.87{\scriptsize$\pm$0.16}) \\
    \rowcolor{Gray}
     +DFRD & \textbf{42.77{\scriptsize$\pm$12.60}} (\textbf{29.41{\scriptsize$\pm$5.53}}) & \textbf{30.17{\scriptsize$\pm$7.26}} (31.56{\scriptsize$\pm$4.56}) & \textbf{24.82{\scriptsize$\pm$6.70}} (\textbf{26.14{\scriptsize$\pm$0.47}}) & \textbf{24.30{\scriptsize$\pm$1.59}} (\textbf{24.70{\scriptsize$\pm$1.58}}) & \textbf{23.78{\scriptsize$\pm$1.77}} (24.35{\scriptsize$\pm$0.67}) & \textbf{19.10{\scriptsize$\pm$0.78}} (26.75{\scriptsize$\pm$1.34}) & \textbf{9.27{\scriptsize$\pm$0.14}} (\textbf{8.09{\scriptsize$\pm$0.20}}) & \textbf{1.50{\scriptsize$\pm$0.12}} (\textbf{8.13{\scriptsize$\pm$0.33}}) & \textbf{0.83{\scriptsize$\pm$0.01}} (\textbf{10.29{\scriptsize$\pm$0.18}}) & \textbf{11.54{\scriptsize$\pm$0.32}} (\textbf{9.28{\scriptsize$\pm$0.07}}) & \textbf{3.05{\scriptsize$\pm$0.73}} (\textbf{10.94{\scriptsize$\pm$0.07}}) & \textbf{1.58{\scriptsize$\pm$0.23}} (\textbf{14.39{\scriptsize$\pm$0.02}}) \\
    \midrule
    FedDP & 28.99{\scriptsize$\pm$17.35} (16.96{\scriptsize$\pm$5.50}) & 24.17{\scriptsize$\pm$18.36} (14.95{\scriptsize$\pm$2.59}) & 20.03{\scriptsize$\pm$13.90} (14.38{\scriptsize$\pm$1.76})  & 17.36{\scriptsize$\pm$3.13} (15.47{\scriptsize$\pm$1.06}) & 12.52{\scriptsize$\pm$1.41} (15.48{\scriptsize$\pm$0.87}) & 11.08{\scriptsize$\pm$1.83} (15.83{\scriptsize$\pm$1.07}) & 6.33{\scriptsize$\pm$0.78} (2.84{\scriptsize$\pm$0.30}) & 1.15{\scriptsize$\pm$0.17} (1.14{\scriptsize$\pm$0.28}) & 0.97{\scriptsize$\pm$0.13} (0.97{\scriptsize$\pm$0.10}) & 7.69{\scriptsize$\pm$0.58} (3.42{\scriptsize$\pm$0.22}) & 1.84{\scriptsize$\pm$0.59} (1.70{\scriptsize$\pm$0.32}) & 1.75{\scriptsize$\pm$0.28} (1.70{\scriptsize$\pm$0.28}) \\
    \rowcolor{Gray}
    +DENSE & 31.15{\scriptsize$\pm$22.51} (18.39{\scriptsize$\pm$6.36}) & 26.38{\scriptsize$\pm$18.60} (15.73{\scriptsize$\pm$4.00}) & 22.33{\scriptsize$\pm$14.46} (14.65{\scriptsize$\pm$2.93}) & 19.66{\scriptsize$\pm$1.15} (15.63{\scriptsize$\pm$0.95}) & 14.55{\scriptsize$\pm$1.50} (15.67{\scriptsize$\pm$1.24}) & 12.22{\scriptsize$\pm$1.89} (15.48{\scriptsize$\pm$1.21}) & 6.54{\scriptsize$\pm$0.39} (2.75{\scriptsize$\pm$0.29}) & 1.23{\scriptsize$\pm$0.25} (1.17{\scriptsize$\pm$0.33}) & 1.03{\scriptsize$\pm$0.10} (0.99{\scriptsize$\pm$0.11}) & 8.19{\scriptsize$\pm$0.91} (3.59{\scriptsize$\pm$0.23}) & 1.98{\scriptsize$\pm$0.33} (1.69{\scriptsize$\pm$0.44}) & 1.66{\scriptsize$\pm$0.49} (1.69{\scriptsize$\pm$0.25}) \\
    +FedFTG & 32.30{\scriptsize$\pm$20.90} (18.00{\scriptsize$\pm$6.10}) & 28.55{\scriptsize$\pm$14.65} (15.04{\scriptsize$\pm$2.80}) & 26.63{\scriptsize$\pm$12.64} (14.39{\scriptsize$\pm$1.60}) & 23.67{\scriptsize$\pm$2.70} (15.68{\scriptsize$\pm$1.24}) & 13.80{\scriptsize$\pm$0.41} (15.51{\scriptsize$\pm$0.73}) & 12.17{\scriptsize$\pm$1.87} (15.61{\scriptsize$\pm$1.24}) & 6.32{\scriptsize$\pm$0.53} (2.84{\scriptsize$\pm$0.25}) & 1.22{\scriptsize$\pm$0.19} (1.16{\scriptsize$\pm$0.31}) & 0.98{\scriptsize$\pm$0.21} (0.96{\scriptsize$\pm$0.17}) & 7.93{\scriptsize$\pm$0.49} (3.44{\scriptsize$\pm$0.22}) & 2.08{\scriptsize$\pm$0.24} (1.65{\scriptsize$\pm$0.45}) & 1.69{\scriptsize$\pm$0.25} (1.68{\scriptsize$\pm$0.28}) \\
    \rowcolor{Gray}
    +DFRD & \textbf{41.48{\scriptsize$\pm$8.11}} (\textbf{19.02{\scriptsize$\pm$4.75}}) & \textbf{37.30{\scriptsize$\pm$12.22}} (\textbf{17.19{\scriptsize$\pm$0.28}}) & \textbf{32.89{\scriptsize$\pm$11.65}} (\textbf{16.14{\scriptsize$\pm$0.30}}) & \textbf{27.93{\scriptsize$\pm$3.63}} (\textbf{16.69{\scriptsize$\pm$0.82}}) & \textbf{20.24{\scriptsize$\pm$1.44}} (\textbf{16.35{\scriptsize$\pm$0.95}}) & \textbf{19.42{\scriptsize$\pm$4.37}} (\textbf{15.97{\scriptsize$\pm$1.25}}) & \textbf{9.01{\scriptsize$\pm$0.42}} (\textbf{3.83{\scriptsize$\pm$0.07}}) & \textbf{1.59{\scriptsize$\pm$0.21}} (\textbf{1.56{\scriptsize$\pm$0.20}}) & \textbf{1.05{\scriptsize$\pm$0.23}} (\textbf{1.25{\scriptsize$\pm$0.08}}) & \textbf{11.28{\scriptsize$\pm$0.44}} (\textbf{4.62{\scriptsize$\pm$0.13}}) & \textbf{2.71{\scriptsize$\pm$0.24}} (\textbf{2.14{\scriptsize$\pm$0.07}}) & \textbf{1.94{\scriptsize$\pm$0.12}} (\textbf{1.82{\scriptsize$\pm$0.01}}) \\
    \midrule
    FedRolex & 34.71{\scriptsize$\pm$13.68} (14.20{\scriptsize$\pm$2.94}) & 23.48{\scriptsize$\pm$13.09} (13.85{\scriptsize$\pm$2.60}) & 22.39{\scriptsize$\pm$15.28} (13.59{\scriptsize$\pm$2.11}) & 21.11{\scriptsize$\pm$1.76} (16.99{\scriptsize$\pm$0.67}) & 16.57{\scriptsize$\pm$4.40} (16.12{\scriptsize$\pm$0.90}) & 14.37{\scriptsize$\pm$2.91} (17.11{\scriptsize$\pm$1.29}) & 9.29{\scriptsize$\pm$0.32} (5.33{\scriptsize$\pm$0.21}) & 5.55{\scriptsize$\pm$0.40} (2.73{\scriptsize$\pm$0.09}) & 2.50{\scriptsize$\pm$0.33} (1.81{\scriptsize$\pm$0.29}) & 10.27{\scriptsize$\pm$1.33} (6.86{\scriptsize$\pm$0.12}) & 5.14{\scriptsize$\pm$3.41} (3.37{\scriptsize$\pm$1.14}) & 3.22{\scriptsize$\pm$0.52} (2.95{\scriptsize$\pm$0.17}) \\
    \rowcolor{Gray}
    +DENSE & 36.58{\scriptsize$\pm$12.53} (17.51{\scriptsize$\pm$2.52}) & 26.69{\scriptsize$\pm$13.11} (14.49{\scriptsize$\pm$1.96}) & 24.34{\scriptsize$\pm$14.81} (14.04{\scriptsize$\pm$1.95}) & 23.72{\scriptsize$\pm$5.48} (17.16{\scriptsize$\pm$0.67}) & 19.65{\scriptsize$\pm$1.47} (15.81{\scriptsize$\pm$0.62}) & 16.44{\scriptsize$\pm$1.89} (17.26{\scriptsize$\pm$1.34}) & 9.33{\scriptsize$\pm$0.06} (5.16{\scriptsize$\pm$0.18}) & 5.40{\scriptsize$\pm$0.40} (2.76{\scriptsize$\pm$0.01}) & 2.40{\scriptsize$\pm$0.20} (1.82{\scriptsize$\pm$0.33}) & 10.83{\scriptsize$\pm$0.78} (6.95{\scriptsize$\pm$0.26}) & 7.54{\scriptsize$\pm$0.46} (3.86{\scriptsize$\pm$0.44}) & 3.07{\scriptsize$\pm$0.53} (2.99{\scriptsize$\pm$0.11}) \\
    +FedFTG & 38.07{\scriptsize$\pm$12.27} (17.64{\scriptsize$\pm$2.54}) & 25.53{\scriptsize$\pm$13.84} (14.51{\scriptsize$\pm$2.21}) & 24.06{\scriptsize$\pm$14.86} (14.03{\scriptsize$\pm$1.91}) & 22.66{\scriptsize$\pm$5.24} (17.16{\scriptsize$\pm$1.03}) & 17.79{\scriptsize$\pm$2.67} (16.25{\scriptsize$\pm$1.56}) & 14.77{\scriptsize$\pm$2.03} (17.70{\scriptsize$\pm$1.14}) & 9.36{\scriptsize$\pm$0.23} (5.18{\scriptsize$\pm$0.18}) & 5.68{\scriptsize$\pm$0.34} (2.75{\scriptsize$\pm$0.05}) & 2.43{\scriptsize$\pm$0.12} (1.81{\scriptsize$\pm$0.23}) & 10.66{\scriptsize$\pm$0.79} (6.85{\scriptsize$\pm$0.11}) & 8.13{\scriptsize$\pm$0.82} (3.75{\scriptsize$\pm$0.48}) & 3.06{\scriptsize$\pm$0.48} (2.92{\scriptsize$\pm$0.19}) \\
    \rowcolor{Gray}
    +DFRD & \textbf{46.30{\scriptsize$\pm$10.12}} (\textbf{18.44{\scriptsize$\pm$1.34}}) & \textbf{34.78{\scriptsize$\pm$9.19}} (\textbf{15.99{\scriptsize$\pm$1.53}}) & \textbf{32.86{\scriptsize$\pm$15.54}} (\textbf{15.17{\scriptsize$\pm$0.66}}) & \textbf{26.68{\scriptsize$\pm$1.21}} (\textbf{17.51{\scriptsize$\pm$0.33}}) & \textbf{25.57{\scriptsize$\pm$1.37}} (\textbf{16.74{\scriptsize$\pm$0.72}}) & \textbf{19.86{\scriptsize$\pm$2.76}} (\textbf{17.77{\scriptsize$\pm$1.16}}) & \textbf{10.93{\scriptsize$\pm$0.05}} (\textbf{6.11{\scriptsize$\pm$0.02}}) & \textbf{6.80{\scriptsize$\pm$0.11}} (\textbf{3.35{\scriptsize$\pm$0.04}}) & \textbf{2.68{\scriptsize$\pm$0.19}} (\textbf{2.22{\scriptsize$\pm$0.08}}) & \textbf{12.70{\scriptsize$\pm$0.79}} (\textbf{8.02{\scriptsize$\pm$0.08}}) & \textbf{10.58{\scriptsize$\pm$0.29}} (\textbf{4.86{\scriptsize$\pm$0.16}}) & \textbf{3.59{\scriptsize$\pm$0.05}} (\textbf{3.34{\scriptsize$\pm$0.15}}) \\
    \bottomrule
    \end{tabular}}%
  \label{table_com_rho:}%
\end{table}%

\subsection{Ablation Study}
\label{Ablation_study:}
In this section, we carefully demonstrate the efficacy and indispensability of core modules and key parameters in our method on SVHN, CIFAR-10 and CIFAR-100.
Thereafter, we resort to FedRolex+DFRD to yield all results.
For SVHN~(CIFAR-10, CIFAR-100), we set $\omega=0.1~(0.1, 1.0)$ and $\rho=10~(10, 5)$. 
Note that we also study the performance of DFRD with different numbers of clients and stragglers, see Appendix~\ref{client_stragglers:} for details.

\begin{wraptable}{r}{7.3cm}
  \centering
  \caption{Test accuracy~(\%) comparison among different transferability constraints over SVHN and CIFAR-10/100.}
    \resizebox{0.5\columnwidth}{!}{
    \begin{tabular}{cp{6.25em}|p{6.25em}|p{6.69em}}
    \toprule
    T. C. & \multicolumn{1}{c|}{SVHN} & \multicolumn{1}{c|}{CIFAR-10} & \multicolumn{1}{c}{CIFAR-100} \\
    \midrule
    $\bigtriangleup$     & 34.09{\scriptsize $\pm$11.45} (15.38{\scriptsize $\pm$2.50}) & 23.45{\scriptsize $\pm$1.13} (15.91{\scriptsize $\pm$1.24}) & 26.62{\scriptsize $\pm$2.39} (12.76{\scriptsize $\pm$0.51}) \\
    \midrule
    $\bigtriangledown$     & 33.89{\scriptsize $\pm$11.22} (15.28{\scriptsize $\pm$2.59}) & 24.27{\scriptsize $\pm$1.82} (16.45{\scriptsize $\pm$1.00}) & 27.18{\scriptsize $\pm$1.78} (12.86{\scriptsize $\pm$0.38}) \\
    \midrule
    $\Diamond$     & \textbf{34.78{\scriptsize $\pm$9.19}} (\textbf{15.99{\scriptsize $\pm$1.53}}) & \textbf{25.57{\scriptsize $\pm$1.37}} (\textbf{16.74{\scriptsize $\pm$0.72}}) & \textbf{28.08{\scriptsize $\pm$0.94}} (\textbf{13.03{\scriptsize $\pm$0.16}}) \\
    \bottomrule
    \end{tabular}}
  \label{table_data_gen:}
\end{wraptable}%
\textbf{Impacts of different transferability constraints.}
It can be observed from Table~\ref{table_data_gen:}  that our proposed transferability constraint $\Diamond$ uniformly beats $\bigtriangleup$ and $\bigtriangledown$ w.r.t. global test performance over three datasets.
This means that $\Diamond$ can guide the generator to generate more effective  synthetic data, thereby improving the test performance of the global model.
Also, we conjecture that generating more synthetic data like those with the black circles~(see Fig.~\ref{pic_Dis_dia:}) may positively impact on the performance of local models, since $\Diamond$ also slightly and consistently trumps other competitors w.r.t. $L.acc$. 

\begin{wraptable}{r}{7.3cm}
  \centering
  \caption{Test accuracy~(\%) comparison among different merger operators over SVHN and CIFAR-10/100.}
    \resizebox{0.5\columnwidth}{!}{
    \begin{tabular}{cp{6.69em}|p{6.315em}|p{6.19em}}
    \toprule
    \multicolumn{1}{c}{M. O.} & \multicolumn{1}{c|}{SVHN} & \multicolumn{1}{c|}{CIFAR-10} & \multicolumn{1}{c}{CIFAR-100} \\
    \midrule
    $mul$   & \textbf{34.78{\scriptsize $\pm$9.19}} (15.99{\scriptsize $\pm$1.53}) & \textbf{25.57{\scriptsize $\pm$1.37}} (16.74{\scriptsize $\pm$0.72}) & \textbf{28.08{\scriptsize $\pm$0.94}} (\textbf{13.03{\scriptsize $\pm$0.16}}) \\
    \midrule 
    $add$   & 32.83{\scriptsize $\pm$8.86} (\textbf{16.04{\scriptsize $\pm$1.25}}) & 23.36{\scriptsize $\pm$1.57} (16.55{\scriptsize $\pm$0.82}) & 23.46{\scriptsize $\pm$2.04} (12.22{\scriptsize $\pm$0.44}) \\
    \midrule
    $cat$   & 32.84{\scriptsize $\pm$10.80} (15.97{\scriptsize $\pm$1.46}) & 21.78{\scriptsize $\pm$0.20} (16.58{\scriptsize $\pm$0.72}) & 24.14{\scriptsize $\pm$2.31} (12.27{\scriptsize $\pm$0.52}) \\
    \midrule 
    $ncat$  & 29.58{\scriptsize $\pm$9.29} (13.37{\scriptsize $\pm$2.87}) & 20.93{\scriptsize $\pm$0.80} (\textbf{16.87{\scriptsize $\pm$0.62}}) & 23.38{\scriptsize $\pm$1.43} (12.30{\scriptsize $\pm$0.16}) \\
    \midrule
    $none$  & 27.79{\scriptsize $\pm$10.70} (13.99{\scriptsize $\pm$2.82}) & 19.64{\scriptsize $\pm$1.08} (16.53{\scriptsize $\pm$0.83}) & 20.95{\scriptsize $\pm$2.32} (12.81{\scriptsize $\pm$2.32}) \\

    \bottomrule
    \end{tabular}}%
  \label{table_noise_label:}%
\end{wraptable}%
\textbf{Impacts of varying merge operators.}
To look into the utility of the merge operator, we consider multiple merge operators, including $mul$, $add$, $cat$, $ncat$ and $none$.
Among them, $mul$ is $o(\bm{z}, y)= \bm{z} \times \mathcal{E}(y)$,  $add$ is $o(\bm{z}, y)= \bm{z} + \mathcal{E}(y)$, $cat$ is $o(\bm{z}, y)= [\bm{z}, \mathcal{E}(y)]$, $n\-cat$ is $o(\bm{z}, y)= [\bm{z}, y]$ and $none$ is $o(\bm{z}, y)= \bm{z}$.
From Table~\ref{table_noise_label:}, we can observe that our proposed $mul$ significantly outperforms other competitors in terms of $G.acc$.
This suggests that $mul$ can better exploit the diversity constraint to make the generator generate more diverse synthetic data, thus effectively fine-tuning the global model.
Also, the visualization of the synthetic data in Appendix~\ref{Vis_syn_samp:} validates this statement.

\textbf{Necessity of each component for DFRD.}
We report the test performance of DFRD after divesting some modules and losses in Table~\ref{table_ablation:}.
Here, EMA indicates the exponential moving average copy of the generator on the server. 
We can evidently observe that removing the EMA generator leads to a significant drop in $G.acc$, which implies that it can generate effective synthetic data for the global model.
The reason why the EMA generator works is that it avoids catastrophic forgetting of the global model and ensures the stability of the global model trained in heterogeneous FL.
We display synthetic data in Appendix ~\ref{Vis_syn_samp:} that further corroborates the above claims.
\begin{wraptable}{r}{7.3cm}
  \centering
  \caption{Impact of each component in DFRD.}
  \resizebox{0.5\columnwidth}{!}{
    \begin{tabular}{cp{6.315em}|p{6.065em}|p{6em}}
    \toprule
          & \multicolumn{1}{c|}{SVHN} & \multicolumn{1}{c|}{CIFAR-10} & \multicolumn{1}{c}{CIFAR-100} \\
    \midrule
    baseline & \textbf{34.78{\scriptsize $\pm$9.19}} (\textbf{15.99{\scriptsize $\pm$1.53}}) & \textbf{25.57{\scriptsize $\pm$1.37}} (\textbf{16.74{\scriptsize $\pm$0.72}}) & \textbf{28.08{\scriptsize $\pm$0.94}} (\textbf{13.03{\scriptsize $\pm$0.16}}) \\
    \midrule
    -EMA  & 26.97{\scriptsize $\pm$12.28} (14.17{\scriptsize $\pm$3.09}) & 19.80{\scriptsize $\pm$2.25} (16.55{\scriptsize $\pm$0.57}) & 24.57{\scriptsize $\pm$0.93} (12.23{\scriptsize $\pm$0.07}) \\
    \midrule
    -$\mathcal{L}_{tran}$ & 29.30{\scriptsize $\pm$9.25} (14.24{\scriptsize $\pm$3.08}) & 22.97{\scriptsize $\pm$1.91} (16.33{\scriptsize $\pm$1.16}) & 27.28{\scriptsize $\pm$0.46} (12.79{\scriptsize $\pm$0.46}) \\
    \rowcolor{Gray}
    -$\mathcal{L}_{div}$ & 27.68{\scriptsize $\pm$9.75} (14.26{\scriptsize $\pm$3.00}) & 22.12{\scriptsize $\pm$1.03} (16.46{\scriptsize $\pm$1.15}) & 26.94{\scriptsize $\pm$1.60} (12.81{\scriptsize $\pm$0.41}) \\
    -$\mathcal{L}_{tran}$, -$\mathcal{L}_{div}$ & 20.32{\scriptsize $\pm$1.93} (13.65{\scriptsize $\pm$1.13}) & 21.97{\scriptsize $\pm$2.48} (16.52{\scriptsize $\pm$1.39}) & 25.50{\scriptsize $\pm$0.91} (12.30{\scriptsize $\pm$0.10}) \\
    
    \bottomrule
    \end{tabular}}
  \label{table_ablation:}
\end{wraptable}
Meanwhile, we perform the leave-one-out test to explore the contributions of $\mathcal{L}_{tran}$ and $\mathcal{L}_{div}$ to DFRD separately, and further report the test results of removing them simultaneously.
From Table~\ref{table_ablation:}, deleting either $\mathcal{L}_{tran}$ or $\mathcal{L}_{div}$ adversely affects the performance of DFRD.
In addition, their joint absence further exacerbates the degradation of $G.acc$.
This suggests that $\mathcal{L}_{tran}$ and $\mathcal{L}_{div}$ are vital for the training of the generator.
Interestingly, $\mathcal{L}_{div}$ benefits more to the global model than $\mathcal{L}_{tran}$. 
We speculate that the diversity of synthetic data is more desired by the global model under the premise of ensuring the fidelity of synthetic data by optimizing $\mathcal{L}_{fid}$.

\begin{wrapfigure}{r}{0cm}
  \centering
  \includegraphics[scale=0.2]{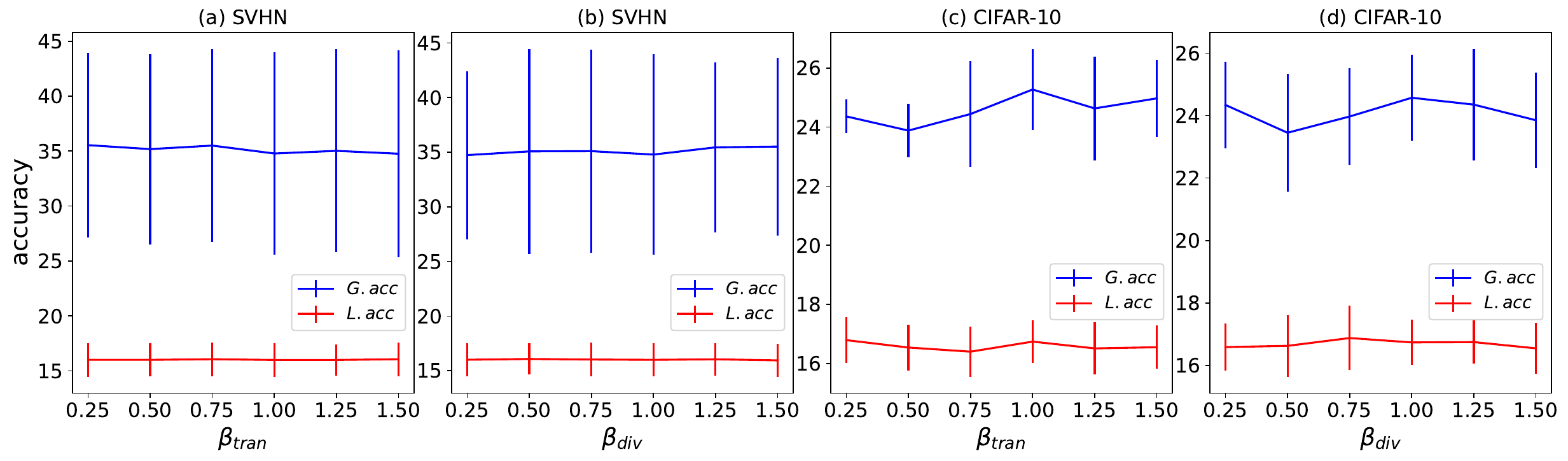}
  \caption{Test accuracy~(\%) with varying $\beta_{tran}$ and $\beta_{div}$.} 
  \label{pic_hyper_parameter:}
  \vspace*{-2ex}
\end{wrapfigure}

\textbf{Varying  $\beta_{tran}$ and $\beta_{div}$.}
We explore the impacts of $\beta_{tran}$ and $\beta_{div}$ on SVHN and CIFAR-10.
We select $\beta_{tran}$ and $\beta_{div}$ from $\{0.25, 0.50, 0.75, 1.00, 1.25, 1.50\}$.
From Fig.~\ref{pic_hyper_parameter:}, we can see that DFRD maintains stable test performance among all selections of $\beta_{tran}$ and $\beta_{div}$ over SVHN. At the same time, $G.acc$
fluctuates slightly with the increases of $\beta_{tran}$ and $\beta_{div}$ on CIFAR-10.
Besides, we observe that the worst $G.acc$ in Fig.~\ref{pic_hyper_parameter:} outperforms the baseline with the best $G.acc$ in Table~\ref{table_com_rho:}.
The above results indicate that DFRD is not sensitive to choices of $\beta_{tran}$ and $\beta_{div}$ over a wide range.

\begin{wrapfigure}{r}{0cm}
  \centering
  \includegraphics[scale=0.25]{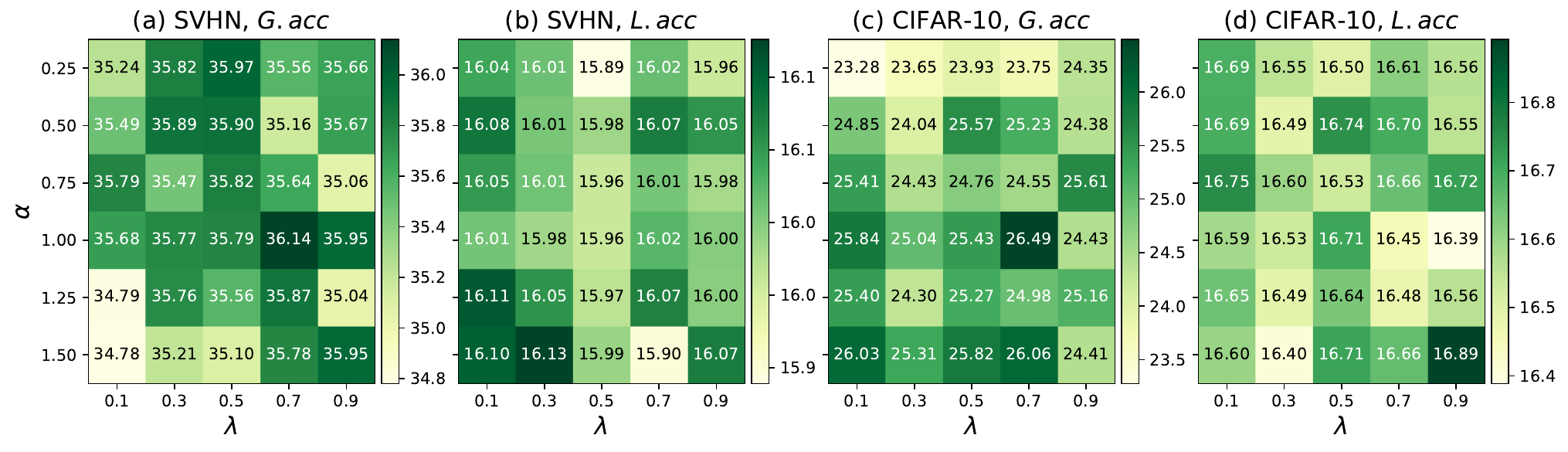}
  \caption{Test accuracy~(\%) with varying ($\alpha$ ,$\lambda$). }
  \label{pic_alpha_beta:} 
  \vspace*{-3ex}
\end{wrapfigure}
\textbf{Varying $\alpha$ and $\lambda$.}
In order to delve into the effect of the EMA generator on DFRD in more details, we perform grid testing on the choices of control parameters $\alpha$ and $\lambda$ over SVHN and CIFAR-10. 
We set $\alpha \in \{0.25, 0.50, 0.75, 1.00, 1.25, 1.50\}$ and $\lambda \in \{0.1, 0.3, 0.5, 0.7, 0.9\}$.
It can be observed from Fig.~\ref{pic_alpha_beta:} that high global test accuracies on SVHN are mainly located in the region of $\alpha<1.25$ and $\lambda>0.5$, while on CIFAR-10 they are mainly located in the region of $\alpha>0.25$  and $\lambda<0.9$. 
According to the above results, 
we deem that the appropriate $\alpha$ and $\lambda$ in a specific task is essential for the utility of the EMA generator.
Notably, high local test accuracies mainly sit in regions that are complementary to those of high global test accuracies, suggesting that pursuing high $G.acc$ and $L.acc$ simultaneously seems to be a dilemma.
How to ensure high $G.acc$ and $L.acc$ simultaneously in the field of FL is an attractive topic that is taken as our future work.

\section{Conclusion} 
In this paper, we propose a new FL method called DFRD, which aims to learn a robust global model in the data-heterogeneous and model-heterogeneous scenarios with the aid of DFKD.
To ensure the utility, DFRD considers a conditional generator and thoroughly studies its training in terms of \textit{fidelity}, \textit{transferability} and \textit{diversity}.
Additionally, DFRD maintains an EMA generator to augment  the global model.
Furthermore, we propose dynamic weighting and label sampling to accurately extract the knowledge of local models.
At last, we conduct extensive experiments to verify the superiority of DFRD. 
Due to space constraints, we discuss in detail the \textbf{limitations} and \textbf{broader impacts} of our work in Appendixes~\ref{app_discussion:} and~\ref{Broader_Impacts:}, respectively.


\section{Acknowledgments}
\label{sec:acknowledge}
This work has been supported by the National Natural Science Foundation of China under Grant No.U1911203, 
and the National Natural Science Foundation of China under Grant No.62377012.

\clearpage
\appendix

\section*{Appendix}

\section{Related Work}
\label{Related_work:}

\textbf{Heterogeneous Federated Learning.}
Federated Learning~(FL) has emerged as a de facto machine learning area and received rapidly increasing research interests from the community.
FedAvg, the classic distributed learning framework for FL, is first proposed by McMahan et al.~\cite{McMahan2017Communication}. 
Although FedAvg provides a practical and simple solution for aggregation, it still suffers from performance degradation or even fails to run due to the vast heterogeneity among real-world clients~\cite{Li2022Federated, Cho2022Heterogeneous}.

On the one hand, the distribution of data among clients may be non-IID~(identical and independently distributed), resulting in \textbf{data heterogeneity}~\cite{Zhao2018Federated, Fallah2020Personalized, Kairouz2021Advances, Li2021Fedbn, Li2022Federated}.
To ameliorate this issue, a myriad of modifications for FedAvg have been proposed~\cite{Li2020Federated, Karimireddy2020Scaffold, Acar2021Federated, Kim2022Multi, Mendieta2022Local, Lee2022Preservation, Zhang2022Federated, Tang2022Virtual, Luo2023GradMA, Li2021Model}. 
For example, 
FedProx~\cite{Li2020Federated} constrains local updates 
by adding a proximal term to the local objectives. 
Scaffold~\cite{Karimireddy2020Scaffold} uses 
control variate
to augment the local updates.
FedDyn~\cite{Acar2021Federated} dynamically regularizes the objectives of 
clients
to align global and local objectives.
Moon~\cite{Li2021Model} corrects the local training by conducting contrastive learning in model-level. 
GradMA~\cite{Luo2023GradMA} takes inspiration from continual learning to simultaneously rectify the server-side and client-side
update directions.
FedMLB~\cite{Kim2022Multi} architecturally regularizes the local objectives via online knowledge distillation. 

On the other hand, due to different hardware and computing resources~\cite{Ignatov2018Ai, Hong2022Efficient}, clients require to customize model capacities, which poses a more practical challenge: \textbf{model heterogeneity}.
Preceding methods, which are developed under the assumption that the local models and the global model have to share the same architecture, cannot work on heterogeneous models.
To address this problem, existing efforts fall broadly into two categories: knowledge distillation~(KD)-based methods~\cite{Li2019Fedmd, He2020Group, Lin2020Ensemble, Afonin2021Towards, Cho2022Heterogeneous, Fang2022Robust} and partial training~(PT)-based methods~\cite{Caldas2018Expanding, Horvath2021Fjord, Diao2020HeteroFL, Alam2022FedRolex}.
For \textbf{the former}, FedDF~\cite{Lin2020Ensemble} conducts KD based on logits averages to refine the network of the same size or several prototypes.
FedGKT~\cite{He2020Group} aggregates knowledge from clients and transfers it to a larger global model on the server based on KD.
FedMD~\cite{Li2019Fedmd} performs communication between clients by using KD, which requires the server to collect logits of the public data on each local model
and compute the average as the consensus for updates.
Fed-ET~\cite{Cho2022Heterogeneous} aims to train a large global model with smaller local models using a weighted consensus distillation scheme with diversity regularization.
RHFL~\cite{Fang2022Robust} aligns the logits outputs of the local models by KD based on public data.
However, most of the mentioned methods are data-dependent, that is, they often depend on access to public data which may not always be available in practice.
For \textbf{the latter}, PT-based methods adaptively extract a matching width-based slimmed-down sub-model from the global model as a local model according to each client's budget, thus averting the requirements for public data.
As with FedAvg, PT-based methods require the server to periodically communicate with the clients. 
Therefore, PT-based methods can be considered as an extension of FedAvg to model-heterogeneous scenarios.
Existing PT-based methods focus on how to extract width-based sub-models from the global model.
For example, HeteroFL~\cite{Diao2020HeteroFL} and FjORD~\cite{Horvath2021Fjord} propose a static extraction scheme in which sub-models are always extracted from a specified part of the global model.
Federated Dropout~\cite{Caldas2018Expanding} proposes a random extraction scheme, which randomly extracts sub-models from the global model.
FedRelox~\cite{Alam2022FedRolex} proposes a rolling sub-model extraction scheme, where the sub-model is extracted from the global model using a rolling window that advances in each communication round.
Going beyond the aforementioned methods, some works overcome model heterogeneity from other perspectives. 
For example, InclusiveFL~\cite{Liu2022No} and DepthFL~\cite{Kim2023DepthFL} consider depth-based sub-models with respect to the global model.
Also, FedProto~\cite{Tan2022Fedproto} and FedHeNN~\cite{Makhija2022Architecture} work on personalized FL with model heterogeneity.

\textbf{Data-Free Knowledge Distillation~(DFKD).} DFKD methods are promising, which transfer knowledge from the teacher model to another student model without any real data.
The generation of synthetic data that facilitates knowledge transfer is crucial in DFKD methods.
Existing DFKD methods can be broadly classified into non-adversarial and adversarial training methods.
Non-adversarial training methods
~\cite{Chen2019Data, Yoo2019Knowledge, Luo2020Large, Nayak2019Zero, Wang2021Data} exploit certain heuristics to search for synthetic data similar to the original training data.
For example, DAFL~\cite{Chen2019Data} and ZSKD~\cite{Nayak2019Zero} regard the predicted probabilities and confidences of the teacher model as heuristics.
Adversarial training methods
~\cite{Fang2021Contrastive, Yin2020Dreaming, Do2022Momentum} take advantage of adversarial learning to explore the distribution space of the original data.
They take the quality and/or diversity of the synthetic data as important objectives.
For example, CMI~\cite{Fang2021Contrastive} improves the diversity of synthetic data by leveraging inverse contrastive loss.
DeepInversion~\cite{Yin2020Dreaming} augments synthetic data by regularizing the distribution of Batch Norm to ensure visual interpretability.
Recently, some efforts~\cite{Do2022Momentum, Binici2022Preventing, Binici2022Robust, Patel2023Learning} have delved into the problem of catastrophic forgetting in adversarial settings.
Among them, MAD~\cite{Do2022Momentum} mitigates the generator's forgetting of previous knowledge by maintaining an exponential moving average copy of the generator. Compared to~\cite{Binici2022Preventing, Binici2022Robust, Patel2023Learning}, MAD is more memory efficient, adapts better to the student update, and is more stable for a continuous stream of synthetic samples. As such, in this paper we opt to maintain an exponential moving average copy of the generator to achieve robust model distillation.

\textbf{Federated Learning with DFKD.} The fact that DFKD does not need real data coincides with the requirement of FL to protect clients' private data.
A series of recent works~\cite{Zhu2021Data, Zhang2022Fine, Zhang2022DENSE, Heinbaugh2023Data} mitigate heterogeneity in FL by using DFKD.
For example, FedGEN~\cite{Zhu2021Data} uploads the last layer of the local models to the server and trains a conditional generator using DFKD to boost the local model update of each client.
FedFTG~\cite{Zhang2022Fine} leverages DFKD to train a conditional generator on the server based on local models, and then fine-tune the global model by KD in model-homogeneous FL.
DENSE~\cite{Zhang2022DENSE} aggregates knowledge from heterogeneous local models based on DFKD to train a global model for one-shot FL.
Most Recently, FEDCVAE-KD~\cite{Heinbaugh2023Data} reframes the local learning task using a conditional variational autoencoder, and leverages KD to compress the ensemble of client decoders into a global decoder. Then, it trains a global model using the synthetic data generated by the global decoder. Note that FEDCVAE-KD considers the one-shot FL.

\section{Datasets}
\label{dataset_app:}
In Table~\ref{Dataset_table:}, we provide information about the image size, the number of classes~(\#class), and the number of training/test samples~(\#train/\#test) of the datasets used in our experiment. 
Of note, we have used Resize$()$ in PyTorch to scale up and down the image size of FMNIST and FOOD101, respectively.
\begin{table}[htbp]
  \centering
  \caption{Statistics of the datasets used in our experiments.}
    \begin{tabular}{c|c|c|c|c}
    \toprule
    Dataset & Image size & \#class & \#train & \#test \\
    \midrule
    FMNIST & 1*32*32 & 10    & 60000 & 10000 \\
    \midrule
    SVHN  & \multirow{3}[6]{*}{3*32*32} & 10    & 50000 & 10000 \\
\cmidrule{1-1}\cmidrule{3-5}    CIFAR-10 &       & 10    & 50000 & 10000 \\
\cmidrule{1-1}\cmidrule{3-5}    CIFAR-100 &       & 100   & 73257 & 26032 \\
    \midrule
    Tiny-ImageNet & \multirow{2}[4]{*}{3*64*64} & 200   & 100000 & 10000 \\
\cmidrule{1-1}\cmidrule{3-5}    FOOD101 &       & 101   & 75750 & 25250 \\
    \bottomrule
    \end{tabular}%
  \label{Dataset_table:}%
\end{table}%

\section{Pseudocode}
\label{sec:pseudo}
In this section, we detail the pseudocode
of DFRD combined with a PT-based method in Algorithm~\ref{alg_all:}.
\begin{algorithm}[h]
  \caption{The pseudocode for DFRD combined with PT-based method.} 
  \label{alg_all:}
  \hspace*{0.02in} {\bf Input:}
  $T$: communication round; $N$: the number of clients; $S$: the number of sampled active clients per communication round; $B$: batch size.
  (\textbf{client side}) $I_c$: synchronization interval; $\eta_c$: learning rate for SGD.
  (\textbf{server side}) 
    $I$: iteration of the training procedure in server; ($I_g$, $I_d$): inner iterations of training the generator and the global model; ($\eta_g$, $b_1$, $b_2$): learning rate and momentums of Adam for the generator; $\eta_d$: learning rate of the global model; ($\beta_{tran}$, $\beta_{div}$): hyper-parameters in \textit{training generator}; ($\lambda$, $\alpha$): control parameters in \textit{robust model distillation}.
\begin{algorithmic}[1]
  \State Initial state $\bm{\theta}^{0}$, $\bm{w}^{0}$ and $\widetilde{\bm{w}}^{0}=\bm{0}$.
  \State Initial width capability $\{R_i\}_{i\in[N]}$. 
  \State Initial weighting counter $\Phi=\{\tau_{i, y}\}_{i\in [N], y \in [C]}\in \mathbbm{R}^{N\times C}$, label distribution $P=\{p(y)\}_{y \in [C]}\in \mathbbm{R}^{C}$ and label counter $LC=\{0\}_{i\in [N], y \in [C]}\in \mathbbm{R}^{N\times C}$.
  \For{$t=0,\cdots, T-1$}
    \State \textbf{On Server:}
    \State Sample a subset $\mathcal{S}_t$ with $S$ active clients.
    \State Transmit $\bm{\theta}_{i}^{t}$~(with $R_i$) extracted from $\bm{\theta}^{t}$ using sub-model extraction scheme~(e.g., \textit{random}~\cite{Caldas2018Expanding}, \textit{static}~\cite{Horvath2021Fjord, Diao2020HeteroFL} or \textit{rolling}~\cite{Alam2022FedRolex}) to client $i \in \mathcal{S}_t$.
    \State \textbf{On Clients:}
    \For{$i \in \mathcal{S}_t$ parallel}
        \State $\bm{\theta}_{i}^{t+1}, L_i=$ Client\_Update($\bm{\theta}_{i}^{t}$, $\eta_c$, $I_c$, $B$), \# see Algorithm~\ref{client_update:}
        \State Send $\bm{\theta}_{i}^{t+1}$ and $L_i$ to server.
    \EndFor
    \State \textbf{end for}
    \State \textbf{On Server:}
    \State 
    Aggregate $\{\bm{\theta}_{i}^{t+1}\}_{i\in \mathcal{S}_t}$ according to Eq.~(\ref{partial_avg:}) and yield $\bm{\theta}^{t+1}$.
    \For{$i \in \mathcal{S}_t$}
        \State $LC[i,:] \leftarrow L_i$.
    \EndFor
    \State \textbf{end for}
    \State Compute $\Phi$ and $P$ according to $LC$ by using Eq.~(\ref{Dyn_wei:}) and~(\ref{lab_samp:}), respectively.
    \State  $\bm{\theta}^{t+1}, \bm{w}^{t+1}=$ Server\_Update($\{\bm{\theta}_{i}^{t+1}\}_{i\in \mathcal{S}_t}$, $\bm{\theta}^{t+1}$,  $\bm{w}^{t}$,  $\widetilde{\bm{w}}^{t}$, $\Phi$, $P$,\\ \quad \quad \quad \quad \quad \quad \quad \quad \quad \quad \quad \quad \quad \quad  $I$, $I_g$, $I_d$, $\eta_g$, $b_1$, $b_2$, $\eta_d$, $\beta_{tran}$, $\beta_{div}$, $\alpha$, $B$). \# see Algorithm~\ref{server_update:}
    \State $\widetilde{\bm{w}}^{t+1} =  \lambda \cdot \widetilde{\bm{w}}^{t}+(1 - \lambda)\cdot \bm{w}^{t+1}$.
    \State $LC=\{0\}_{i\in [N], y \in [C]}$.
  \EndFor
  \State \textbf{end for}
\end{algorithmic}
\hspace*{0.02in} {\bf Output:} 
  $\bm{\theta}^T$ 
\end{algorithm}

\begin{algorithm}[h]
  \caption{Client\_Update($\bm{\theta}$, $\eta$, $I$, $B$) \# Take client $i$ as an example}
  \label{client_update:}
\begin{algorithmic}[1]
    \State Set $\bm{\theta}_i = \bm{\theta}$ and $L_i=\{0\}_{y\in [C]} \in \mathbbm{R}^{C}$.
    \State Cache data $Cache=[]$.
    \For{$e = 0,1,\ldots, I-1$}
        \State Sample $\mathcal{B} = \{(\bm{x}_i^b, y_i^b)\}_{b=1}^B$ from $\{(\bm{X}_i, \bm{Y}_i)\}$.
        \State $\bm{\theta}_{i}\leftarrow\bm{\theta}_{i}- \frac{\eta}{B}\sum_{b\in[B]} \nabla \mathcal{L}_{i}(f_i(\bm{x}_i^b,\bm{\theta}_{i}), y_i^b )$, where $\mathcal{L}_{i}$ denotes the cross-entropy loss.
        \For{$(\bm{x}, y)$ in $\mathcal{B}^e$}
            \If{$(\bm{x}, y)$ not in $Cache$}
                \State $Cache\leftarrow Cache \cup \{(\bm{x}, y)\}$.
            \EndIf
            \State \textbf{end if}
        \EndFor
        \State \textbf{end for}
        \For {$(\bm{x}, y)$ in $Cache$}
            \State $L_i^y\leftarrow L_i^y + 1$, where $L_i^y$ denotes the value of the $y^{th}$ element in $L_i$.
        \EndFor
        \State \textbf{end for}
    \EndFor
    \State \textbf{end for}
    \State {\bfseries Output:} $\bm{\theta}_{i}$, $L_i$.
\end{algorithmic}
\end{algorithm}

\begin{algorithm}[h]
  \caption{Server\_Update($\{\bm{\theta}_{i}\}_{i\in \mathcal{S}_t}$, $\bm{\theta}$,  $\bm{w}$,  $\widetilde{\bm{w}}$, $\Theta$ , $P$, $I$, $I_g$, $I_d$, $\eta_g$,  $b_1$, $b_2$, $\eta_d$, $\beta_{tran}$, $\beta_{div}$, $\alpha^\star$, $B$)}
  \label{server_update:}
\begin{algorithmic}[1]
    
    \For{$e = 0,1,\ldots, I-1$}
        \State Sample a batch of $\{\bm{z}^b, y^b\}_{b=1}^B$, where $\bm{z}\sim \mathcal{N}(\bm{0}, \bm{I})$ and $y\sim p(y):=P[y]$. 
        \State Set $\bm{m}=0$ and $\bm{v}=0$.
        \For{$e_g = 0,1,\ldots, I_g-1$}
            \State Generate $\{\bm{s}^b\}_{b=1}^B$ with $\{\bm{z}^b, y^b\}_{b=1}^B$ and $G(\cdot)$,
            \State Compute generator loss $\mathcal{L}_{gen}=\mathcal{L}_{fid}+\beta_{tran}\cdot\mathcal{L}_{tran}+\beta_{div} \cdot\mathcal{L}_{div}$ \\
            \quad \quad \quad \quad \quad \quad \quad \quad \quad \quad \quad \quad \quad \quad \quad \quad \quad \quad \quad \quad \quad \quad \quad \quad \quad \quad by using $\Theta$ and Eq.~(\ref{L_fidelity:})-(\ref{L_div:}),
            \State $\bm{g} = \frac{1}{B}\sum_{b\in [B]}\nabla_{\bm{w}} \mathcal{L}_{gen}$, 
            \State $\bm{m} \leftarrow b_1 \cdot \bm{m} + (1-b_1)\cdot\bm{g}$, $\bm{v} \leftarrow b_2\cdot\bm{v} + (1-b_2)\cdot \bm{g}^2$,
            \State $\hat{\bm{m}}\leftarrow \bm{m}/ (1-b_1)$, $\hat{\bm{v}}\leftarrow \bm{v}/ (1-b_2)$,
            \State  $\bm{w}\leftarrow \bm{w}-\eta_g \hat{\bm{m}} / (\sqrt{\hat{\bm{v}}} + 10^{-8})$.
        \EndFor
        \State \textbf{end for}
        \For{$e_d = 0,1,\ldots, I_d-1$}
            \State Generate $\{\bm{s}^b\}_{b=1}^B$ with $\{\bm{z}^b, y^b\}_{b=1}^B$ and $G(\cdot)$.
            \If{$\widetilde{\bm{w}}\neq\bm{0}$}
                \State Sample a batch of $\{\widetilde{\bm{z}}^b, \widetilde{y}^b\}_{b=1}^B$, where $\widetilde{\bm{z}}\sim \mathcal{N}(\bm{0}, \bm{I})$ and $\widetilde{y}\sim p(\widetilde{y}):=P[\widetilde{y}]$,
                \State Generate $\{\widetilde{\bm{s}}^b\}_{b=1}^B$ with $\{\widetilde{\bm{z}}^b, \widetilde{y}^b\}_{b=1}^B$ and $\widetilde{G}(\cdot)$,
                \State Compute global model loss $\mathcal{L}_{md}$ by using $\Theta$ and Eq.~(\ref{L_kl:}) with $\alpha = \alpha^\star$,
                \State $\bm{\theta} \leftarrow \bm{\theta}-\frac{\eta_d}{B}\sum_{b\in [B]}\nabla_{\bm{\theta}} \mathcal{L}_{md}$. 
            \Else
                \State Compute global model loss $\mathcal{L}_{kl}$ by using $\Theta$ and Eq.~(\ref{L_kl:}) with $\alpha = 0$,
                \State $\bm{\theta} \leftarrow \bm{\theta}-\frac{\eta_d}{B}\sum_{b\in [B]}\nabla_{\bm{\theta}} \mathcal{L}_{kl}$. 
            \EndIf
            \State \textbf{end if}

        \EndFor
        \State \textbf{end for}
    \EndFor
    \State \textbf{end for}
    \State {\bfseries Output:} $\bm{\theta}$, $\bm{w}$.
\end{algorithmic}
\end{algorithm}

\section{Budget Distribution}
\label{Budget_Distribution_app:}
In our work, in order to simulate model-heterogeneous scenarios and investigate the effect of the different model heterogeneity distributions. 
We formulate exponentially distributed budgets for a given $N$: $R_i = [\frac{1}{2}]^{\min\{\sigma, \lfloor\frac{\rho \cdot i}{N}\rfloor\}} ( i \in [N])$, where $\sigma$ and $\rho$ are both positive integers.
Specifically, $\sigma$ controls the smallest capacity model, i.e., $\left[\frac{1}{2}\right]^\sigma$-width w.r.t. global model. 
Also, $\rho$ manipulates client budget distribution. The larger the $\rho$ value, the higher the proportion of the smallest capacity models. 
We now present client budget distributions under several different settings about parameters $\sigma$ and $\rho$.
For example, we fix $\sigma=4$, that is, the smallest capacity model is $\frac{1}{16}$-width w.r.t. global model. 
Assuming $10$ clients participate in federated learning, we have: For any $i \in [10]$,
\begin{align}
    R_i = [\frac{1}{2}]^{\min\{4, \lfloor\frac{\rho \cdot i}{10}\rfloor\}}.
\end{align}
\begin{itemize}
    \item If we set $\rho=5$, $\{R_i\}_{i\in[10]}=\{1, \frac{1}{2}, \frac{1}{2}, \frac{1}{4}, \frac{1}{4}, \frac{1}{8}, \frac{1}{8}, \frac{1}{16}, \frac{1}{16}, \frac{1}{16}, \frac{1}{16}\}$;
    \item If we set $\rho=10$, $\{R_i\}_{i\in[10]}=\{\frac{1}{2}, \frac{1}{4}, \frac{1}{8}, \frac{1}{16}, \frac{1}{16}, \frac{1}{16}, \frac{1}{16}, \frac{1}{16}, \frac{1}{16}, \frac{1}{16}\}$;
    \item If we set $\rho=40$, $\{R_i\}_{i\in[10]}=\{\frac{1}{16}, \frac{1}{16}, \frac{1}{16}, \frac{1}{16}, \frac{1}{16}, \frac{1}{16}, \frac{1}{16}, \frac{1}{16}, \frac{1}{16}, \frac{1}{16}\}$. 
\end{itemize}
Note that if $\rho\geq4N$, each client holds $[\frac{1}{2}]^\sigma$-width w.r.t. global model. More budget distributions can refer to the literature~\cite{ Hong2022Efficient, Diao2020HeteroFL, Alam2022FedRolex}.

\section{Complete Experimental Setup}
\label{Com_Exp_Setup:}
In this section, for the convenience of the reader, we review the experimental setup for all the implemented algorithms on top of FMNIST, SVHN, CIFAR-10, CIFAR-100, Tiny-ImageNet and FOOD101.

\textbf{Datasets.} In this paper, we evaluate different methods with six real-world image classification task-related datasets, namely Fashion-MNIST~\cite{xiao2017fashion}~(FMNIST in short), SVHN~\cite{Netzer2011Reading}, CIFAR-10, CIFAR-100~\cite{Krizhevsky2009Learning}, Tiny-imageNet\footnote{http://cs231n.stanford.edu/tiny-imagenet-200.zip} and Food101~\cite{Bossard2014Food}.
We detail the six datasets in Appendix~\ref{dataset_app:}.
To simulate data heterogeneity across clients, as in previous works~\cite{Luo2023GradMA, Yurochkin2019Bayesian, Wang2020Federated}, we use Dirichlet process $Dir(\omega)$ to partition the training set for each dataset, thereby allocating local training data for each client.  
It is worth noting that $\omega$ is the concentration parameter and smaller $\omega$ corresponds to stronger data heterogeneity.  

\textbf{Baselines.} 
We compare DFRD to FedFTG~\cite{Zhang2022Fine} and DENSE~\cite{Zhang2022DENSE}, which are the most relevant methods to our work.
To verify the superiority of DFRD, on the one hand, DFRD, FedFTG and DENSE are directly adopted on the server to transfer the knowledge of the local models to a randomly initialized global model. 
We call them collectively \textbf{data-free methods}.
On the other hand, they are utilized as \textbf{fine-tuning methods} to improve the global model's performance after 
computing weighted average per communication round. 
In this case, in each communication round, the preliminary global model is obtained using FedAvg~\cite{McMahan2017Communication} in FL with homogeneous models, whereas in FL with heterogeneous models, the PT-based methods, including HeteroFL~\cite{Diao2020HeteroFL}, Federated Dropout~\cite{Caldas2018Expanding}~(FedDp for short) and FedRolex~\cite{Alam2022FedRolex}, are employed to get the preliminary global model.
\footnote{As an example, DFRD+FedRelox indicates that DFRD is used as a fine-tuning method to improve the performance of FedRelox, and others are similar. See Tables~\ref{table_com_omega:} and~\ref{table_com_rho:} for details.}

\begin{figure*}[h]\captionsetup[subfigure]{font=scriptsize}
    \centering
    \begin{subfigure}{0.3\linewidth}
        \centering
        \includegraphics[width=1.0\linewidth]{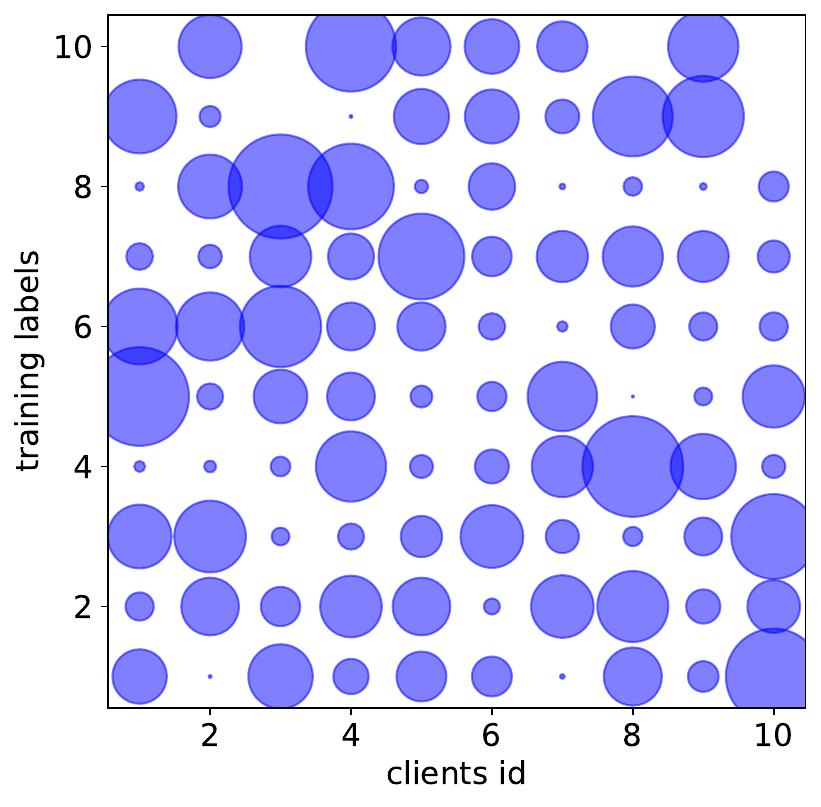}
        \caption{FMNIST, $\omega=1.0$}
        \label{chutian3}
    \end{subfigure}
    \centering
    \begin{subfigure}{0.3\linewidth}
        \centering
        \includegraphics[width=1.0\linewidth]{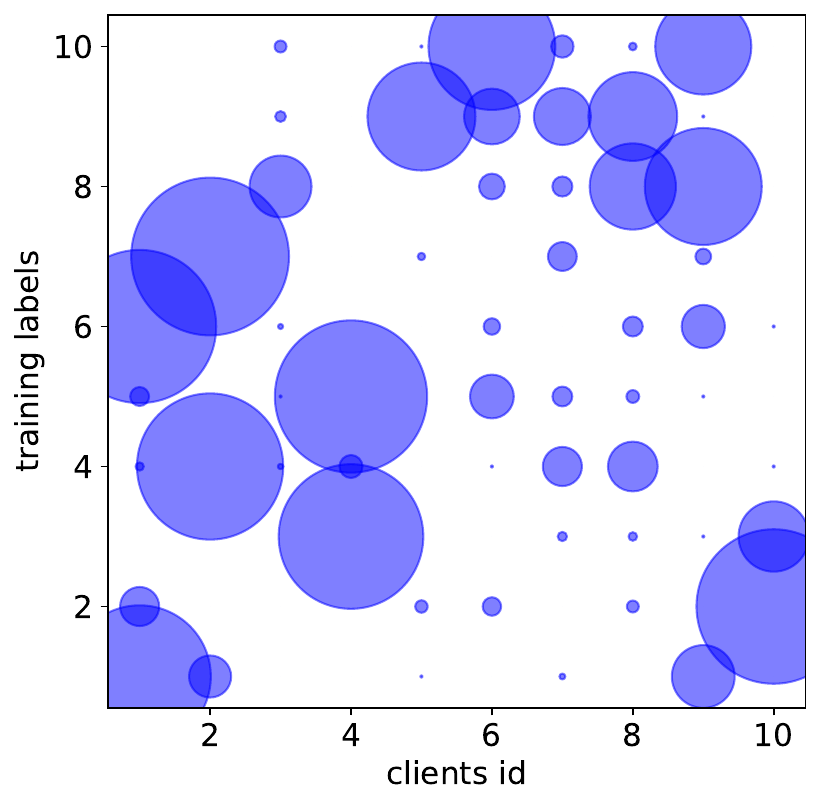}
        \caption{FMNIST, $\omega=0.1$}
        \label{chutian3}
    \end{subfigure} 
    \centering
    \begin{subfigure}{0.3\linewidth}
        \centering
        \includegraphics[width=1.0\linewidth]{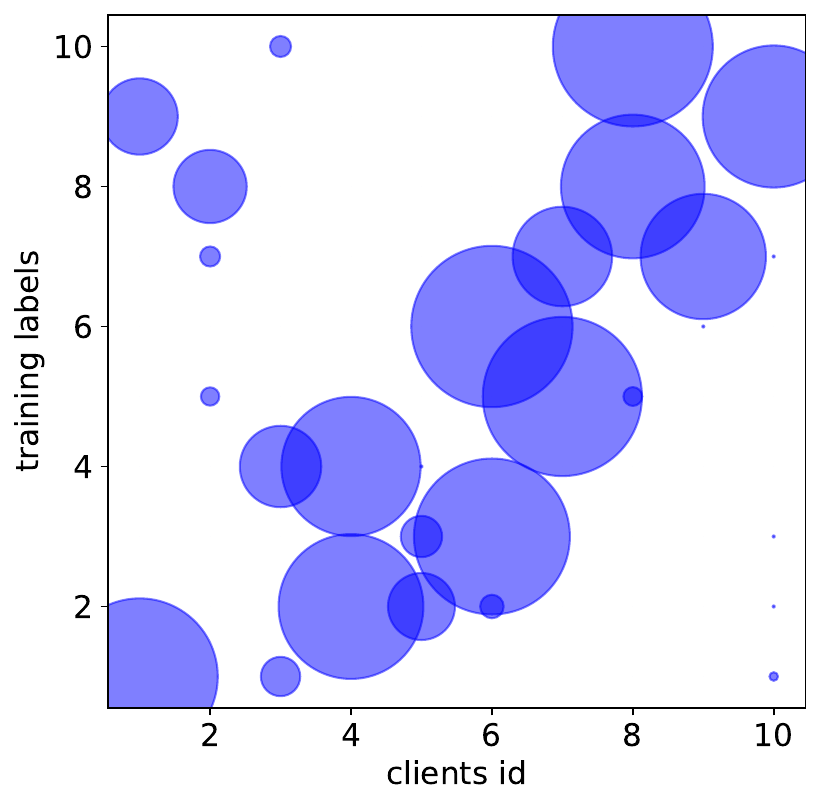}
        \caption{FMNIST, $\omega=0.01$}
        \label{chutian3}
    \end{subfigure} \\
    \centering
    \begin{subfigure}{0.3\linewidth}
        \centering
        \includegraphics[width=1.0\linewidth]{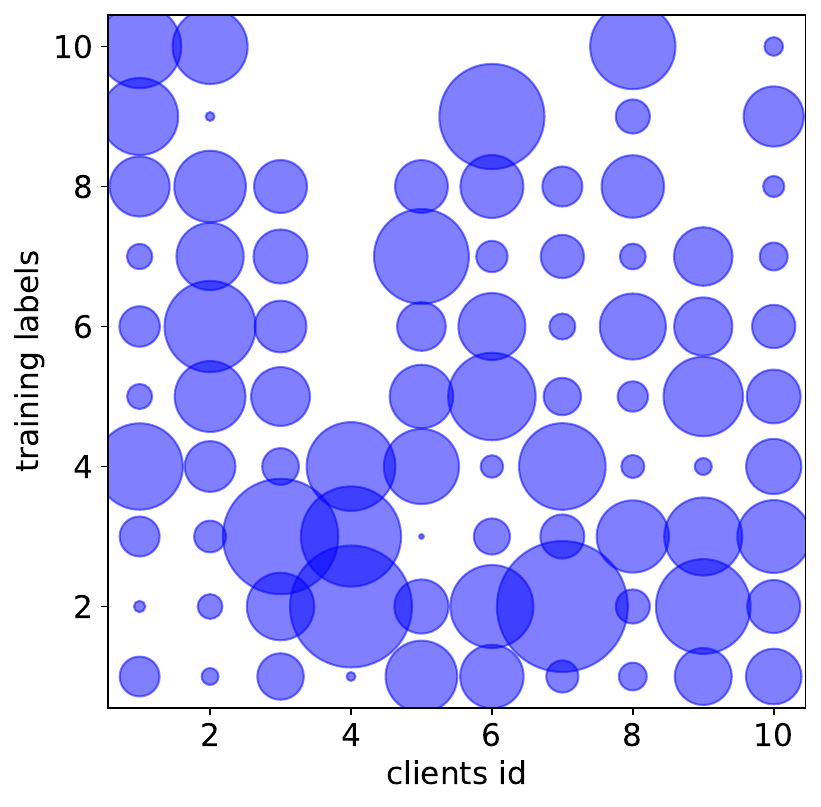}
        \caption{SVHN, $\omega=1.0$}
        \label{chutian3}
    \end{subfigure} 
    \centering
    \begin{subfigure}{0.3\linewidth}
        \centering
        \includegraphics[width=1.0\linewidth]{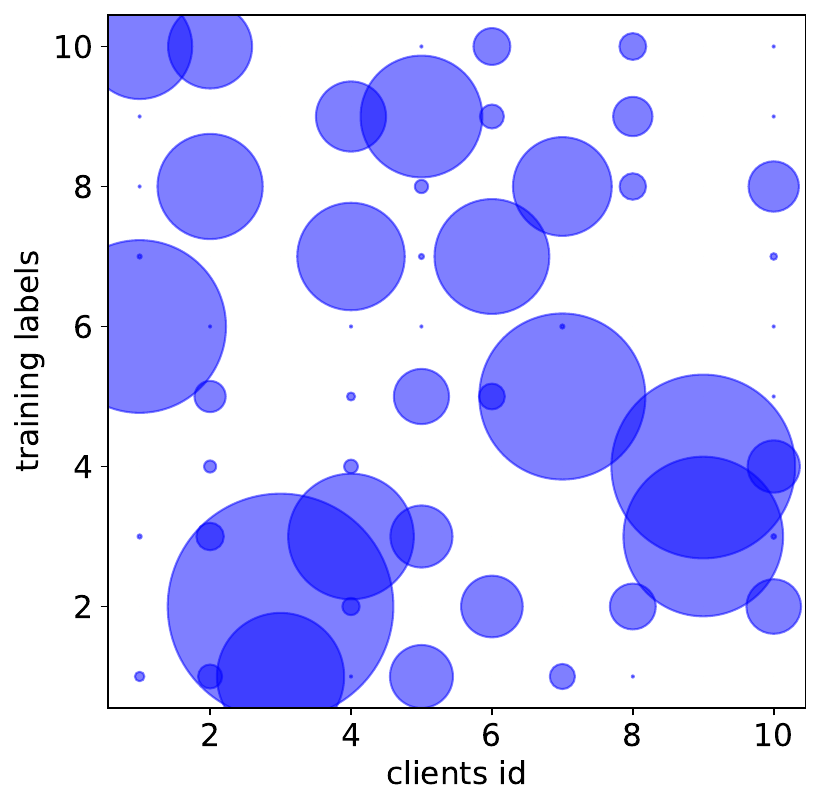}
        \caption{SVHN, $\omega=0.1$}
        \label{chutian3}
    \end{subfigure}
    \centering
    \begin{subfigure}{0.3\linewidth}
        \centering
        \includegraphics[width=1.0\linewidth]{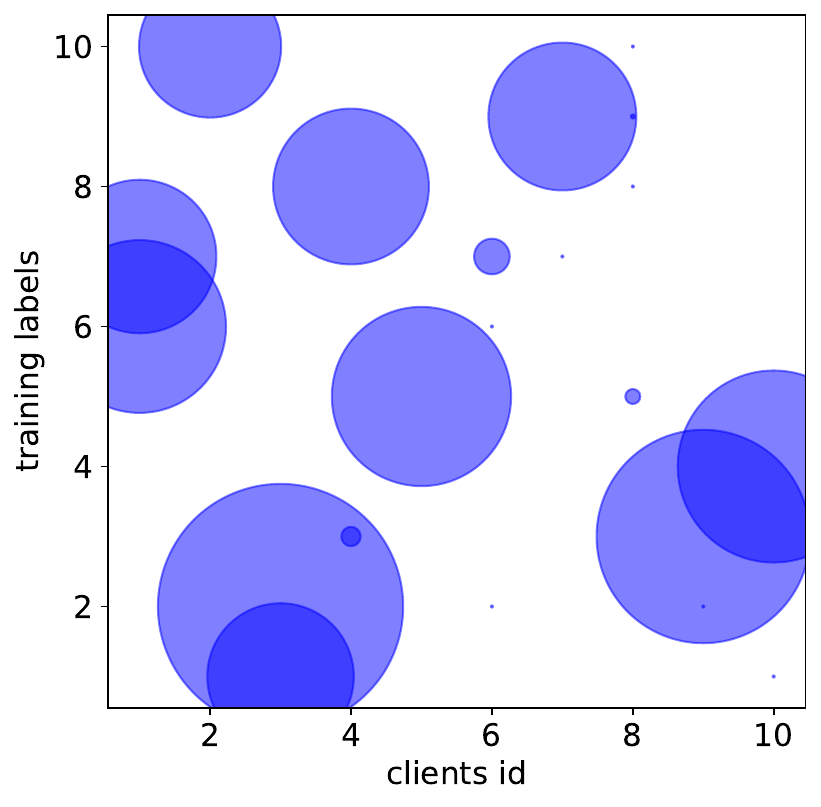}
        \caption{SVHN, $\omega=0.01$} 
        \label{chutian3}
    \end{subfigure} \\
    \centering
    \begin{subfigure}{0.3\linewidth}
        \centering
        \includegraphics[width=1.0\linewidth]{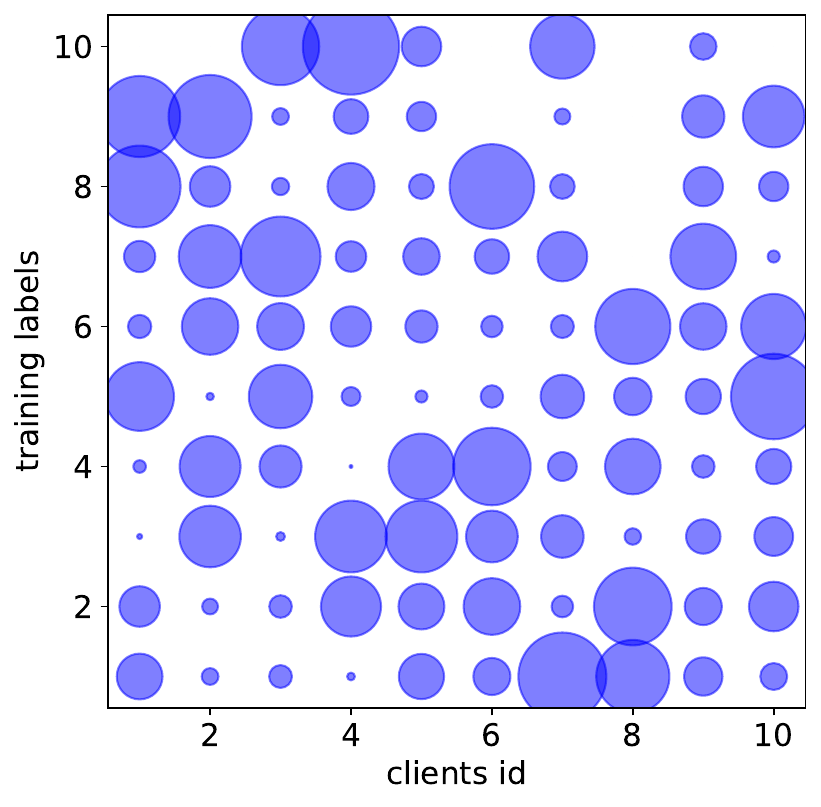}
        \caption{CIFAR-10, $\omega=1.0$}
        \label{chutian3}
    \end{subfigure}
    \centering
    \begin{subfigure}{0.3\linewidth}
        \centering
        \includegraphics[width=1.0\linewidth]{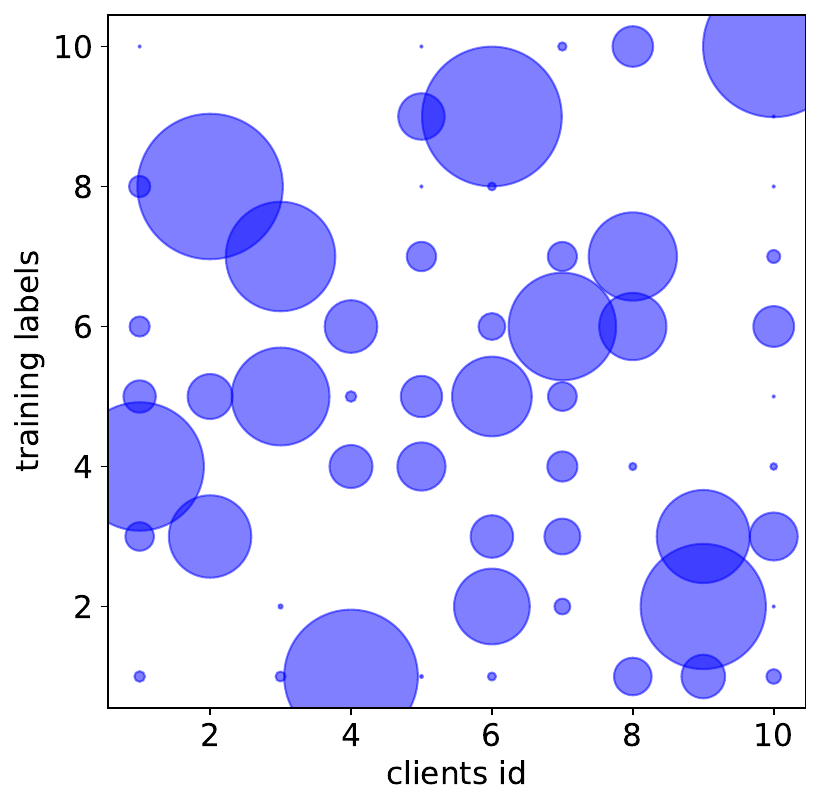}
        \caption{CIFAR-10, $\omega=0.1$}
        \label{chutian3}
    \end{subfigure}
    \centering
    \begin{subfigure}{0.3\linewidth}
        \centering
        \includegraphics[width=1.0\linewidth]{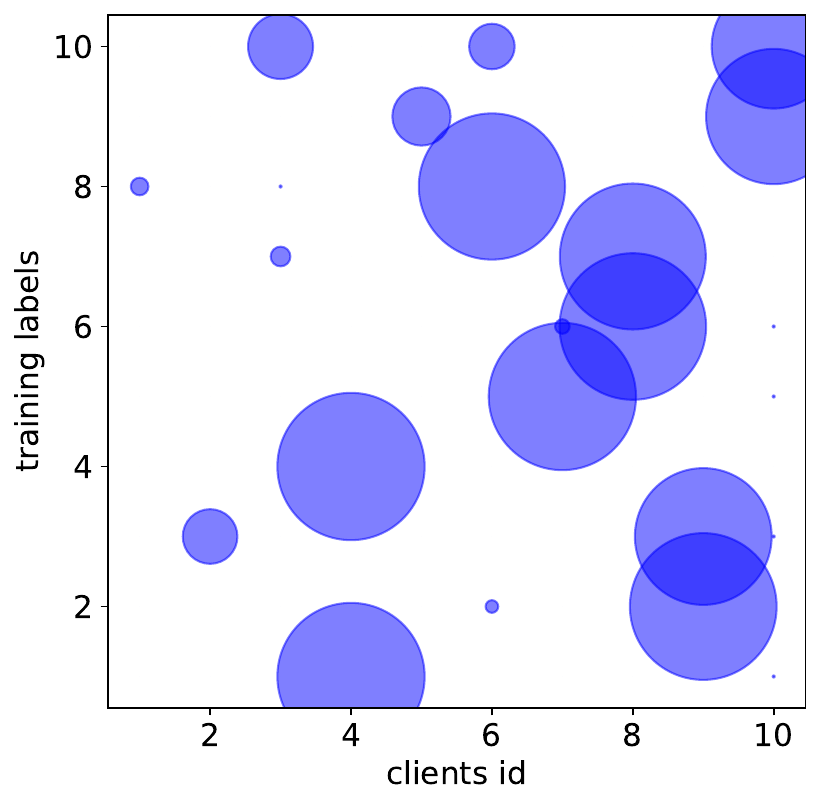}
        \caption{CIFAR-10, $\omega=0.01$}
        \label{chutian3}
    \end{subfigure} 
    \caption{Data heterogeneity among clients is visualized on three datasets~(FMNIST, SVHN, CIFAR-10), where the $x$-axis represents the clients id, the $y$-axis represents the class labels on the training set, and the size of scattered points represents the number of training samples with available labels for that client.} 
    \label{data_par_sum_appendix_1:}
\end{figure*}

\begin{figure*}[h]\captionsetup[subfigure]{font=scriptsize}
    \centering
    \begin{subfigure}{0.3\linewidth}
        \centering
        \includegraphics[width=1.0\linewidth]{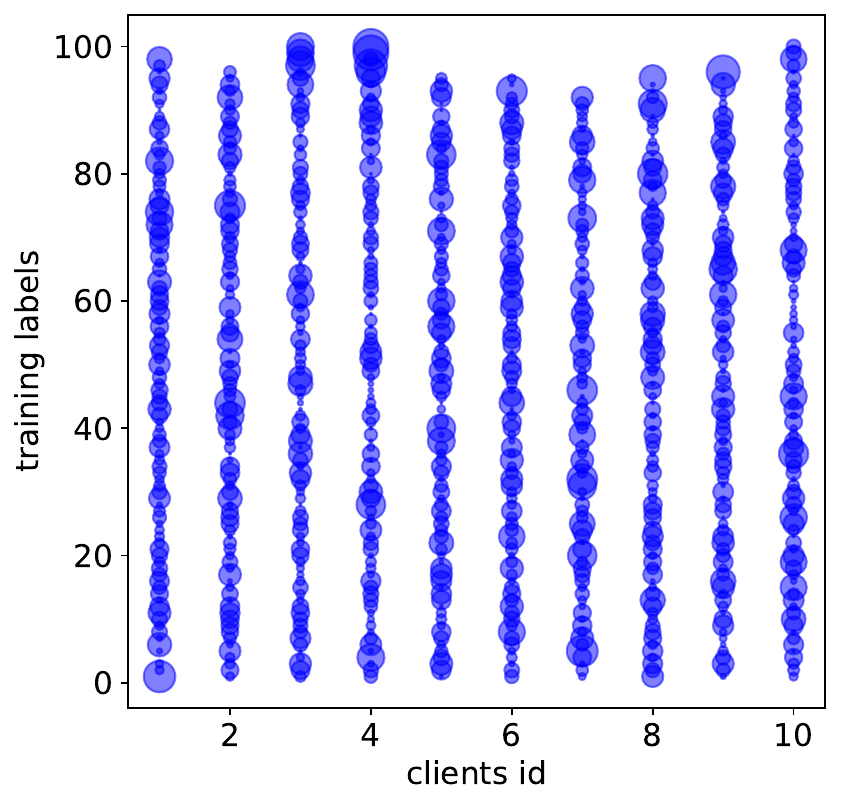}
        \caption{CIFAR-100, $\omega=1.0$}
        \label{chutian3}
    \end{subfigure}
    \centering
    \begin{subfigure}{0.3\linewidth}
        \centering
        \includegraphics[width=1.0\linewidth]{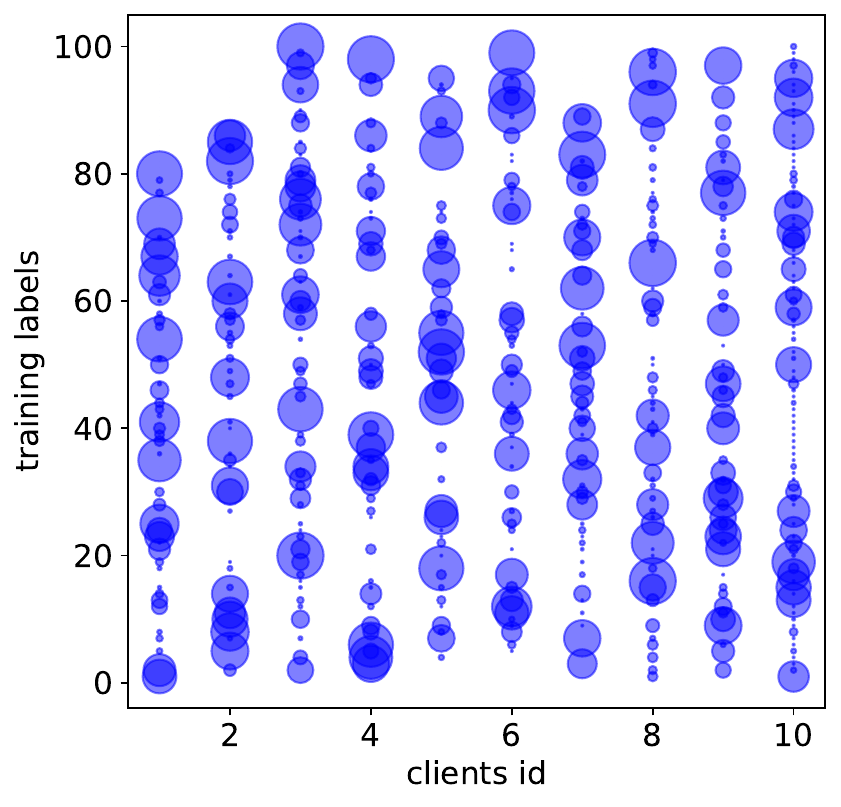}
        \caption{CIFAR-100, $\omega=0.1$}
        \label{chutian3}
    \end{subfigure}
    \centering
    \begin{subfigure}{0.3\linewidth}
        \centering
        \includegraphics[width=1.0\linewidth]{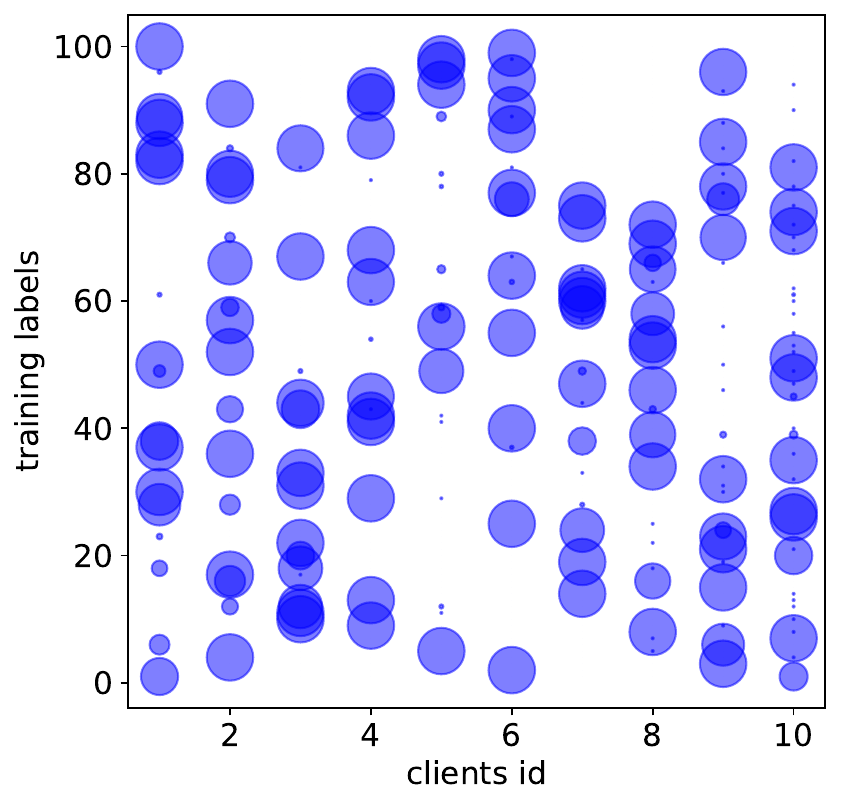}
        \caption{CIFAR-100, $\omega=0.01$}
        \label{chutian3}
    \end{subfigure} \\
    \centering
    \begin{subfigure}{0.3\linewidth}
        \centering
        \includegraphics[width=1.0\linewidth]{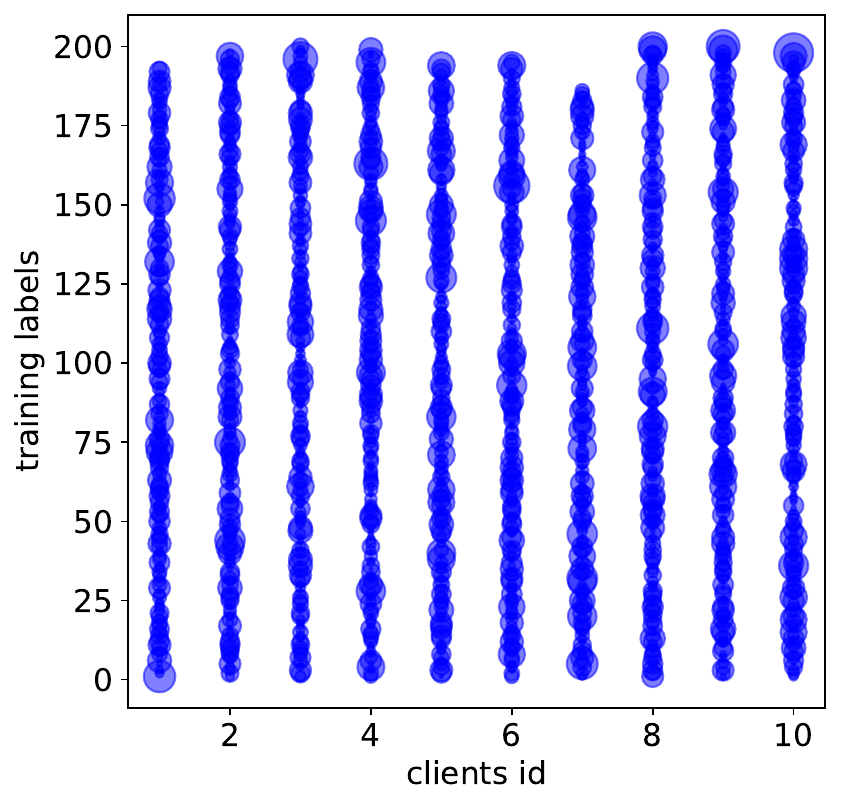}
        \caption{Tiny-ImageNet, $\omega=1.0$}
        \label{chutian3}
    \end{subfigure}
    \centering
    \begin{subfigure}{0.3\linewidth}
        \centering
        \includegraphics[width=1.0\linewidth]{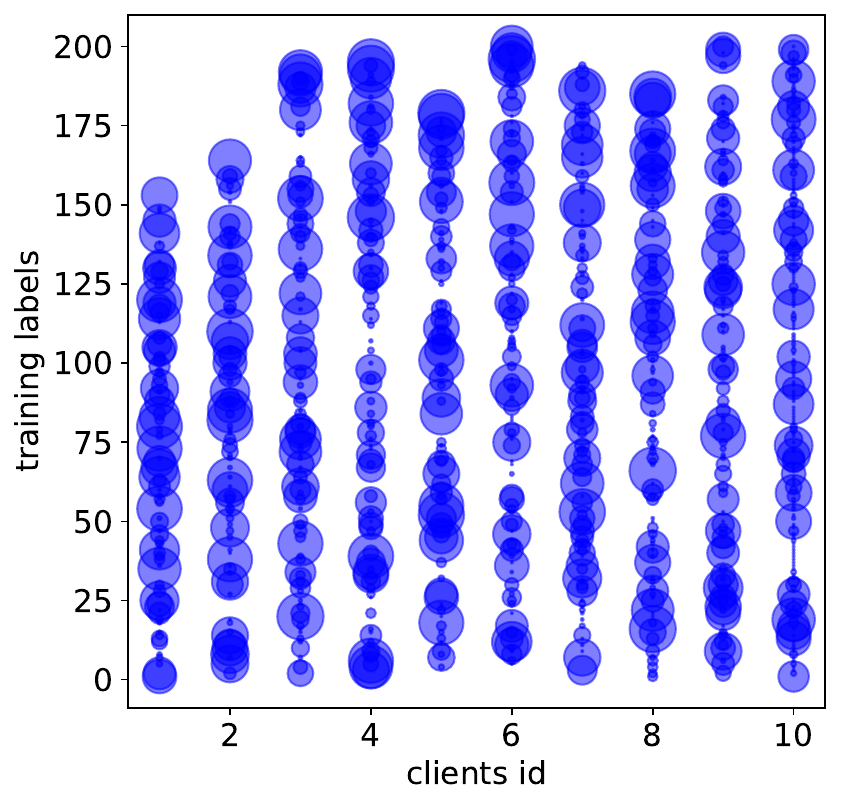}
        \caption{Tiny-ImageNet, $\omega=0.1$}
        \label{chutian3}
    \end{subfigure}
    \centering
    \begin{subfigure}{0.3\linewidth}
        \centering
        \includegraphics[width=1.0\linewidth]{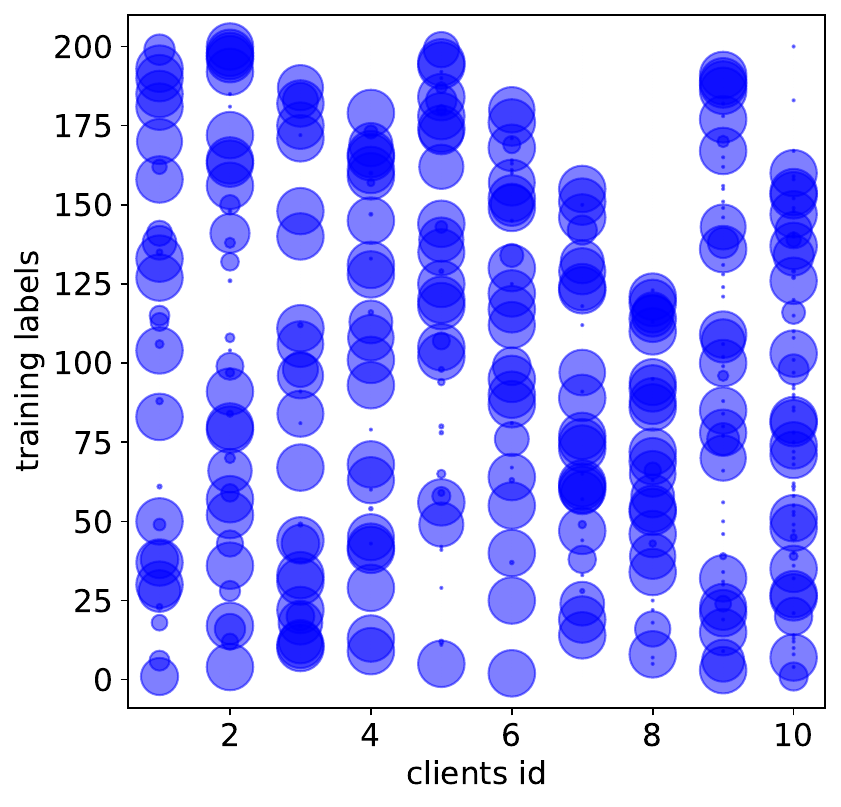}
        \caption{Tiny-ImageNet, $\omega=0.01$}
        \label{chutian3}
    \end{subfigure} \\
    \centering
    \begin{subfigure}{0.3\linewidth}
        \centering
        \includegraphics[width=1.0\linewidth]{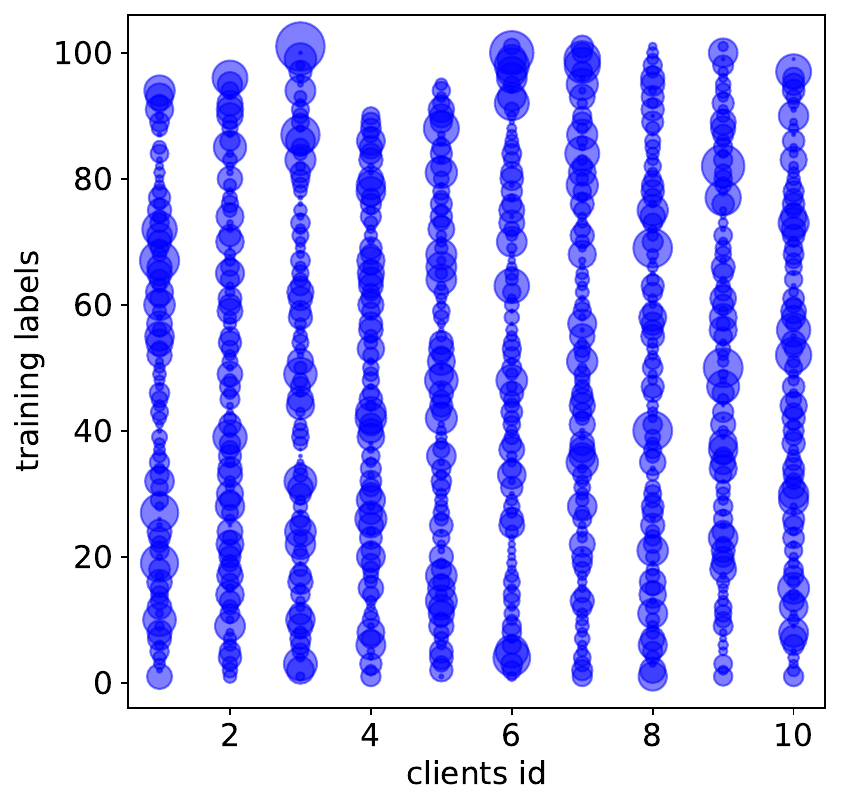}
        \caption{FOOD101, $\omega=1.0$}
        \label{chutian3}
    \end{subfigure}
    \centering
    \begin{subfigure}{0.3\linewidth}
        \centering
        \includegraphics[width=1.0\linewidth]{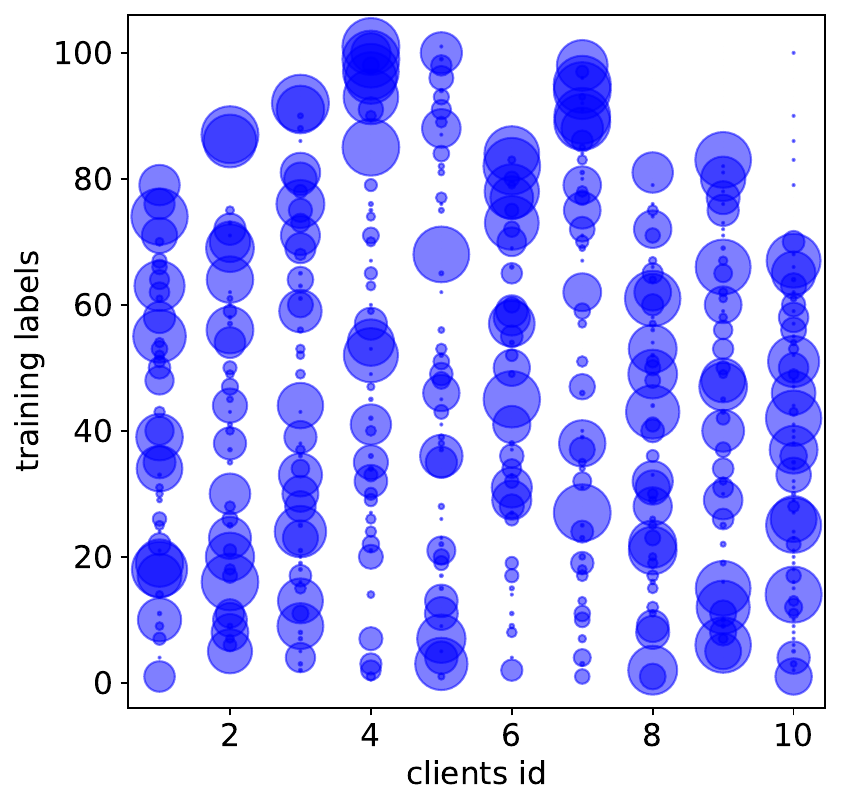}
        \caption{FOOD101, $\omega=0.1$}
        \label{chutian3}
    \end{subfigure}
    \centering
    \begin{subfigure}{0.3\linewidth}
        \centering
        \includegraphics[width=1.0\linewidth]{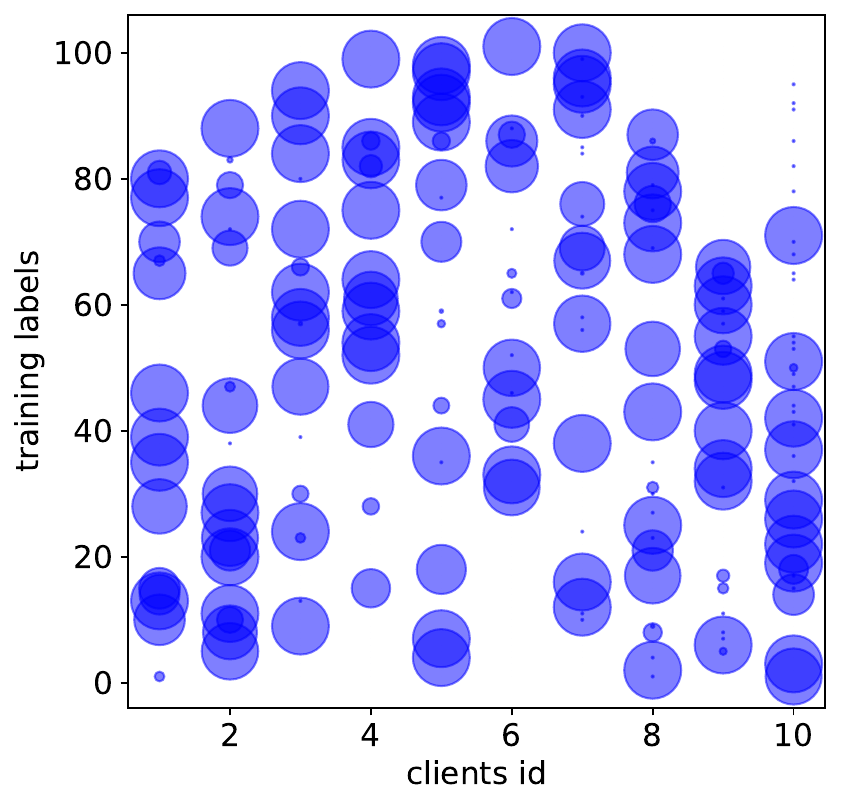}
        \caption{FOOD101, $\omega=0.01$}
        \label{chutian3}
    \end{subfigure}
    \caption{Data heterogeneity among clients is visualized on three datasets~(CIFAR-100, Tiny-ImageNet and FOOD101), where the $x$-axis represents the clients id, the $y$-axis represents the class labels on the training set, and the size of scattered points represents the number of training samples with available labels for that client.} 
    \label{data_par_sum_appendix_2:}
\end{figure*}

\textbf{Configurations.} 
Unless otherwise stated, all experiments are performed on a centralized network with $N=10$ active clients.
We set $\omega \in \{0.01, 0.1, 1.0\}$ to mimic different data heterogeneity scenarios. 
A visualization of the data partitions for the six datasets at varying $\omega$ values can be found in Fig.~\ref{data_par_sum_appendix_1:} and ~\ref{data_par_sum_appendix_2:}.
To simulate model-heterogeneous scenarios, we formulate exponentially distributed budgets for a given $N$: $R_i = [\frac{1}{2}]^{\min\{\sigma, \lfloor\frac{\rho \cdot i}{N}\rfloor\}} (i \in [N])$, where $\sigma$ and $\rho$ are both positive integers. 
We fix $\sigma=4$ and consider $\rho\in \{5, 10, 40\}$.
See Appendix~\ref{Budget_Distribution_app:} for more details.
Unless otherwise specified, we set $\beta_{tran}$ and $\beta_{div}$ both to $1$ in \textit{training generator}, while in \textit{robust model distillation}, we set $\lambda=0.5$ and $\alpha=0.5$.
For each client, we fix synchronization interval $I_c=20$ and select $\eta_c$ from $\{0.001, 0.01, 0.1\}$. 
For fairness, the popular SGD procedure is employed to perform local update steps for each client.
On the server, we take the values $10$, $5$ and $2$ for $I$, $I_g$ and $I_d$ respectively.
Also, SGD and Adam are applied to optimize the global model and generator, respectively.
The learning rate $\eta_d$ for SGD is searched over the range of $\{0.001, 0.01, 0.1\}$ and the best one is picked.
We set $\eta_g=0.0002$, $b_1=0.5$ and $b_2=0.999$ for Adam. 
For all update steps, we set batch size $B$ to $64$.

To verify the superiority of DFRD, we execute substantial comparative experiments in terms of data and model heterogeneity for FL, respectively.

Firstly, we investigate the performance of DFRD against that of baselines on FMNIST, SVHN, CIFAR-10 and CIFAR-100 w.r.t. different levels of data heterogeneity in FL with homogeneous models.
To be specific, For FMNIST, SVHN and CIFAR-10, a four-layer convolutional neural network with BatchNorm is implemented for each
client. 
We fix the total communication rounds to $100$, i.e., $T=100$. 
For CIFAR-100, each client implements a ResNet18~\cite{He2016Deep} architecture. 
We fix the total communication rounds to $500$, i.e., $T=500$.

Then, we look into the performance of DFRD against that of baselines on SVHN, CIFAR-10, Tiny-ImageNet and FOOD101 in terms of different levels of model heterogeneity distribution in FL. 
Concretely, for SVHN~(CIFAR-10), we set concentration parameter $\omega=0.1$. We equip the server with a four-layer convolutional neural network with BatchNorm as the global model, and extract suitable sub-models from the global model according to the width capability of each client. 
We fix the total communication rounds to $200$~($300$), i.e., $T=200$~($300$).
For Tiny-ImageNet~(FOOD101), we set the concentration parameter $\omega=1.0$. 
Meanwhile, a ResNet20~\cite{He2016Deep} architecture is equipped on the server as the global model, and appropriate sub-models are extracted from the global model according to the width capability of each client.
We fix the total communication rounds to $600$~($600$), i.e., $T=600$~($600$).
In addition, it is crucial to equip each dataset with a suitable generator on the server. 
The details of generator frameworks are shown in Tables~\ref{gen_FMNIST_tab:}-\ref{gen_tinyimagenet_food101_tab:}.
Moreover, for FMNIST, SVHN and CIFAR-10, we set the random noise dimension $d=64$. For CIFAR-100, Tiny-ImageNet and FOOD101, we set the random noise dimension $d=100$.
All baselines leverage the same setting as ours.

\begin{table}[htbp]
  \centering
  \caption{Generator for FMNIST}
    \begin{tabular}{c}
    \toprule
    $\bm{z}\in \mathbbm{R}^d \sim \mathcal{N}(\bm{0}, \bm{I})$, $y\in[C]$ \\
    \midrule
    $\bm{h}=o(\bm{z}, y)\rightarrow {d}$ \\
    \midrule
    Reshape($\bm{h}$)$\rightarrow{ d\times1\times1}$ \\
    \midrule
    Relu(ConvTransposed2d($d$, $512$)) $\rightarrow{B\times 512\times4\times4}$ \\
    \midrule
    Relu(BN(ConvTransposed2d($512$, $256$))) $\rightarrow{ 256\times8\times8}$ \\
    \midrule
    Relu(BN(ConvTransposed2d($256$, $128$))) $\rightarrow{128\times16\times16}$ \\
    \midrule
    Tanh(ConvTransposed2d($128$, $1$)) $\rightarrow{ 1\times32\times32}$ \\
    \bottomrule
    \end{tabular}%
  \label{gen_FMNIST_tab:}%
\end{table}%

\begin{table}[htbp]
  \centering
  \caption{Generator for SVHN, CIFAR-10 and CIFAR-100}
    \begin{tabular}{c}
    \toprule
    $\bm{z}\in \mathbbm{R}^d \sim \mathcal{N}(\bm{0}, \bm{I})$, $y\in[C]$ \\
    \midrule
    $\bm{h}=o(\bm{z}, y)\rightarrow {d}$ \\
    \midrule
    Reshape($\bm{h}$)$\rightarrow{ d\times1\times1}$ \\
    \midrule
    Relu(ConvTransposed2d($d$, $512$)) $\rightarrow{B\times 512\times4\times4}$ \\
    \midrule
    Relu(BN(ConvTransposed2d($512$, $256$))) $\rightarrow{ 256\times8\times8}$ \\
    \midrule
    Relu(BN(ConvTransposed2d($256$, $128$))) $\rightarrow{128\times16\times16}$ \\
    \midrule
    Tanh(ConvTransposed2d($128$, $3$)) $\rightarrow{ 3\times32\times32}$ \\
    \bottomrule
    \end{tabular}%
  \label{gen_SVHN_CIFAR10_100_tab:}%
\end{table}%

\begin{table}[htbp]
  \centering
  \caption{Generator for Tiny-ImageNet and FOOD101}
    \begin{tabular}{c}
    \toprule
    $\bm{z}\in \mathbbm{R}^d \sim \mathcal{N}(\bm{0}, \bm{I})$, $y\in[C]$ \\
    \midrule
    $\bm{h}=o(\bm{z}, y)\rightarrow {d}$ \\
    \midrule
    Reshape($\bm{h}$)$\rightarrow{ d\times1\times1}$ \\
    \midrule
    Relu(ConvTransposed2d($d$, $512$)) $\rightarrow{B\times 512\times4\times4}$ \\
    \midrule
    Relu(BN(ConvTransposed2d($512$, $256$))) $\rightarrow{ 256\times8\times8}$ \\
    \midrule
    Relu(BN(ConvTransposed2d($256$, $128$))) $\rightarrow{128\times16\times16}$ \\
    \midrule
    Relu(BN(ConvTransposed2d($128$, $64$))) $\rightarrow{64\times32\times32}$ \\
    \midrule
    Tanh(ConvTransposed2d($64$, $3$)) $\rightarrow{ 3\times64\times64}$ \\
    \bottomrule
    \end{tabular}%
  \label{gen_tinyimagenet_food101_tab:}%
\end{table}%

\textbf{Computing devices and platforms:}
\begin{itemize}
    \item OS: Ubuntu 20.04.3 LTS
    \item CPU: AMD EPYC 7763 64-Core Processor
    \item CPU Memory: 2 T
    \item GPU: NVIDIA Tesla A100 PCIe
    \item GPU Memory:  80GB
    \item Programming platform: Python 3.8.16
    \item Deep learning platform: PyTorch 1.12.1
\end{itemize}

\section{Dynamic Weighting and Label Sampling}
\label{dy_wei_La_sam:}
Here, we study how to determine $\tau_{i, y}$ and $p(y)$. We first consider $\tau_{i, y}$.
Existing works either assign the same weights to the logits outputs of different local models, i.e., $\tau_{i, y}=\frac{1}{|\mathcal{S}_t|}$~\cite{Lin2020Ensemble, Zhu2021Data, Zhang2022DENSE} ($\blacktriangle$), or assign weights to them based on the respective static data distribution of the clients, i.e., $\tau_{i, y} = n_i^{y} / n_{\mathcal{S}_t}^{y}$, where $n_{\mathcal{S}_t}^{y} = \sum_{ j\in [\mathcal{S}_t]}n_j^{y}$ and $n_i^{y}$ denotes the number of data points with label $y$ on the $i$-th client~\cite{Zhang2022Fine}($\blacktriangledown$).
For the former, since the importance of knowledge is different among local models
in FL with label distribution skewness~(i.e., data heterogeneity), the important knowledge cannot be correctly identified and utilized.
The latter, meanwhile, may not be feasible in practical FL, where on the one hand the label distribution on each client may be constantly changing~\cite{Singhal2021Federated, Zhang2021Real, Li2022Data, Zeng2023Hfedms}, and on the other hand, there may exist labels held by clients not involved in training under batch training.
Therefore, in FL with heterogeneous data, neither $\blacktriangle$ nor $\blacktriangledown$ can reasonably control the weight of the logits outputs among local models and yield a biased ensemble model~(i.e., $\sum_{i \in \mathcal{S}_t}\tau_{i, y} f_i(\bm{s}, \bm{\theta}_i)$ in Eq.~(\ref{L_fidelity:}), (\ref{L_tran:}) and (\ref{L_kl:})), thereby misleading the training of generator and global model.

To overcome the mentioned problems, we propose dynamic weighting, which only calculates the proportion of data involved in the local update in $\mathcal{S}_t$, i.e.,
\begin{align}
     \label{Dyn_wei:}
     \tau_{i, y} = n_{i,t}^{y} / n_{\mathcal{S}_t, t}^{y},
\end{align}
where $n_{\mathcal{S}_t, t}^{y} = \sum_{ j\in [\mathcal{S}_t]}n_{j,t}^{y}$ and  $n_{i,t}^{y}(\leq n_{i}^{y})$ denotes the number of data with label $y$ involved in training on client $i$  at round $t$~($\blacksquare$).
If $n_{i,t}^{y}=0$, it means the data with label $y$ held by client $i$ is not involved in training at $t$ round, or client $i$ has no data with label $y$.

We now give the label distribution $p(y)$.
Typically, $p(y)$ correlates with $\tau_{i, y}$.
For $\blacksquare$,
\begin{align}
     \label{lab_samp:}
     p(y)=n_{\mathcal{S}_t, t}^{y} / \sum_{y\in[C]}n_{\mathcal{S}_t, t}^{y}.
\end{align}
Similarly, $p(y)=\frac{1}{C}$ for $\blacktriangle$ and $p(y)=n_{\mathcal{S}_t}^{y} / \sum_{y\in[C]}n_{\mathcal{S}_t}^{y}$ for $\blacktriangledown$. 
\footnote{
Note that $\blacktriangle$, $\blacktriangledown$ and $\blacksquare$ denote the average, static and dynamic weighting schemes as well as the corresponding label sampling schemes, respectively.
}
Intuitively, $\tau_{i, y}$ and $p(y)$~(i.e. $\blacksquare$) obtained from the label distribution of clients participating in local training ensure that the generator accurately approximates the data distribution from the labels that they know best.


To verify the utility of $\blacktriangle$, $\blacktriangledown$ and $\blacksquare$, we conduct
experiments on SVHN, CIFAR-10 and CIFAR-100, and the test results are presented in Table~\ref{table_agg_strategy:}.
Note that the experimental setup here is consistent with Section~\ref{Ablation_study:} in the main paper and we do not repeat it.
From Table~\ref{table_agg_strategy:}, we can see that $\blacksquare$ significantly outperforms $\blacktriangle$ and $\blacktriangledown$ in terms of global test accuracy, which indicates that $\blacksquare$ is more effective in aggregating logits outputs of local models.
We attribute the merits of the proposed $\blacksquare$ to two things.
Firstly, in the local update phase, $\blacksquare$ only needs to compute the real data with labels involved in the training, which allows DFRD to effectively overcome the drift among local models, and thus enable a compliant global training space.
Secondly, the label distribution derived from $\blacksquare$ generates labels that participate in local updates on a probabilistic basis, so as to avoid misleading the training of the conditional generator $G(\cdot)$.

\begin{table}
  \centering
  \caption{Test accuracy~(\%) comparison among $\blacktriangle$, $\blacktriangledown$ and $\blacksquare$ over SVHN, CIFAR-10/100.}
  
    \begin{tabular}{cp{6.25em}|p{6.25em}|p{6.69em}}
    \toprule
    \multicolumn{1}{c}{W. S.} & \multicolumn{1}{c|}{SVHN} & \multicolumn{1}{c|}{CIFAR-10} & \multicolumn{1}{c}{CIFAR-100} \\
    \midrule
    $\blacktriangle$     & 30.56{\scriptsize $\pm$14.34} (15.93{\scriptsize $\pm$1.42}) & 21.61{\scriptsize $\pm$1.71} (15.77{\scriptsize $\pm$0.96}) & 24.98{\scriptsize $\pm$0.99} (12.19{\scriptsize $\pm$0.40}) \\
    \midrule
    $\blacktriangledown$     & 31.40{\scriptsize $\pm$12.72} (15.89{\scriptsize $\pm$1.35}) & 22.80{\scriptsize $\pm$1.34} (16.29{\scriptsize $\pm$1.43}) & 26.75{\scriptsize $\pm$2.00} (12.66{\scriptsize $\pm$0.56}) \\
    \midrule
    $\blacksquare$     & \textbf{34.78{\scriptsize $\pm$9.19}} (\textbf{15.99{\scriptsize $\pm$1.53}}) & \textbf{25.57{\scriptsize $\pm$1.37}} (\textbf{16.74{\scriptsize $\pm$0.72}}) & \textbf{28.08{\scriptsize $\pm$0.94}} (\textbf{13.03{\scriptsize $\pm$0.16}}) \\

    \bottomrule
    \end{tabular}
  \label{table_agg_strategy:}%
\end{table}%

\section{Impacts of the number of clients and stragglers}
\label{client_stragglers:}
In real-world FL scenarios with both data and model heterogeneity, the number of clients participating in the federation may be large~\cite{Nguyen2021Federated}.
Also, due to differences in available modes among clients, some clients may abort the training midway~(i.e., stragglers)~\cite{Luo2023GradMA}.
Therefore, in this section, we evaluate the test performance of FedRelox~(as a baseline) and FedRelox+DFRD with varying numbers of clients and stragglers on SVHN and CIFAR-10.
For the number of clients $N$, we select $N$ from $\{10, 20, 50, 100, 200\}$.
For stragglers, we consider its opposite, that is, the number of active clients participating in training per communication round.
Specifically, we fix $N=100$ and set the number of active clients $S\in \{5, 10, 50, 100\}$. 
Moreover, to ensure consistency, we set the width capability of each client to $\frac{1}{16}$ when $N$ is given.

Table~\ref{table_num_clients:} shows the test performances among different $N$.
We can observe that as $N$ increases, the global test accuracy of FedRolex+DFRD decreases uniformly on SVHN, while it increases and then decreases on CIFAR-10.
For the latter, this seems inconsistent with the observations in~\cite{McMahan2017Communication, Zhang2022DENSE}.
This is an interesting result. 
We conjecture that for a given task, increasing $N$ within a certain range, FedRolex+DFRD may enhance the global model, but the performance of the global model still degrades if a threshold is exceeded.
Besides, FedRolex+DFRD significantly outperforms FedRolex for each given $N$ in terms of global test accuracy, which improves by an average of $9.75\%$ on SVHN and $8.01\%$ on CIFAR-10.
This shows that DFRD can effectively augment the global model for different $N$ in FL with both data and model heterogeneity.

Table~\ref{table_num_active_clients:} reports the test results for different $S$.
A higher $S$ means more active clients upload model per communication round.
Note that due to the existence of stragglers, the number of rounds and times of training among clients may not be identical. 
For brevity, we only display the global test accuracy.
We can clearly see that the performance of FedRolex and FedRolex+DFRD improves uniformly with increasing $S$, where FedRolex+DFRD consistently dominates FedRolex in terms of global test accuracy.
This means that stragglers are detrimental to the global model.
As such, stragglers pose new challenges for data-free knowledge distillation, which we leave for future work.

\begin{table}[h]
  \centering
  \caption{Test accuracy~(\%) over different $N$ on SVHN and CIFAR-10.}
    \begin{tabular}{c|p{6.19em}p{6.94em}|p{6.125em}p{6.565em}}
    \toprule
    Dataset & \multicolumn{2}{c|}{SVHN} & \multicolumn{2}{c}{CIFAR-10} \\
    \midrule
    $N$     & \multicolumn{1}{c}{FedRolex} & FedRolex+DFRD & \multicolumn{1}{c}{FedRolex} & FedRolex+DFRD \\
    \midrule
    10    & 22.39{\scriptsize$\pm$15.28} (13.59{\scriptsize$\pm$2.11}) & \textbf{32.86{\scriptsize$\pm$15.54}} (\textbf{15.17{\scriptsize$\pm$0.66}}) & 14.37{\scriptsize$\pm$2.91} (17.11{\scriptsize$\pm$1.29}) & \textbf{19.86{\scriptsize$\pm$2.76}} (\textbf{17.77{\scriptsize$\pm$1.16}}) \\
    \midrule
    20    & 18.20{\scriptsize$\pm$1.43} (12.48{\scriptsize$\pm$0.47}) & \textbf{31.32{\scriptsize$\pm$10.16}} (\textbf{13.69{\scriptsize$\pm$1.38}}) & 13.60{\scriptsize$\pm$0.68} (16.30{\scriptsize$\pm$0.98}) & \textbf{20.20{\scriptsize$\pm$1.30}} (\textbf{17.11{\scriptsize$\pm$0.92}}) \\
    \midrule
    50    & 16.83{\scriptsize$\pm$1.02} (11.73{\scriptsize$\pm$0.45}) & \textbf{27.62{\scriptsize$\pm$3.41}} (\textbf{12.48{\scriptsize$\pm$0.90}}) & 12.75{\scriptsize$\pm$0.71} (16.50{\scriptsize$\pm$1.14}) & \textbf{21.14{\scriptsize$\pm$2.17}} (\textbf{16.53{\scriptsize$\pm$0.78}}) \\
    \midrule
    100   & 15.38{\scriptsize$\pm$1.22} (10.80{\scriptsize$\pm$0.14}) & \textbf{24.22{\scriptsize$\pm$4.75}} (\textbf{11.56{\scriptsize$\pm$0.53}}) & 12.25{\scriptsize$\pm$1.39} (16.26{\scriptsize$\pm$1.01}) & \textbf{23.68{\scriptsize$\pm$1.25}} (\textbf{16.88{\scriptsize$\pm$0.72}}) \\
    \midrule
    200   & 10.33{\scriptsize$\pm$0.17} (9.45{\scriptsize$\pm$1.59}) & \textbf{15.85{\scriptsize$\pm$2.64}} (\textbf{10.01{\scriptsize$\pm$0.21}}) & 10.14{\scriptsize$\pm$2.11} (14.65{\scriptsize$\pm$0.99}) & \textbf{18.27{\scriptsize$\pm$0.56}} (\textbf{15.44{\scriptsize$\pm$0.53}}) \\
    \bottomrule
    \end{tabular}%
  \label{table_num_clients:}%
\end{table}%

\begin{table}[h]
  \centering
  \caption{Test accuracy~(\%) over different $S$ per communication round on SVHN and CIFAR-10.}
    \begin{tabular}{c|p{6.19em}p{6.5em}|p{6.125em}p{6.69em}}
    \toprule
    Dataset & \multicolumn{2}{c|}{SVHN} & \multicolumn{2}{c}{CIFAR-10} \\
    \midrule
    $S$     & FedRolex & FedRolex+DFRD & FedRolex & FedRolex+DFRD \\
    \midrule
    5     & 13.86{\scriptsize $\pm$2.06}  & \textbf{17.58{\scriptsize $\pm$0.47}} & 10.22{\scriptsize$\pm$0.46} & \textbf{18.59{\scriptsize$\pm$0.44}} \\
    \midrule
    10    & 14.64{\scriptsize$\pm$1.17}  & \textbf{19.29{\scriptsize$\pm$0.17}} & 11.67{\scriptsize$\pm$1.34} & \textbf{21.22{\scriptsize$\pm$0.08}} \\
    \midrule
    50    & 15.17{\scriptsize$\pm$1.63} & \textbf{22.85{\scriptsize$\pm$4.23}} & 12.73{\scriptsize$\pm$0.71} & \textbf{23.18{\scriptsize$\pm$0.31}} \\
    \midrule
    100   & 15.38{\scriptsize$\pm$1.22} & \textbf{24.22{\scriptsize$\pm$4.75}} & 12.79{\scriptsize$\pm$0.71} & \textbf{23.68{\scriptsize$\pm$1.25}} \\
    \bottomrule
    \end{tabular}%
  \label{table_num_active_clients:}%
  \vspace{-0.3cm}
\end{table}%

\section{Full learning curves.}
\label{full_learn_curve:}
\begin{figure*}[h]\captionsetup[subfigure]{font=scriptsize}
    \centering
    \begin{subfigure}{0.45\linewidth}
        \centering
        \includegraphics[width=1.0\linewidth]{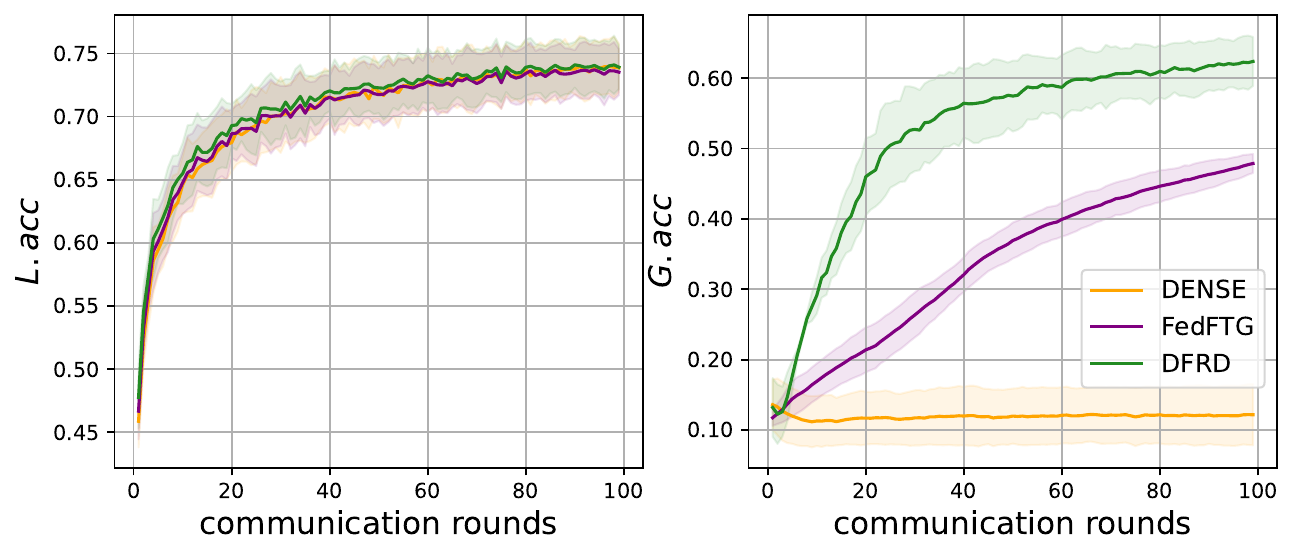}
        \caption{FMNIST, $\omega=1.0$}
        \label{chutian3}
    \end{subfigure}
    \centering
    \begin{subfigure}{0.45\linewidth}
        \centering
        \includegraphics[width=1.0\linewidth]{data_free_FMNIST_0.1_comm_round.pdf}
        \caption{FMNIST, $\omega=0.1$}
        \label{chutian3}
    \end{subfigure} \\
    \centering
    \begin{subfigure}{0.45\linewidth}
        \centering
        \includegraphics[width=1.0\linewidth]{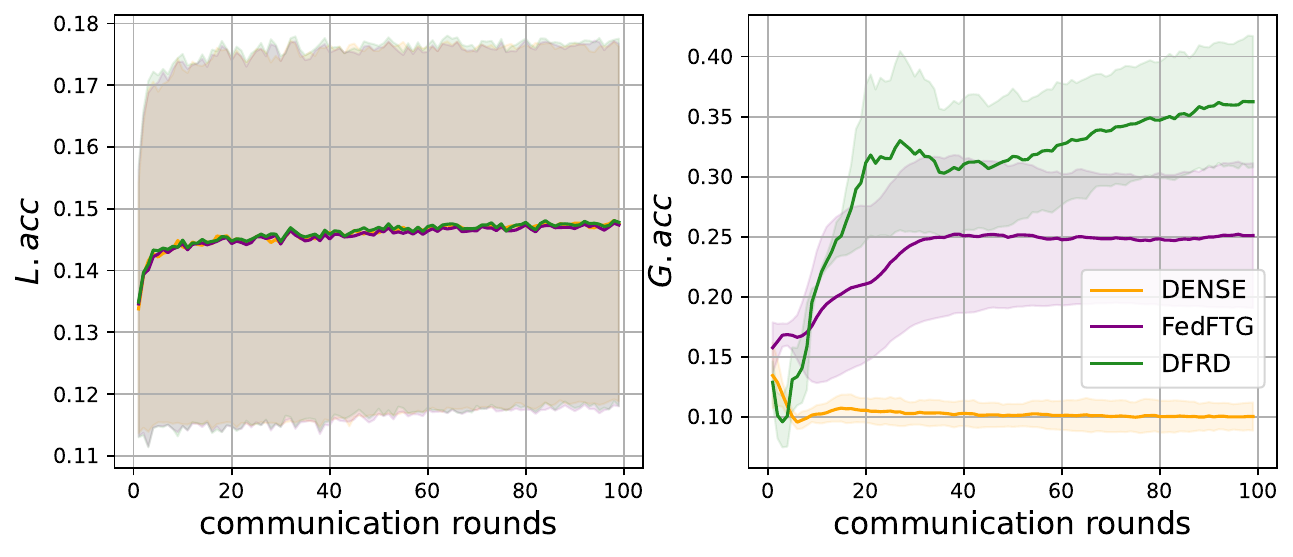}
        \caption{FMNIST, $\omega=0.01$}
        \label{chutian3}
    \end{subfigure}
    \centering
    \begin{subfigure}{0.45\linewidth}
        \centering
        \includegraphics[width=1.0\linewidth]{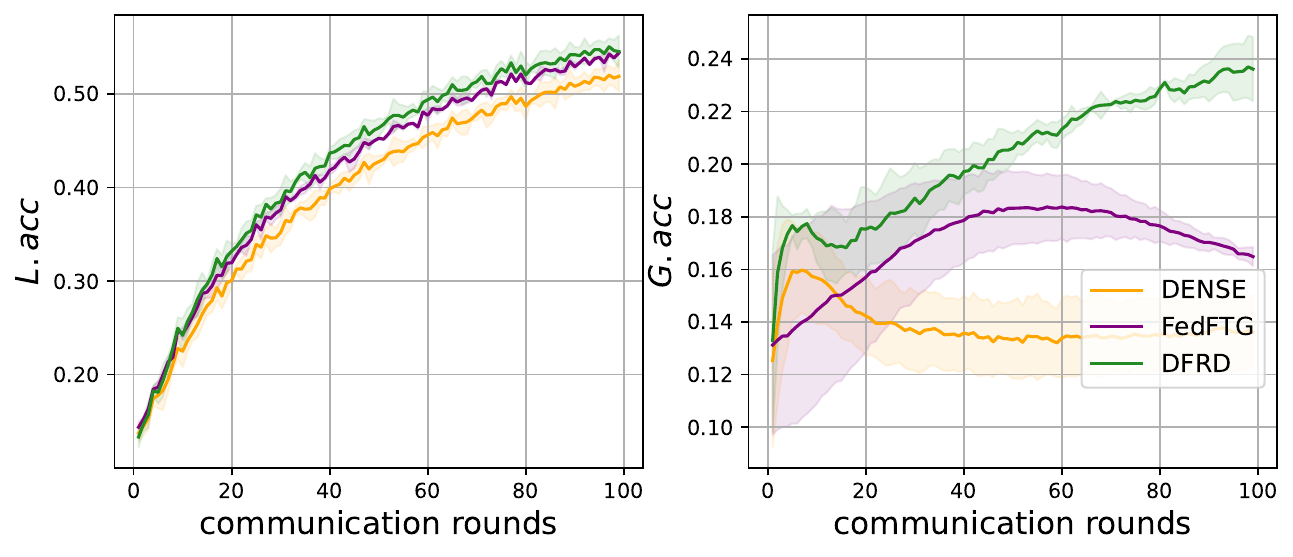}
        \caption{SVHN, $\omega=1.0$}
        \label{chutian3}
    \end{subfigure} \\
    \centering
    \begin{subfigure}{0.45\linewidth}
        \centering
        \includegraphics[width=1.0\linewidth]{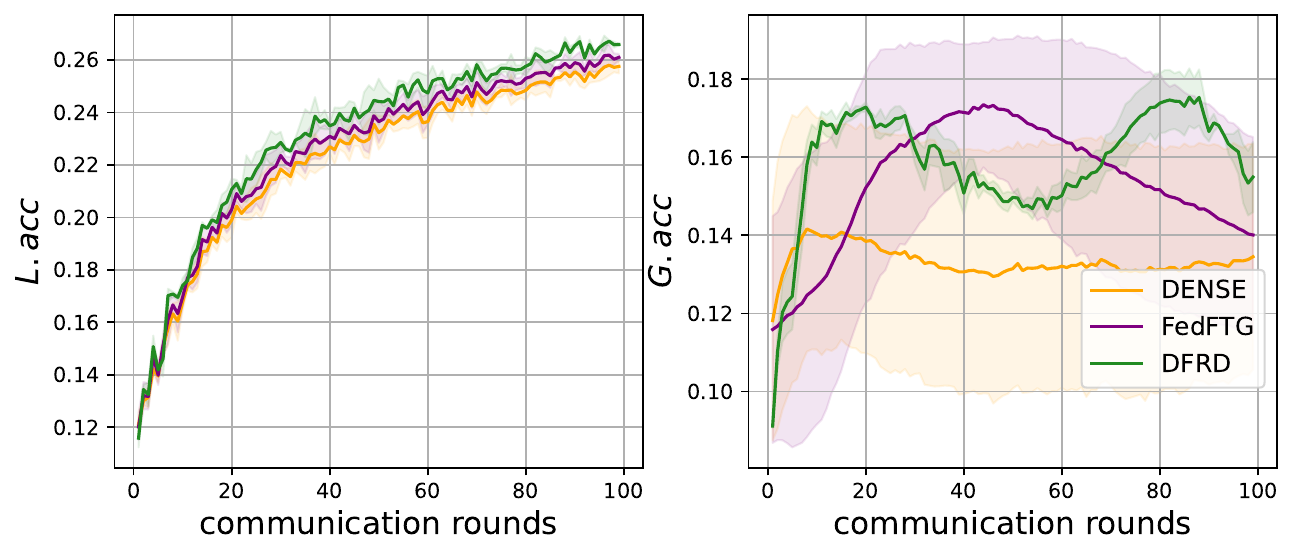}
        \caption{SVHN, $\omega=0.1$}
        \label{chutian3}
    \end{subfigure}
    \centering
    \begin{subfigure}{0.45\linewidth}
        \centering
        \includegraphics[width=1.0\linewidth]{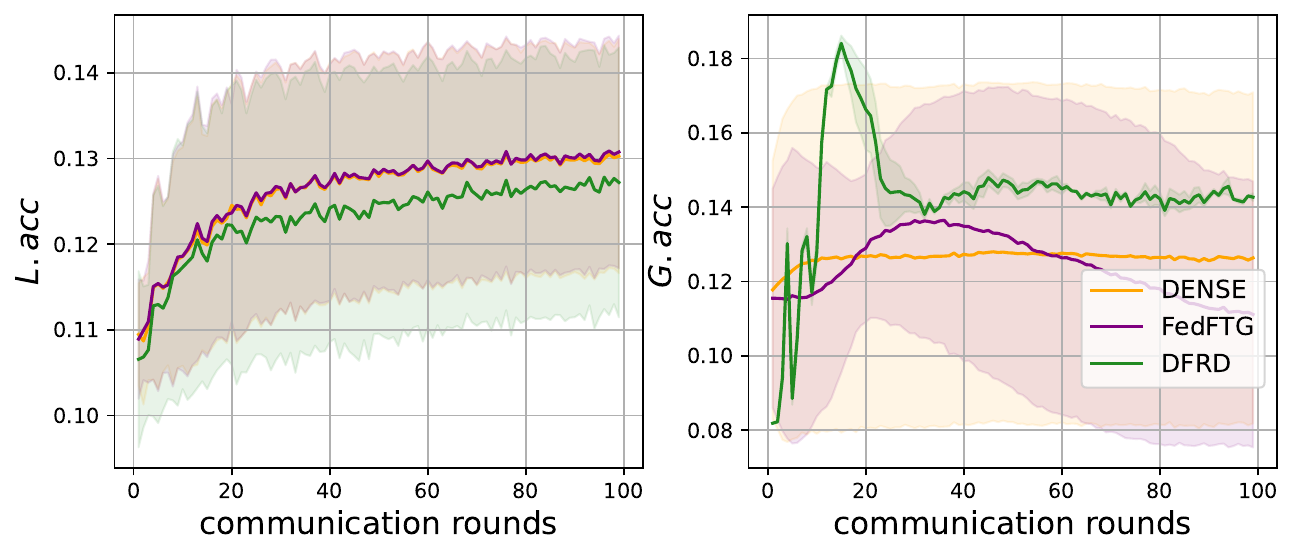}
        \caption{SVHN, $\omega=0.01$}
        \label{chutian3}
    \end{subfigure}\\
    \centering
    \begin{subfigure}{0.45\linewidth}
        \centering
        \includegraphics[width=1.0\linewidth]{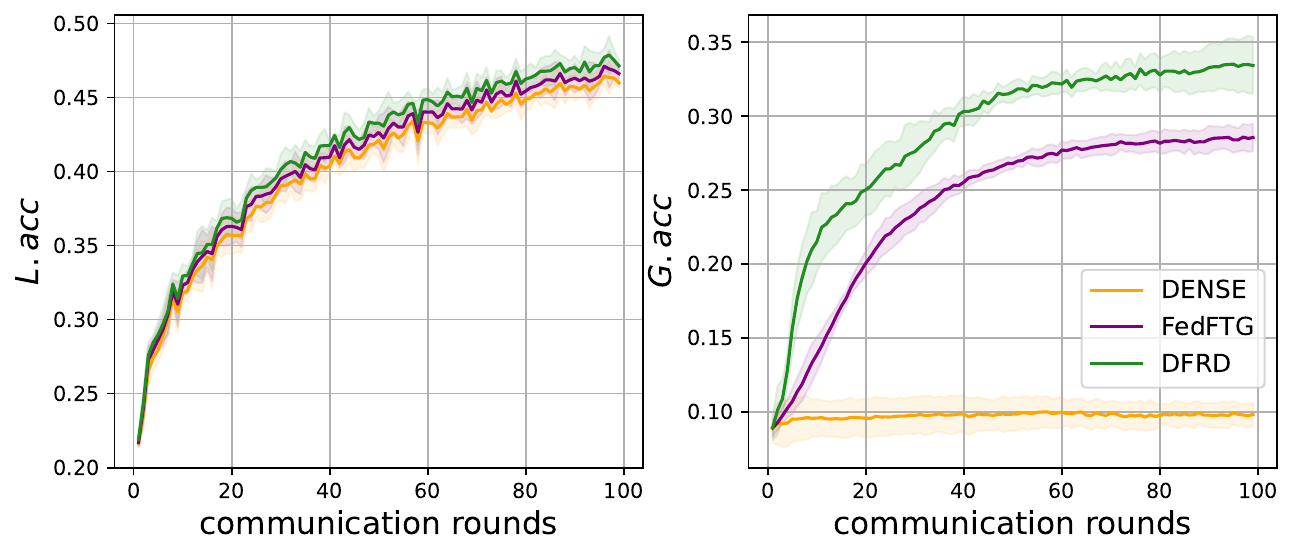}
        \caption{CIFAR-10, $\omega=1.0$}
        \label{chutian3}
    \end{subfigure}
    \centering
    \begin{subfigure}{0.45\linewidth}
        \centering
        \includegraphics[width=1.0\linewidth]{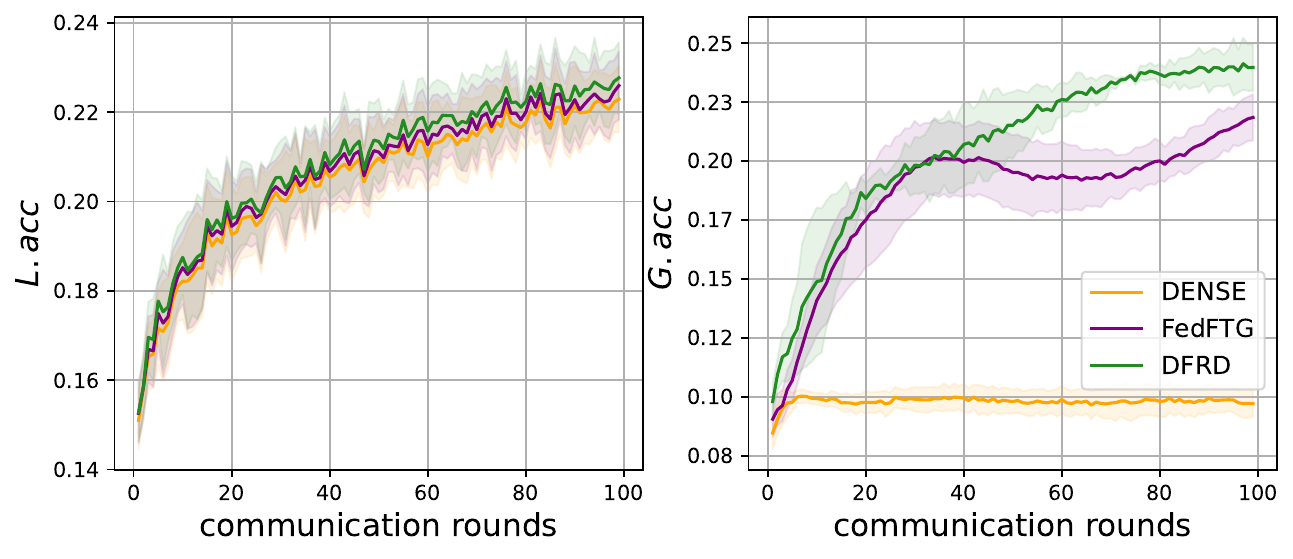}
        \caption{CIFAR-10, $\omega=0.1$}
        \label{chutian3}
    \end{subfigure}\\
    \centering
    \begin{subfigure}{0.45\linewidth}
        \centering
        \includegraphics[width=1.0\linewidth]{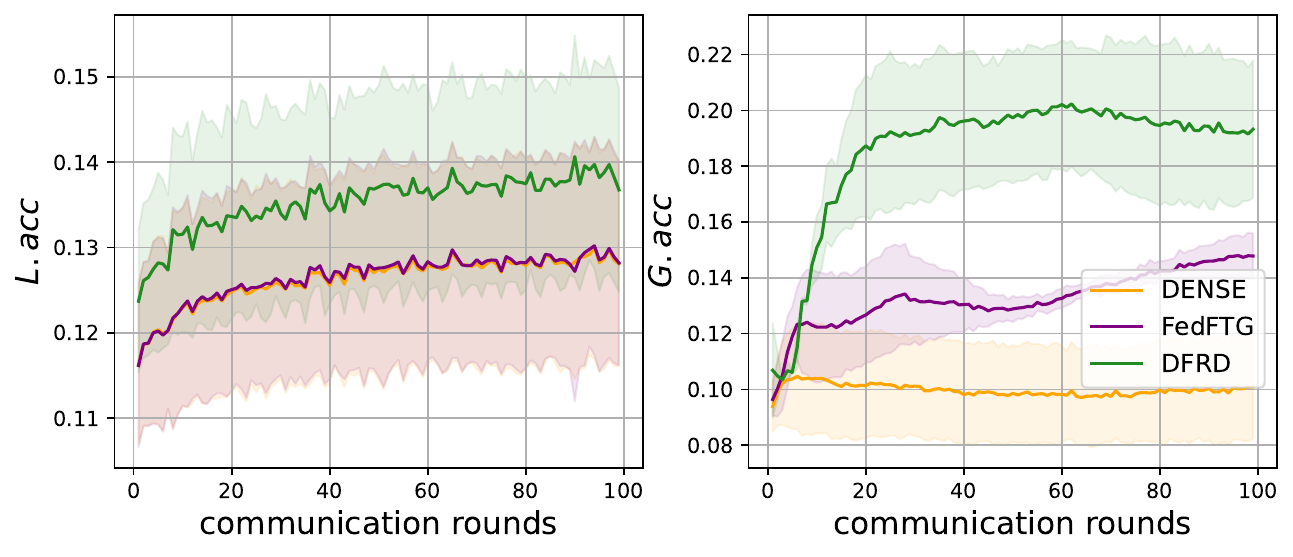}
        \caption{CIFAR-10, $\omega=0.01$}
        \label{chutian3}
    \end{subfigure}
    \centering
    \begin{subfigure}{0.45\linewidth}
        \centering
        \includegraphics[width=1.0\linewidth]{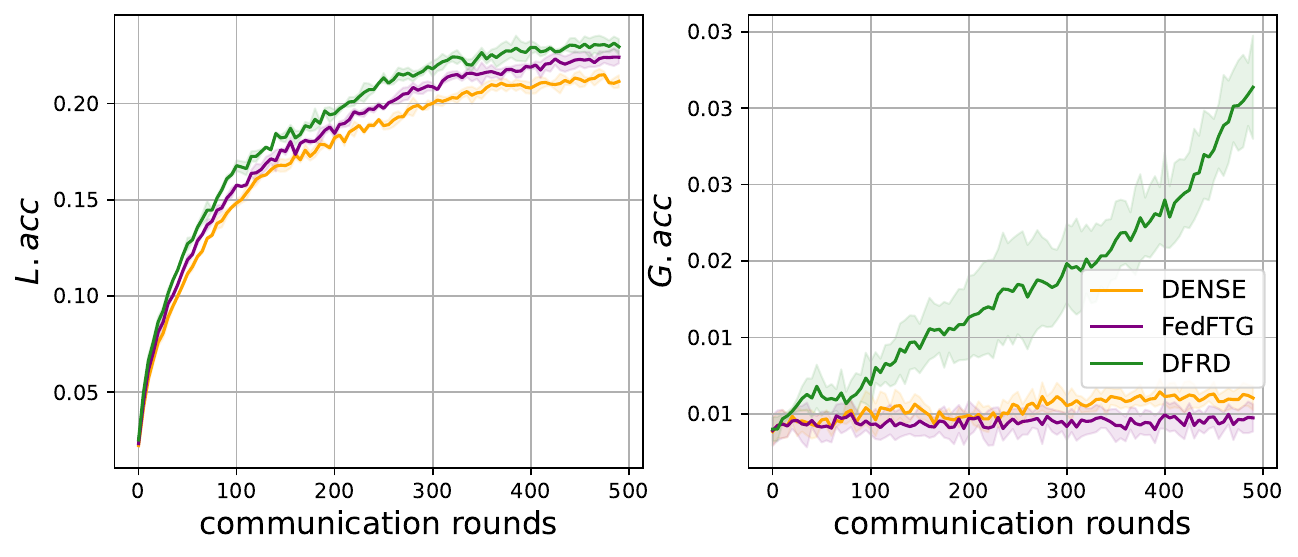}
        \caption{CIFAR-100, $\omega=1.0$}
        \label{chutian3}
    \end{subfigure}\\
    \centering
    \begin{subfigure}{0.45\linewidth}
        \centering
        \includegraphics[width=1.0\linewidth]{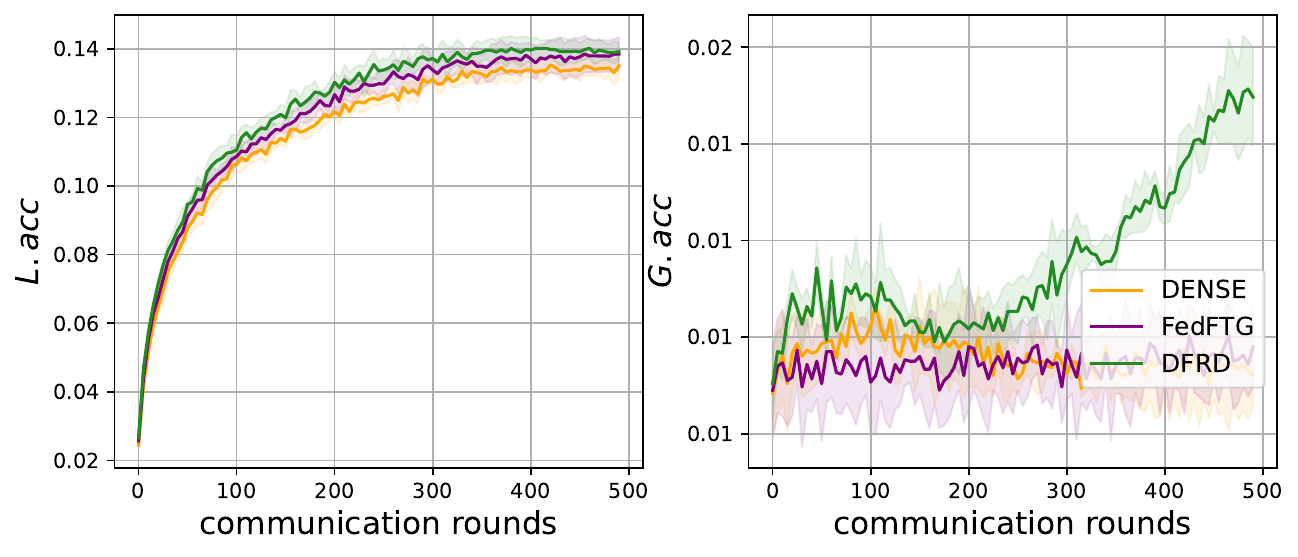}
        \caption{CIFAR-100, $\omega=0.1$}
        \label{chutian3}
    \end{subfigure}
    \centering
    \begin{subfigure}{0.45\linewidth}
        \centering
        \includegraphics[width=1.0\linewidth]{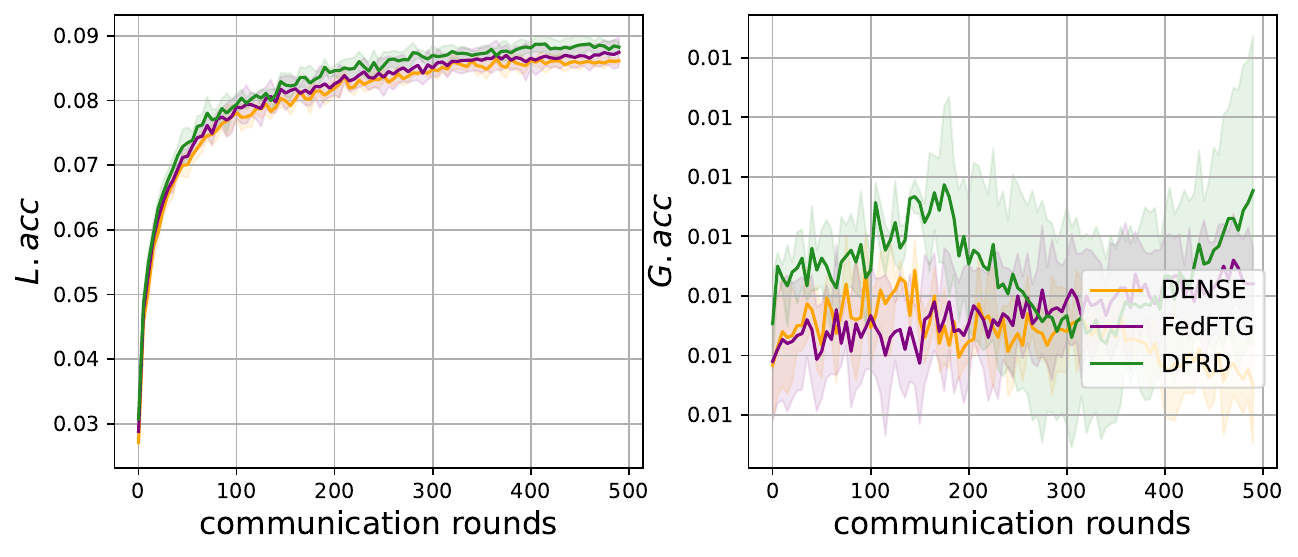}
        \caption{CIFAR-100, $\omega=0.01$}
        \label{chutian3}
    \end{subfigure}
    \caption{Learning curves of distinct data-free methods across $\omega \in \{0.01, 0.1, 1.0\}$ on FMNIST, SVHN, CIFAR-10 and CIFAR-100.} 
    \label{model_homo_data_free:}
\end{figure*}

\begin{figure*}[h]\captionsetup[subfigure]{font=scriptsize}
    \centering
    \begin{subfigure}{0.45\linewidth}
        \centering
        \includegraphics[width=1.0\linewidth]{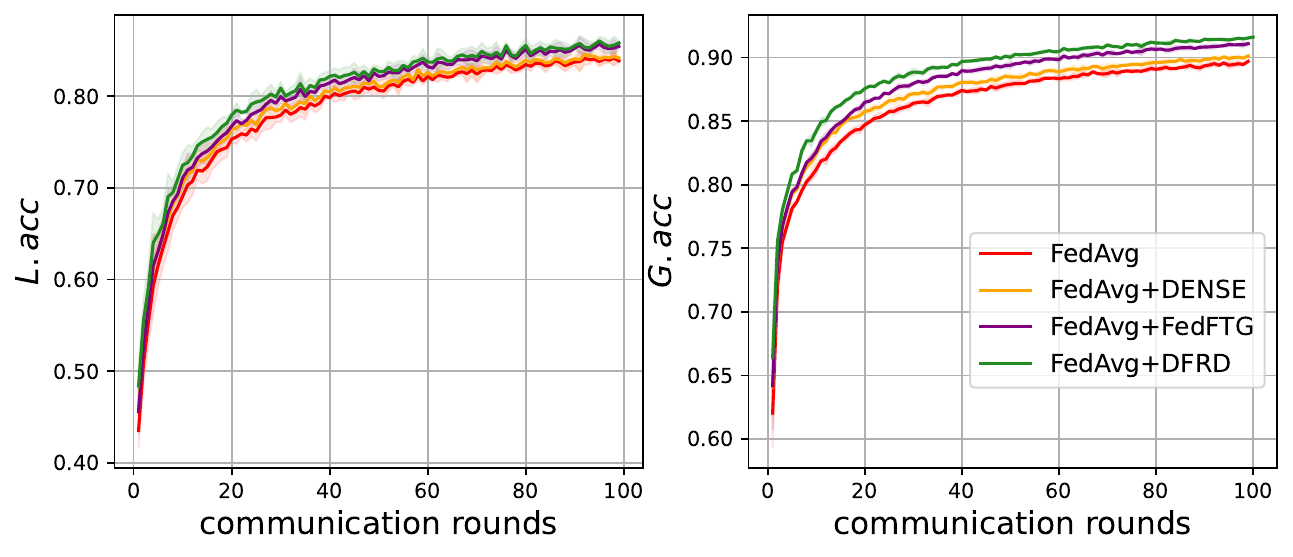}
        \caption{FMNIST, $\omega=1.0$}
        \label{chutian3}
    \end{subfigure}
    \centering
    \begin{subfigure}{0.45\linewidth}
        \centering
        \includegraphics[width=1.0\linewidth]{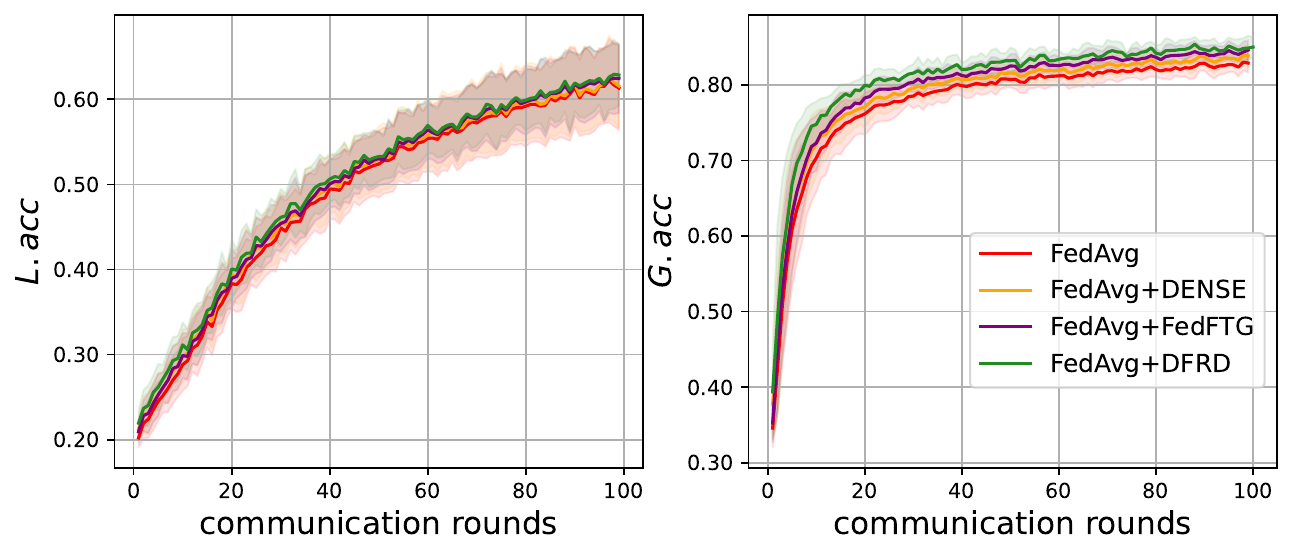}
        \caption{FMNIST, $\omega=0.1$}
        \label{chutian3}
    \end{subfigure} \\
    \centering
    \begin{subfigure}{0.45\linewidth}
        \centering
        \includegraphics[width=1.0\linewidth]{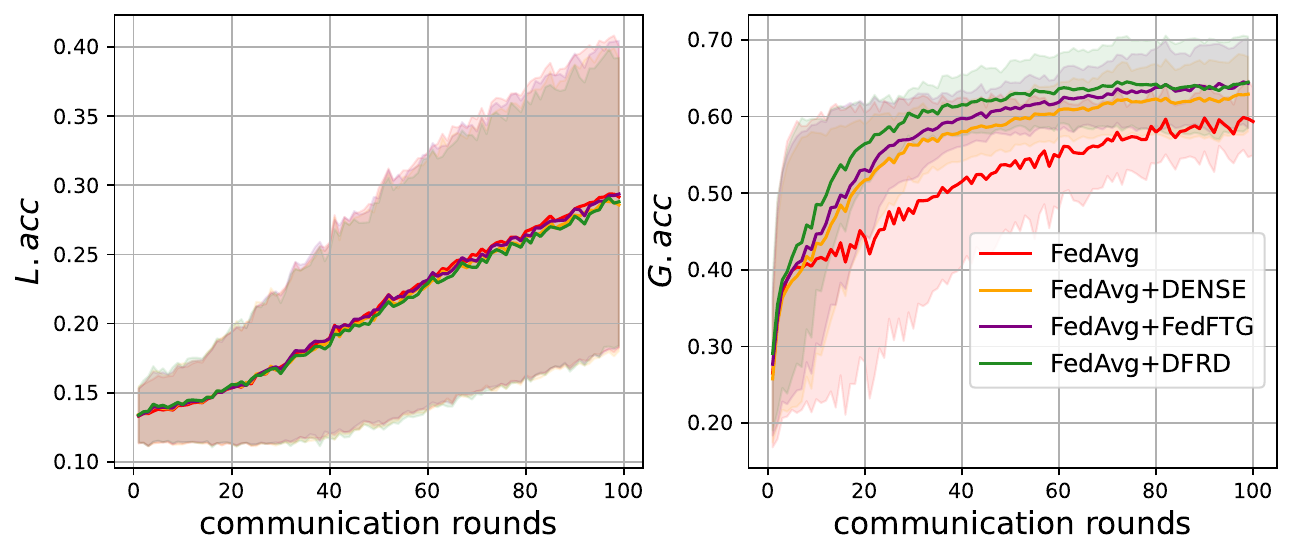}
        \caption{FMNIST, $\omega=0.01$}
        \label{chutian3}
    \end{subfigure}
    \centering
    \begin{subfigure}{0.45\linewidth}
        \centering
        \includegraphics[width=1.0\linewidth]{fine_tune_SVHN_1.0_comm_round.pdf}
        \caption{SVHN, $\omega=1.0$}
        \label{chutian3}
    \end{subfigure} \\
    \centering
    \begin{subfigure}{0.45\linewidth}
        \centering
        \includegraphics[width=1.0\linewidth]{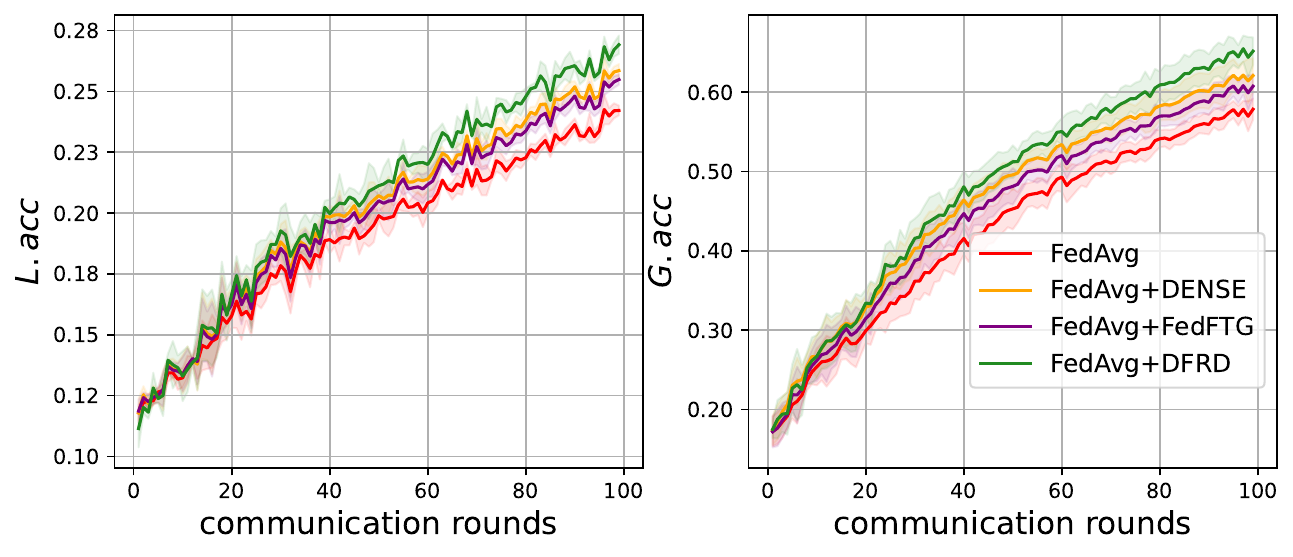}
        \caption{SVHN, $\omega=0.1$}
        \label{chutian3}
    \end{subfigure}
    \centering
    \begin{subfigure}{0.45\linewidth}
        \centering
        \includegraphics[width=1.0\linewidth]{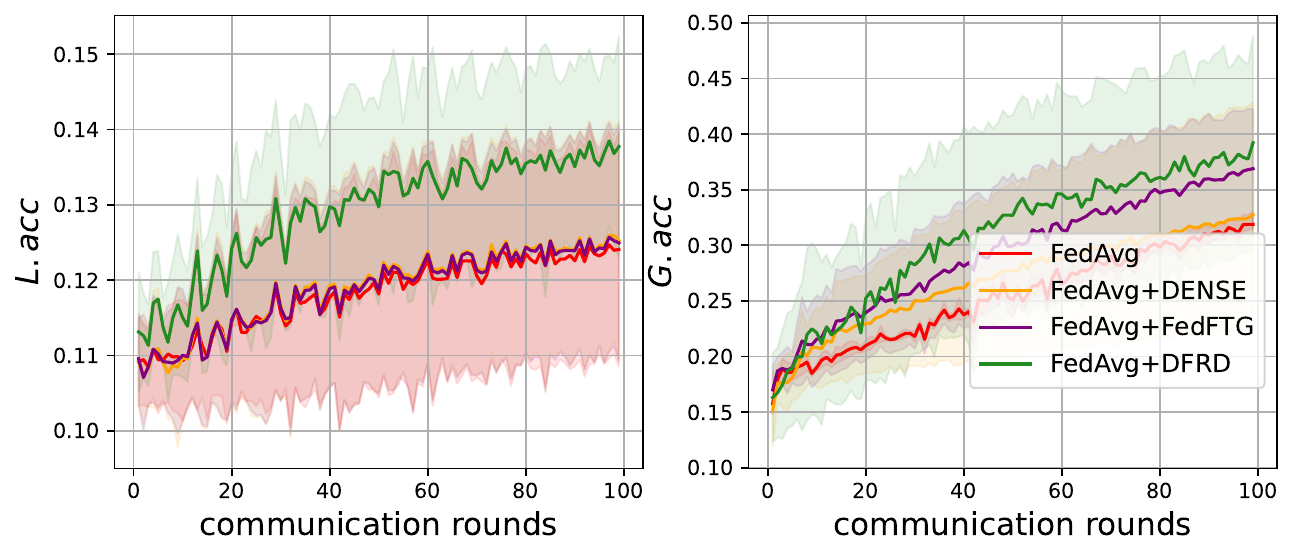}
        \caption{SVHN, $\omega=0.01$}
        \label{chutian3}
    \end{subfigure}\\
    \centering
    \begin{subfigure}{0.45\linewidth}
        \centering
        \includegraphics[width=1.0\linewidth]{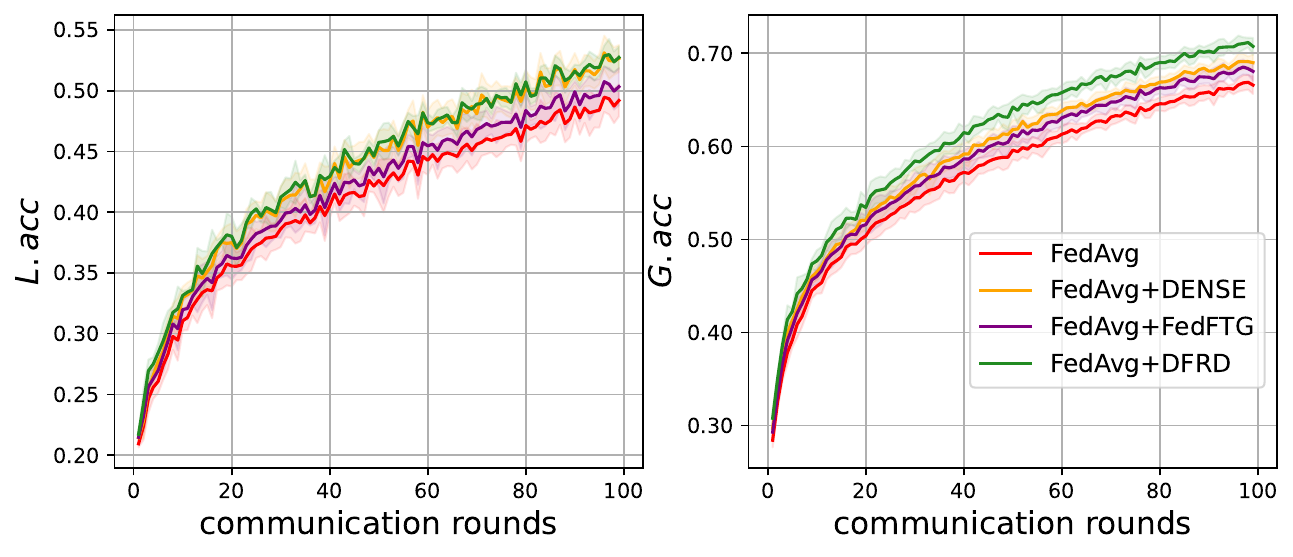}
        \caption{CIFAR-10, $\omega=1.0$}
        \label{chutian3}
    \end{subfigure}
    \centering
    \begin{subfigure}{0.45\linewidth}
        \centering
        \includegraphics[width=1.0\linewidth]{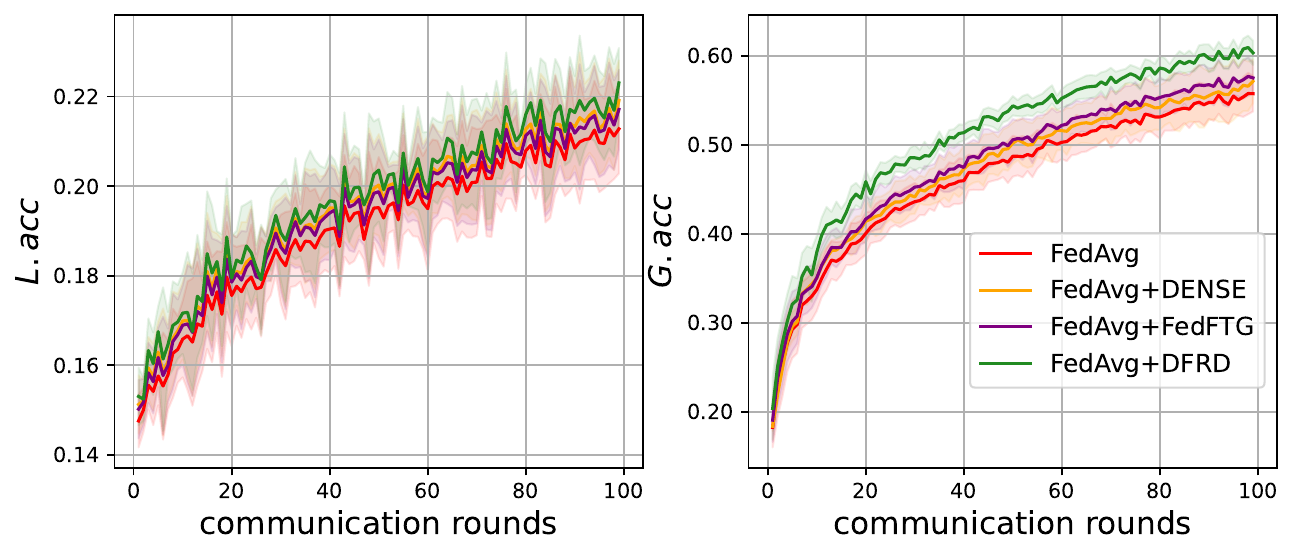}
        \caption{CIFAR-10, $\omega=0.1$}
        \label{chutian3}
    \end{subfigure}\\
    \centering
    \begin{subfigure}{0.45\linewidth}
        \centering
        \includegraphics[width=1.0\linewidth]{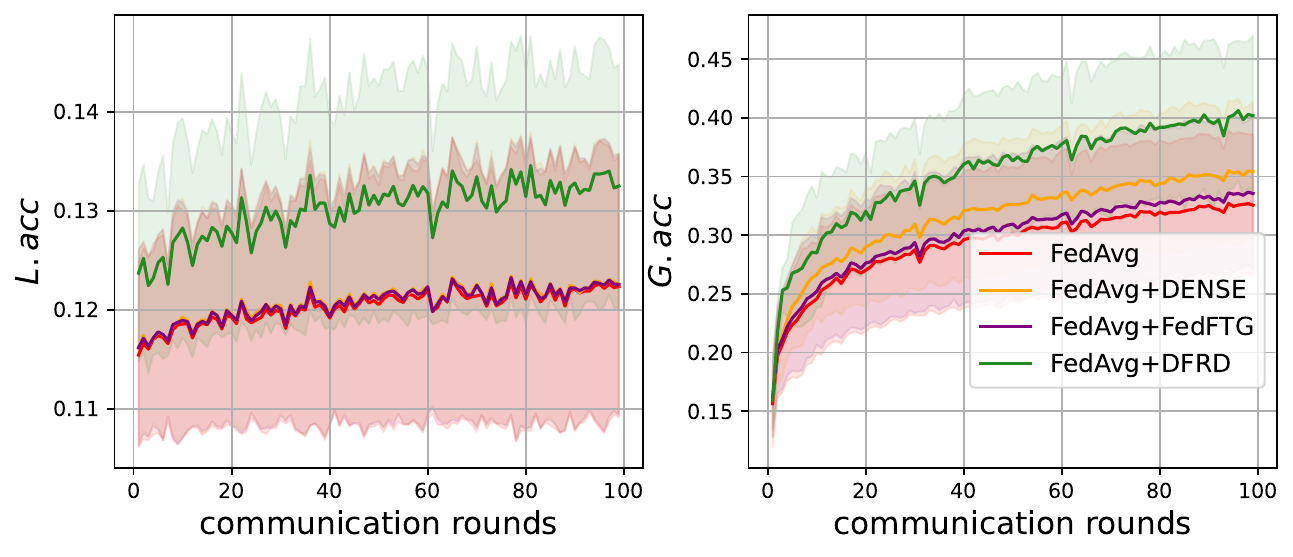}
        \caption{CIFAR-10, $\omega=0.01$}
        \label{chutian3}
    \end{subfigure}
    \centering
    \begin{subfigure}{0.45\linewidth}
        \centering
        \includegraphics[width=1.0\linewidth]{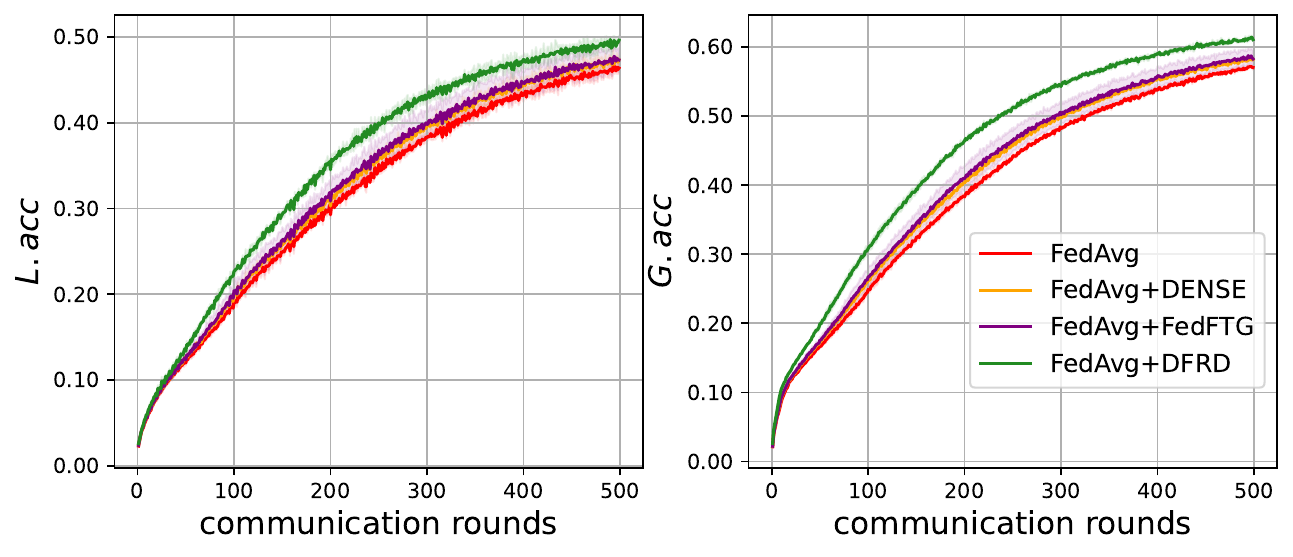}
        \caption{CIFAR-100, $\omega=1.0$}
        \label{chutian3}
    \end{subfigure}\\
    \centering
    \begin{subfigure}{0.45\linewidth}
        \centering
        \includegraphics[width=1.0\linewidth]{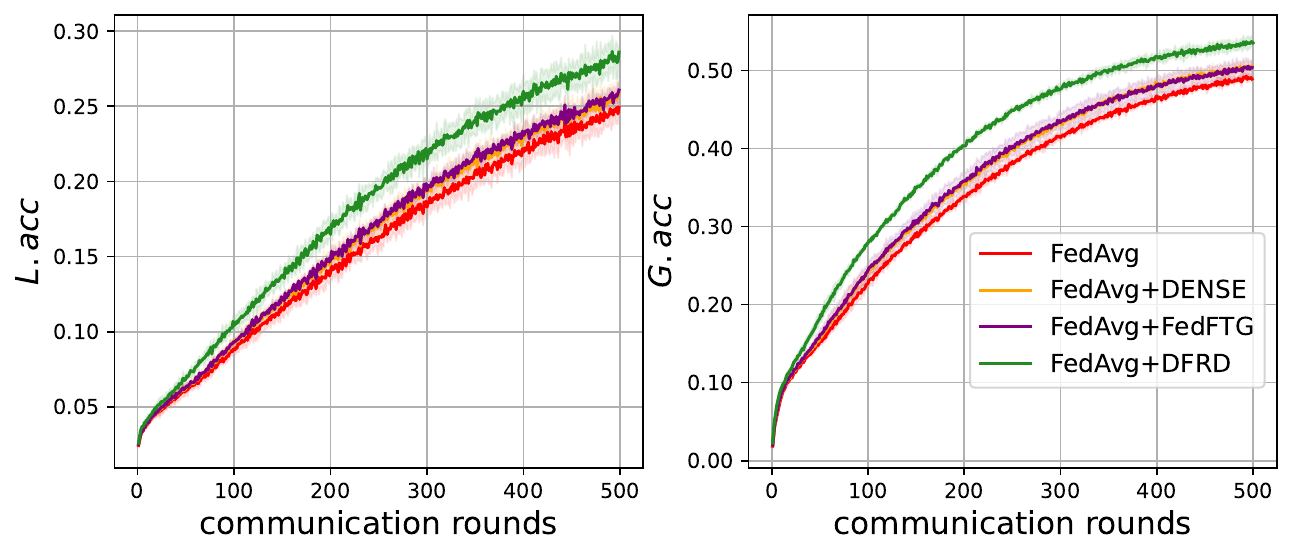}
        \caption{CIFAR-100, $\omega=0.1$}
        \label{chutian3}
    \end{subfigure}
    \centering
    \begin{subfigure}{0.45\linewidth}
        \centering
        \includegraphics[width=1.0\linewidth]{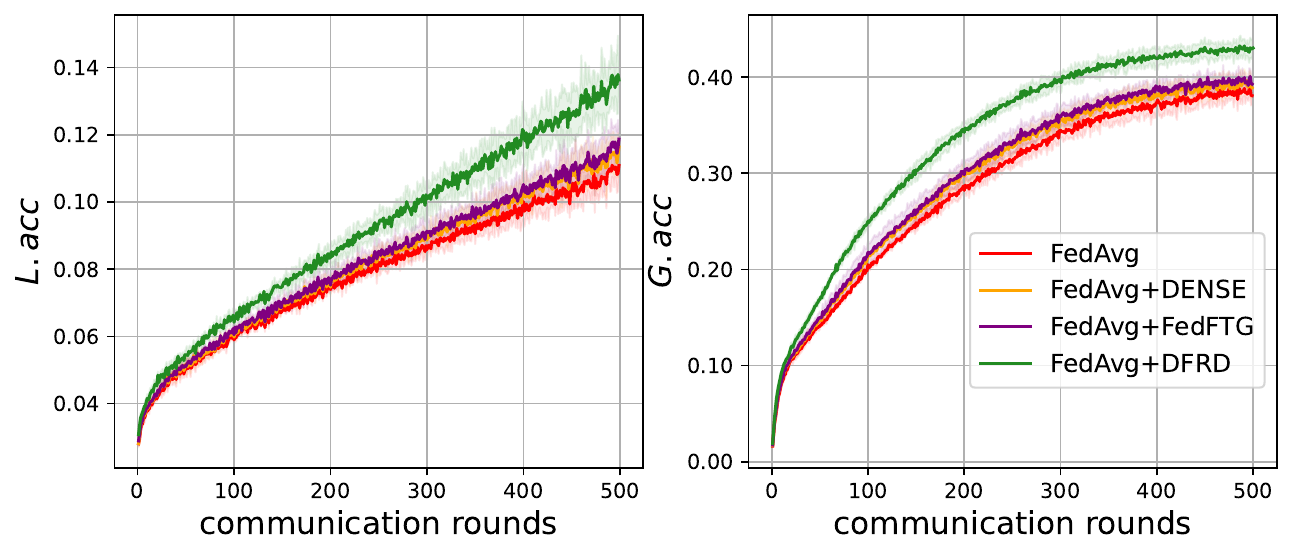}
        \caption{CIFAR-100, $\omega=0.01$}
        \label{chutian3}
    \end{subfigure}
    \caption{Learning curves of distinct fine-tuning methods based on FedAvg across $\omega \in \{0.01, 0.1, 1.0\}$ on FMNIST, SVHN, CIFAR-10 and CIFAR-100.} 
    \label{model_homo_fine_tune:}
\end{figure*}

\begin{figure*}[h]\captionsetup[subfigure]{font=scriptsize}
    \centering
    \begin{subfigure}{0.45\linewidth}
        \centering
        \includegraphics[width=1.0\linewidth]{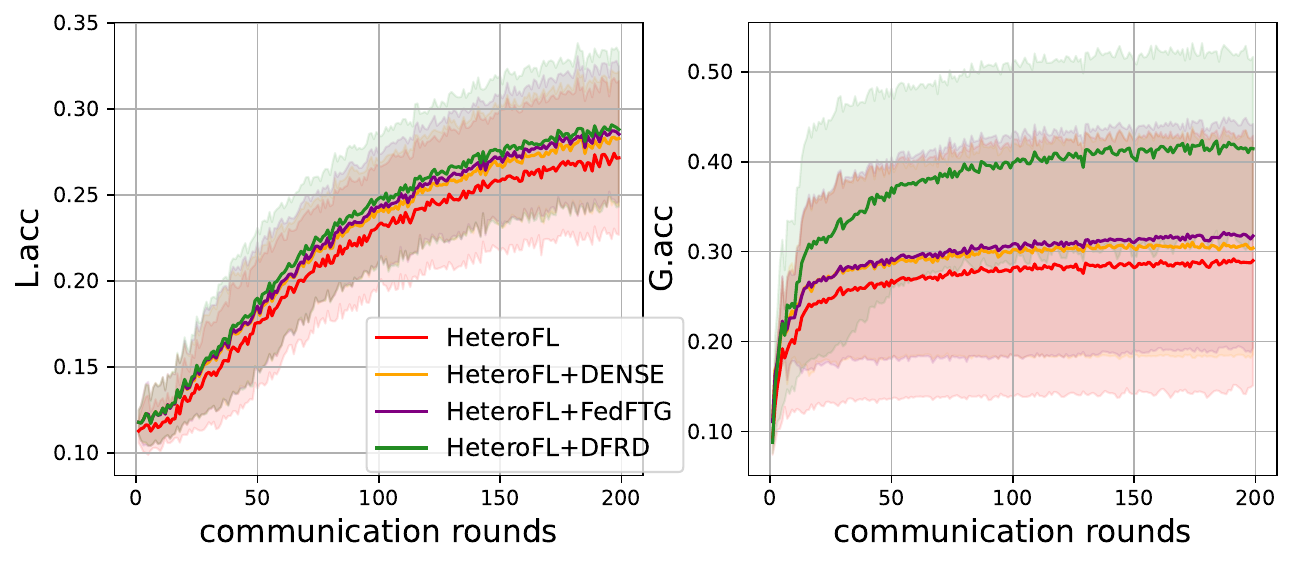}
        \caption{SVHN, $\rho=5$}
        \label{chutian3}
    \end{subfigure}
    \centering
    \begin{subfigure}{0.45\linewidth}
        \centering
        \includegraphics[width=1.0\linewidth]{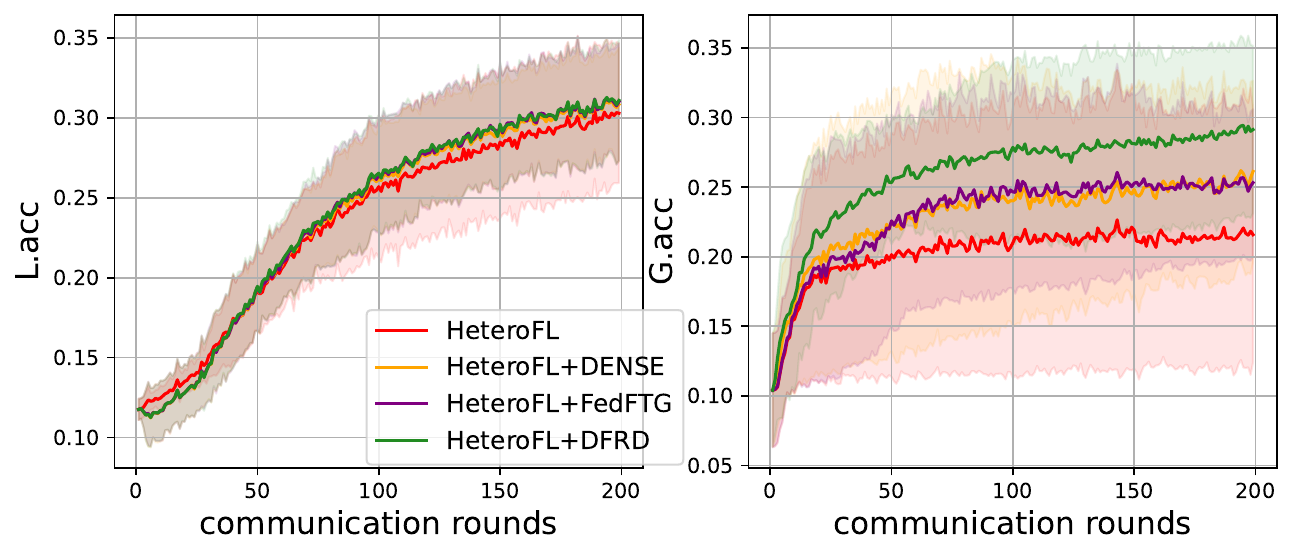}
        \caption{SVHN, $\rho=10$}
        \label{chutian3}
    \end{subfigure} \\
    \centering
    \begin{subfigure}{0.45\linewidth}
        \centering
        \includegraphics[width=1.0\linewidth]{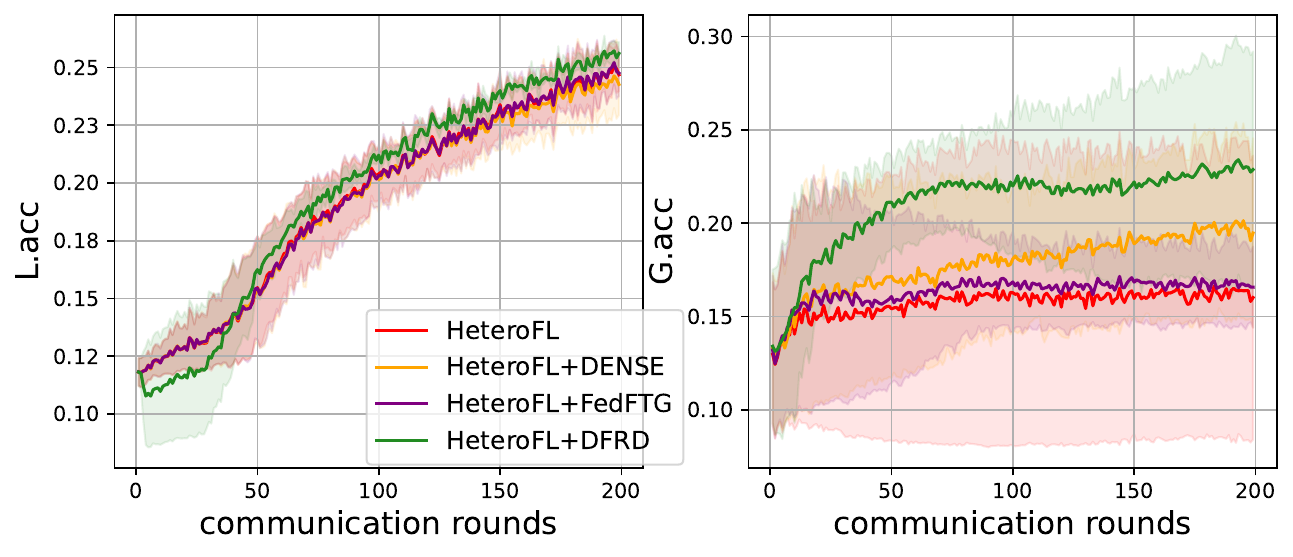}
        \caption{SVHN, $\rho=40$}
        \label{chutian3}
    \end{subfigure}
    \centering
    \begin{subfigure}{0.45\linewidth}
        \centering
        \includegraphics[width=1.0\linewidth]{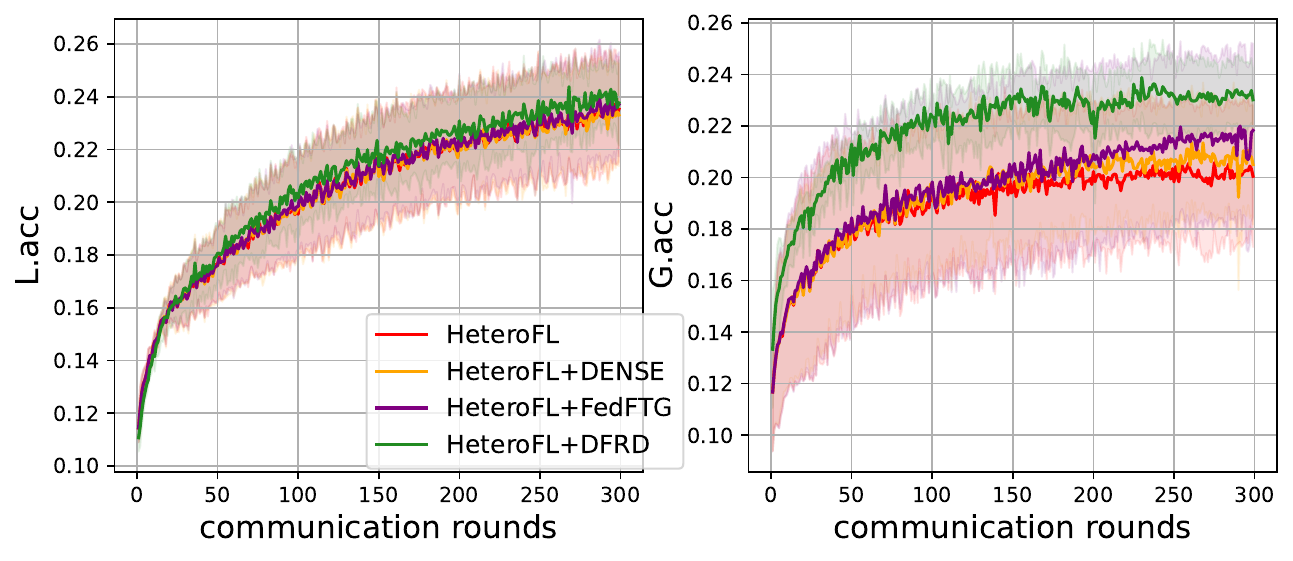}
        \caption{CIFAR-10, $\rho=5$}
        \label{chutian3}
    \end{subfigure} \\
    \centering
    \begin{subfigure}{0.45\linewidth}
        \centering
        \includegraphics[width=1.0\linewidth]{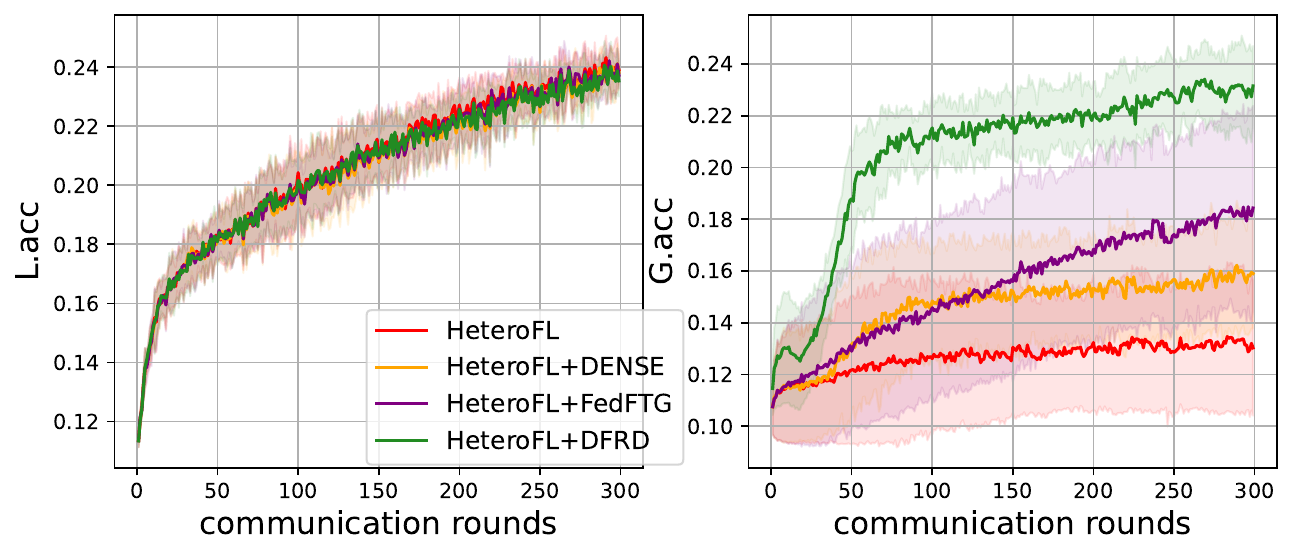}
        \caption{CIFAR-10, $\rho=10$}
        \label{chutian3}
    \end{subfigure}
    \centering
    \begin{subfigure}{0.45\linewidth}
        \centering
        \includegraphics[width=1.0\linewidth]{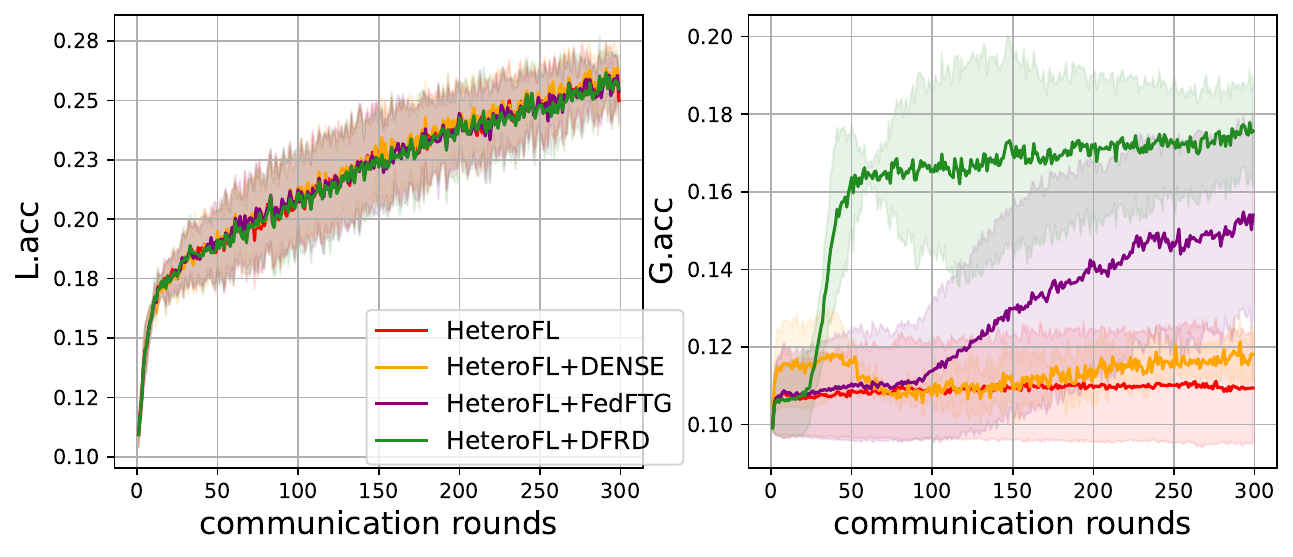}
        \caption{CIFAR-10, $\rho=40$}
        \label{chutian3}
    \end{subfigure}\\
    \centering
    \begin{subfigure}{0.45\linewidth}
        \centering
        \includegraphics[width=1.0\linewidth]{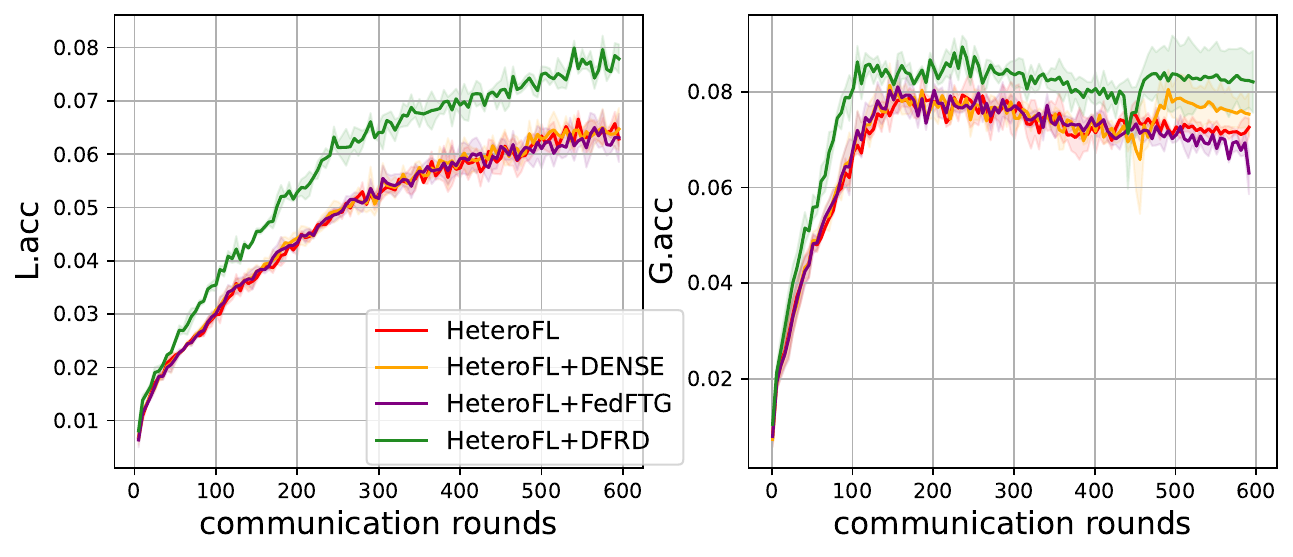}
        \caption{Tiny-ImageNet, $\rho=5$}
        \label{chutian3}
    \end{subfigure}
    \centering
    \begin{subfigure}{0.45\linewidth}
        \centering
        \includegraphics[width=1.0\linewidth]{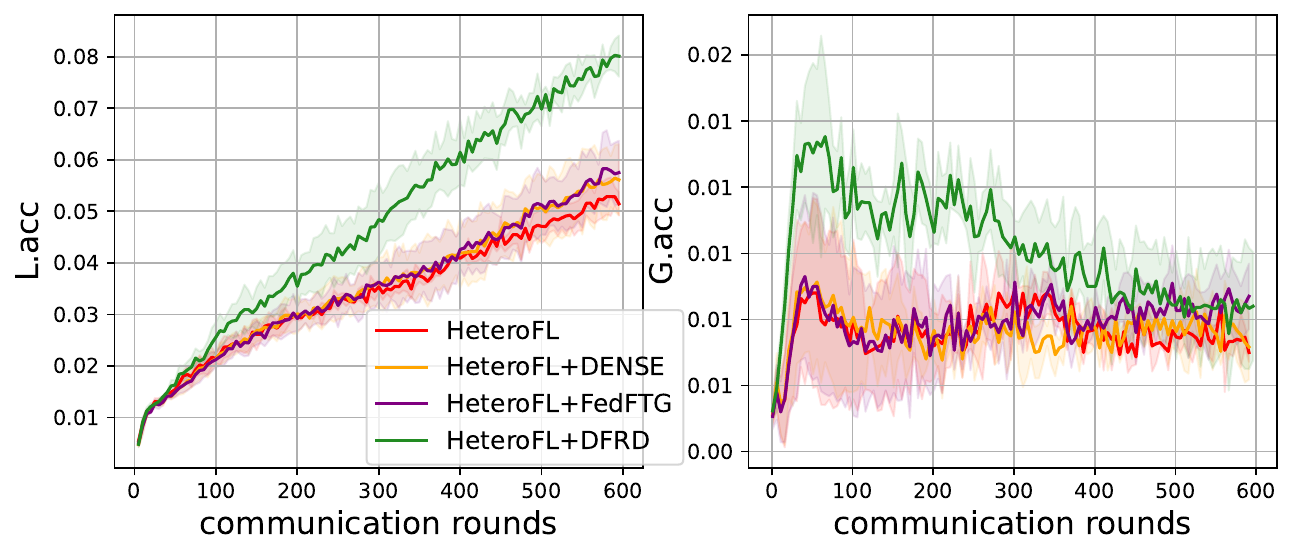}
        \caption{Tiny-ImageNet, $\rho=10$}
        \label{chutian3}
    \end{subfigure}\\
    \centering
    \begin{subfigure}{0.45\linewidth}
        \centering
        \includegraphics[width=1.0\linewidth]{Tiny-Imagenet_static_level_10_comm_round.pdf}
        \caption{Tiny-ImageNet, $\rho=40$}
        \label{chutian3}
    \end{subfigure}
    \centering
    \begin{subfigure}{0.45\linewidth}
        \centering
        \includegraphics[width=1.0\linewidth]{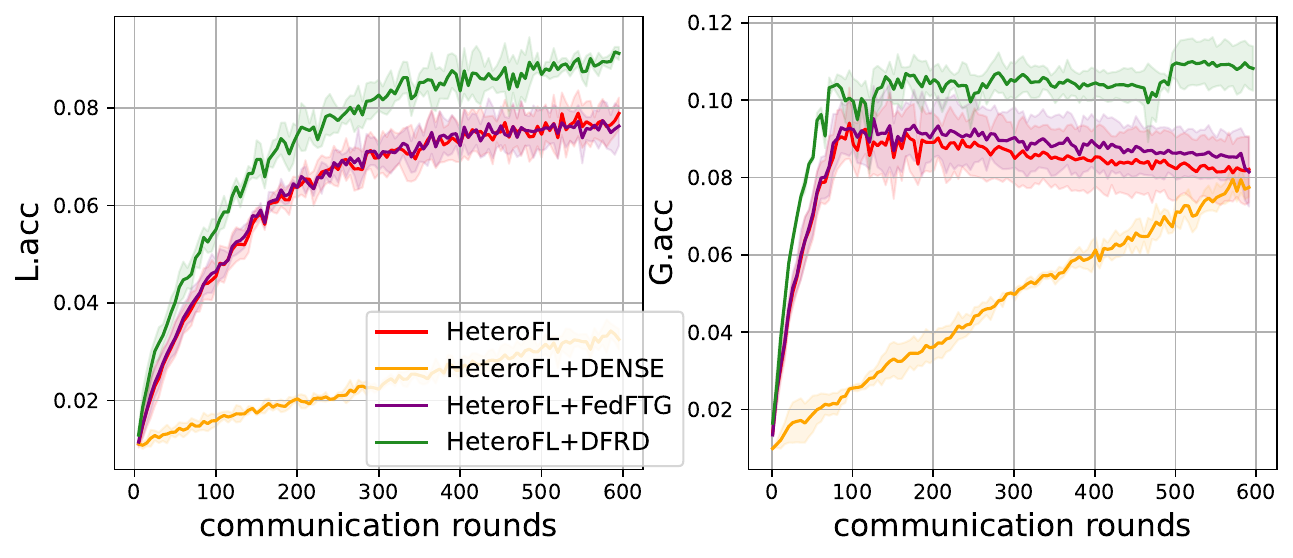}
        \caption{FOOD101, $\rho=5$}
        \label{chutian3}
    \end{subfigure}\\
    \centering
    \begin{subfigure}{0.45\linewidth}
        \centering
        \includegraphics[width=1.0\linewidth]{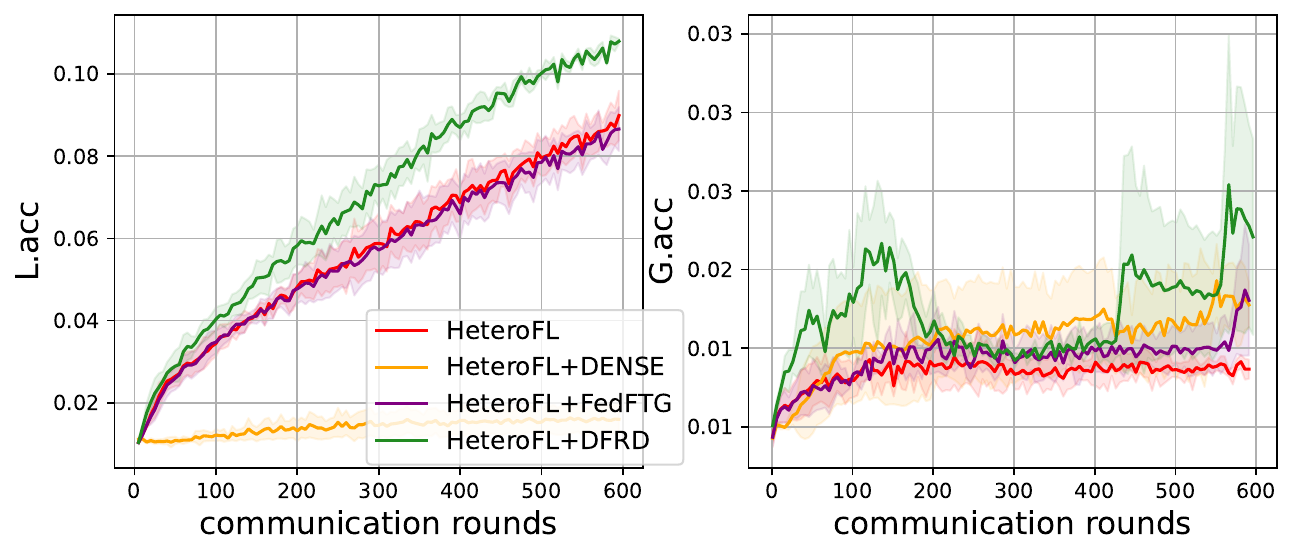}
        \caption{FOOD101, $\rho=10$}
        \label{chutian3}
    \end{subfigure}
    \centering
    \begin{subfigure}{0.45\linewidth}
        \centering
        \includegraphics[width=1.0\linewidth]{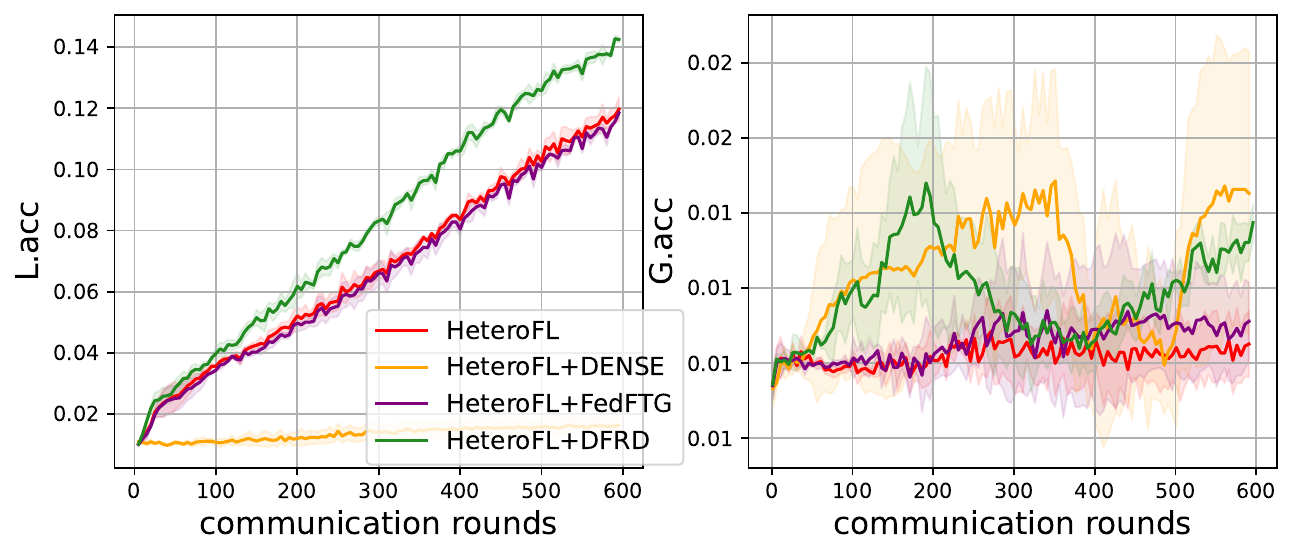}
        \caption{FOOD101, $\rho=40$}
        \label{chutian3}
    \end{subfigure}
    \caption{Learning curves of distinct fine-tuning methods based on HeteroFL across $\rho \in \{5, 10, 40\}$ on SVHN, CIFAR-10, Tiny-ImageNet and FOOD101.} 
    \label{model_static_heter:}
\end{figure*}

\begin{figure*}[h]\captionsetup[subfigure]{font=scriptsize}
    \centering
    \begin{subfigure}{0.45\linewidth}
        \centering
        \includegraphics[width=1.0\linewidth]{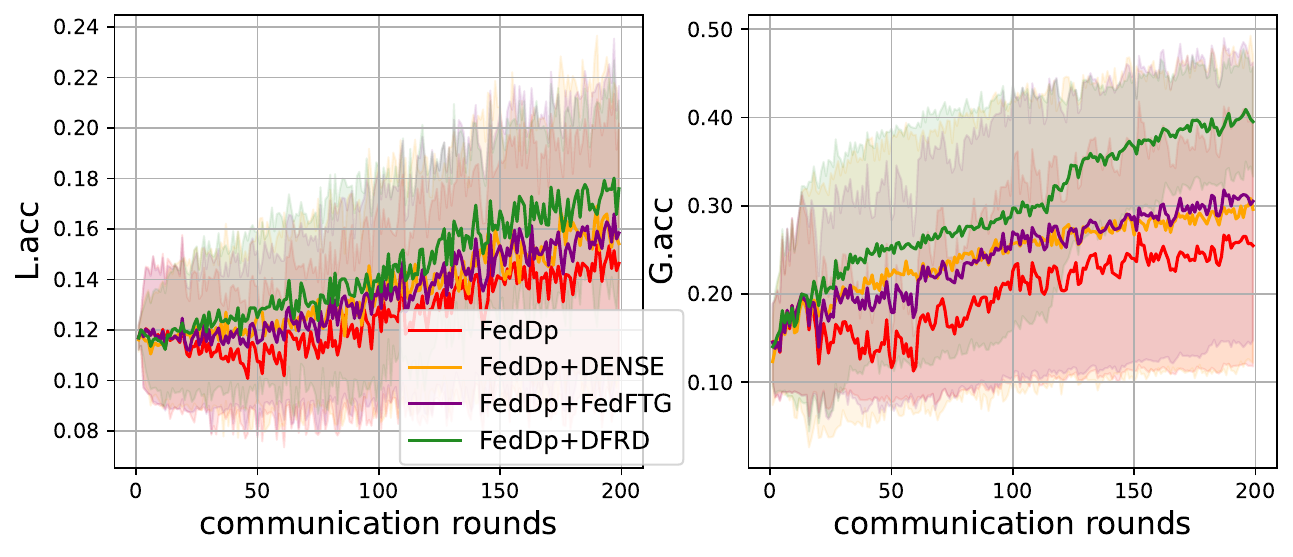}
        \caption{SVHN, $\rho=5$}
        \label{chutian3}
    \end{subfigure}
    \centering
    \begin{subfigure}{0.45\linewidth}
        \centering
        \includegraphics[width=1.0\linewidth]{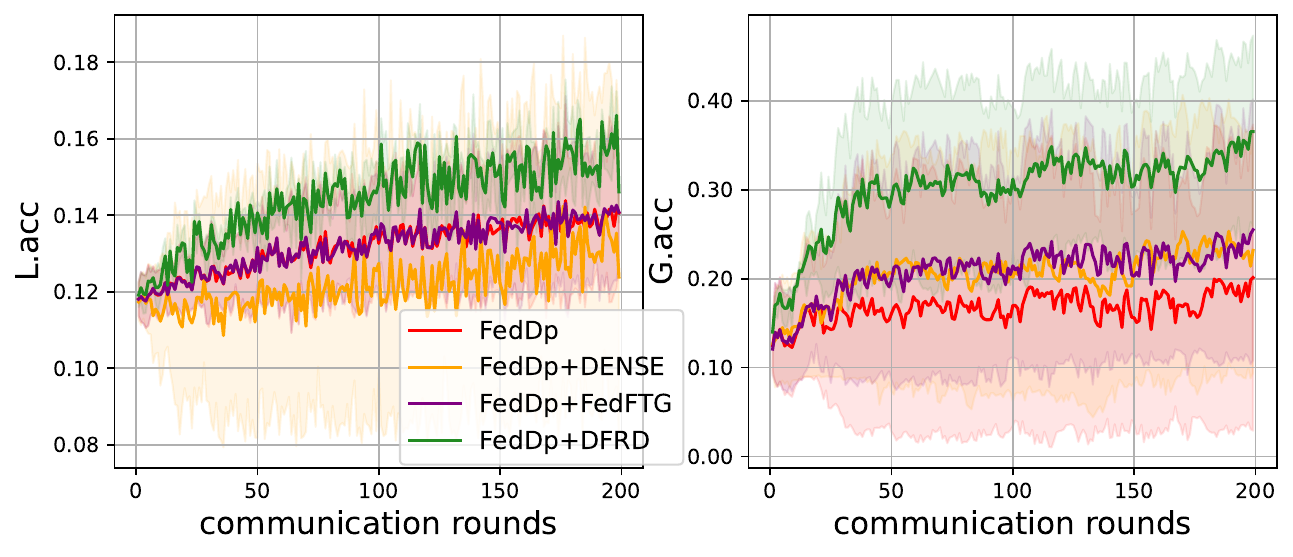}
        \caption{SVHN, $\rho=10$}
        \label{chutian3}
    \end{subfigure} \\
    \centering
    \begin{subfigure}{0.45\linewidth}
        \centering
        \includegraphics[width=1.0\linewidth]{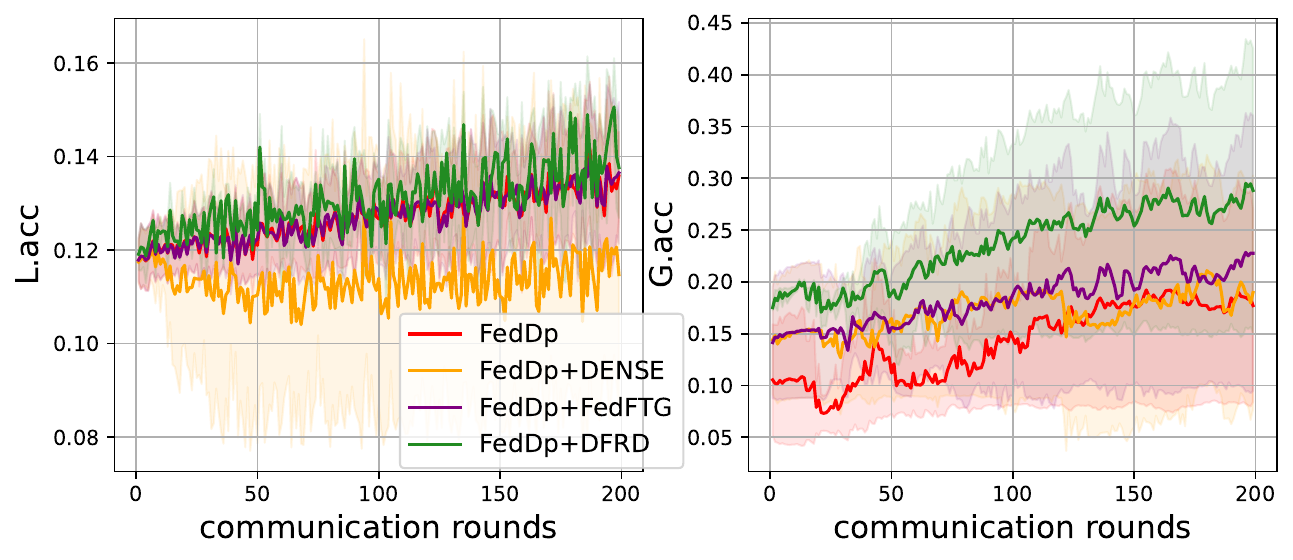}
        \caption{SVHN, $\rho=40$}
        \label{chutian3}
    \end{subfigure}
    \centering
    \begin{subfigure}{0.45\linewidth}
        \centering
        \includegraphics[width=1.0\linewidth]{CIFAR-10_random_level_5_comm_round.pdf}
        \caption{CIFAR-10, $\rho=5$}
        \label{chutian3}
    \end{subfigure} \\
    \centering
    \begin{subfigure}{0.45\linewidth}
        \centering
        \includegraphics[width=1.0\linewidth]{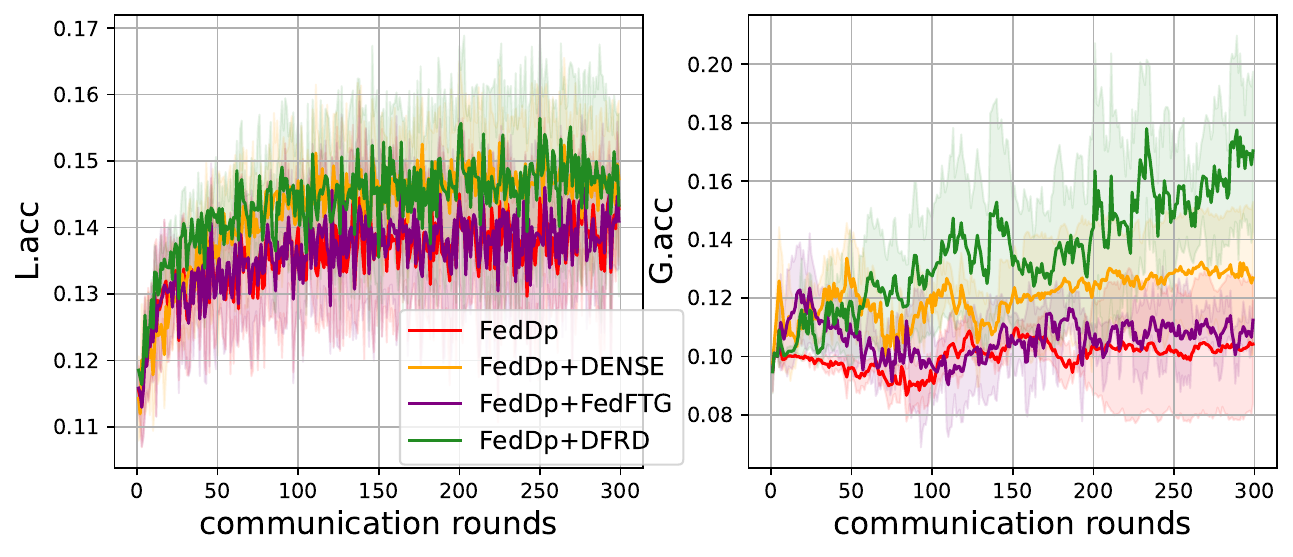}
        \caption{CIFAR-10, $\rho=10$}
        \label{chutian3}
    \end{subfigure}
    \centering
    \begin{subfigure}{0.45\linewidth}
        \centering
        \includegraphics[width=1.0\linewidth]{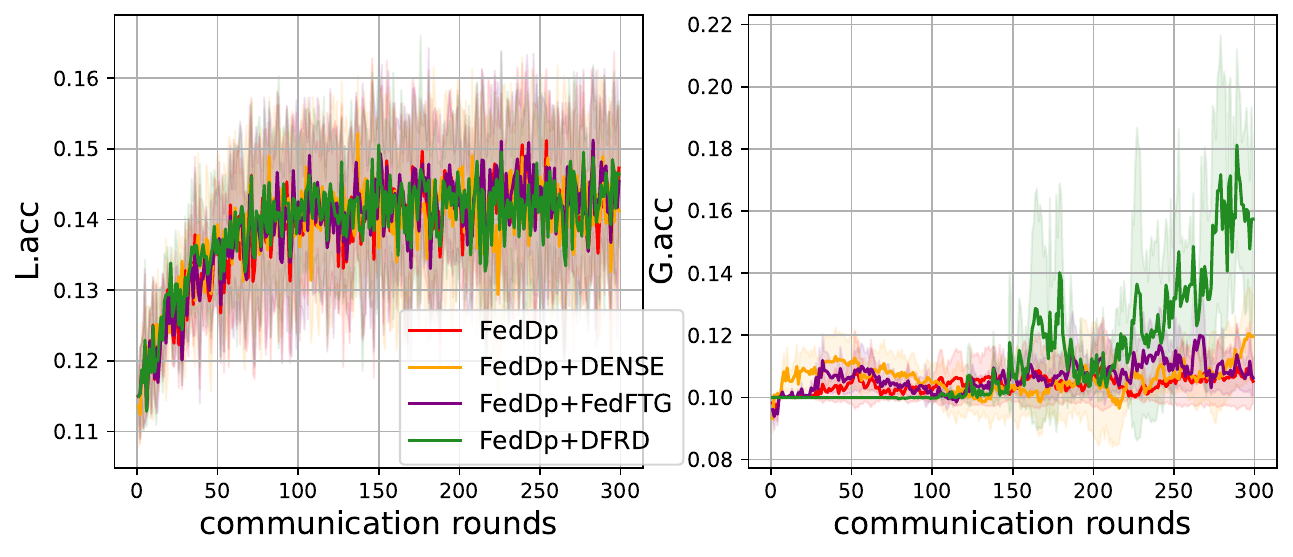}
        \caption{CIFAR-10, $\rho=40$}
        \label{chutian3}
    \end{subfigure}\\
    \centering
    \begin{subfigure}{0.45\linewidth}
        \centering
        \includegraphics[width=1.0\linewidth]{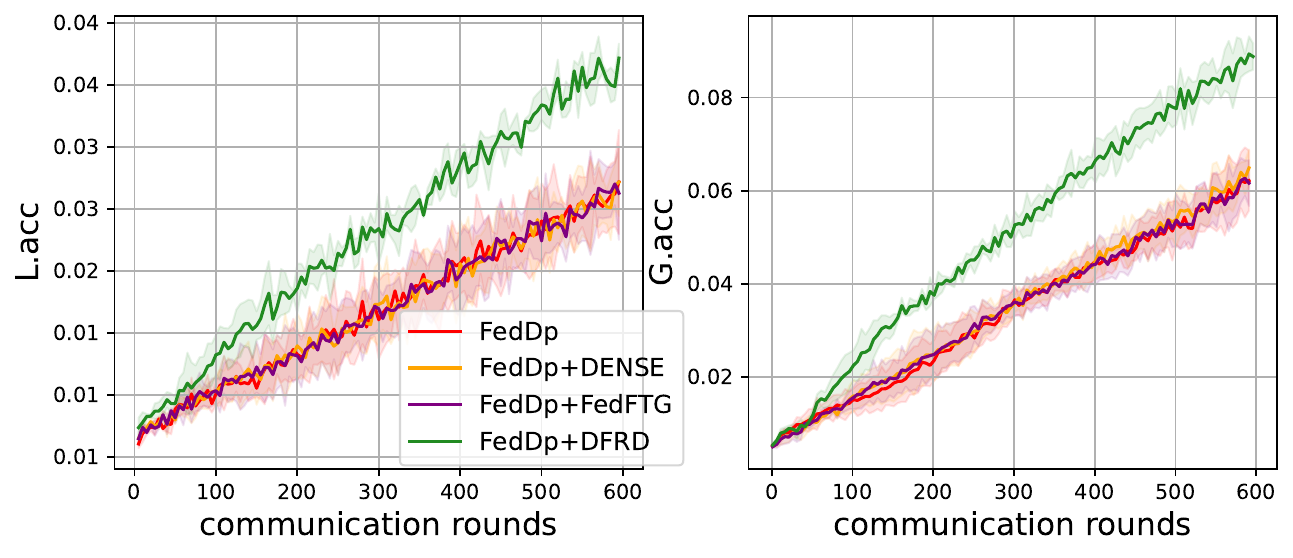}
        \caption{Tiny-ImageNet, $\rho=5$}
        \label{chutian3}
    \end{subfigure}
    \centering
    \begin{subfigure}{0.45\linewidth}
        \centering
        \includegraphics[width=1.0\linewidth]{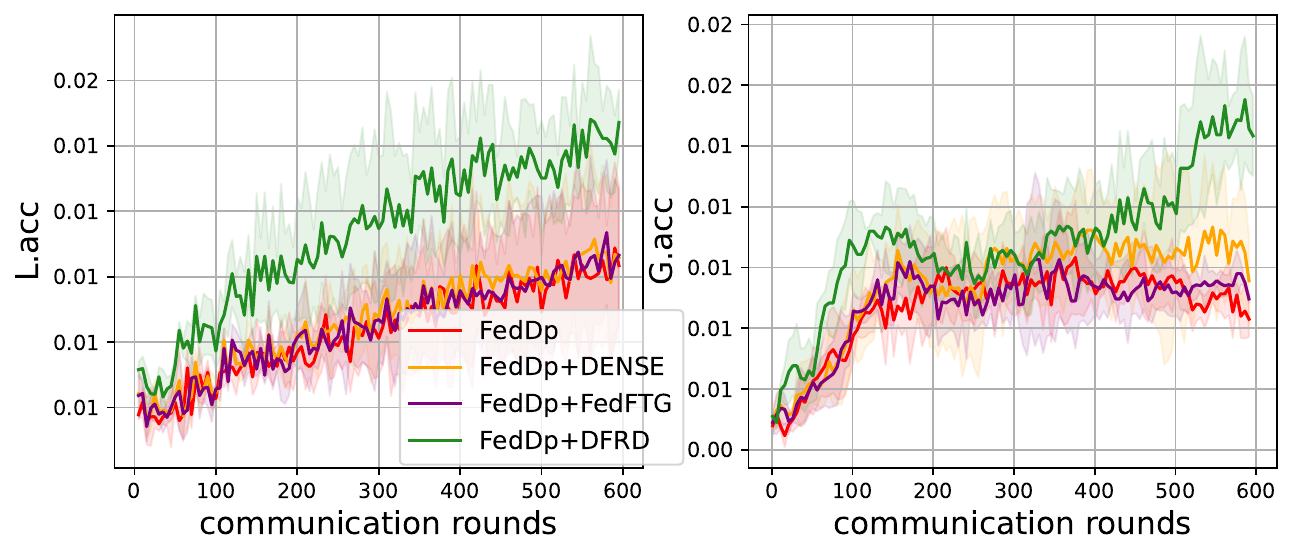}
        \caption{Tiny-ImageNet, $\rho=10$}
        \label{chutian3}
    \end{subfigure}\\
    \centering
    \begin{subfigure}{0.45\linewidth}
        \centering
        \includegraphics[width=1.0\linewidth]{Tiny-Imagenet_random_level_10_comm_round.pdf}
        \caption{Tiny-ImageNet, $\rho=40$}
        \label{chutian3}
    \end{subfigure}
    \centering
    \begin{subfigure}{0.45\linewidth}
        \centering
        \includegraphics[width=1.0\linewidth]{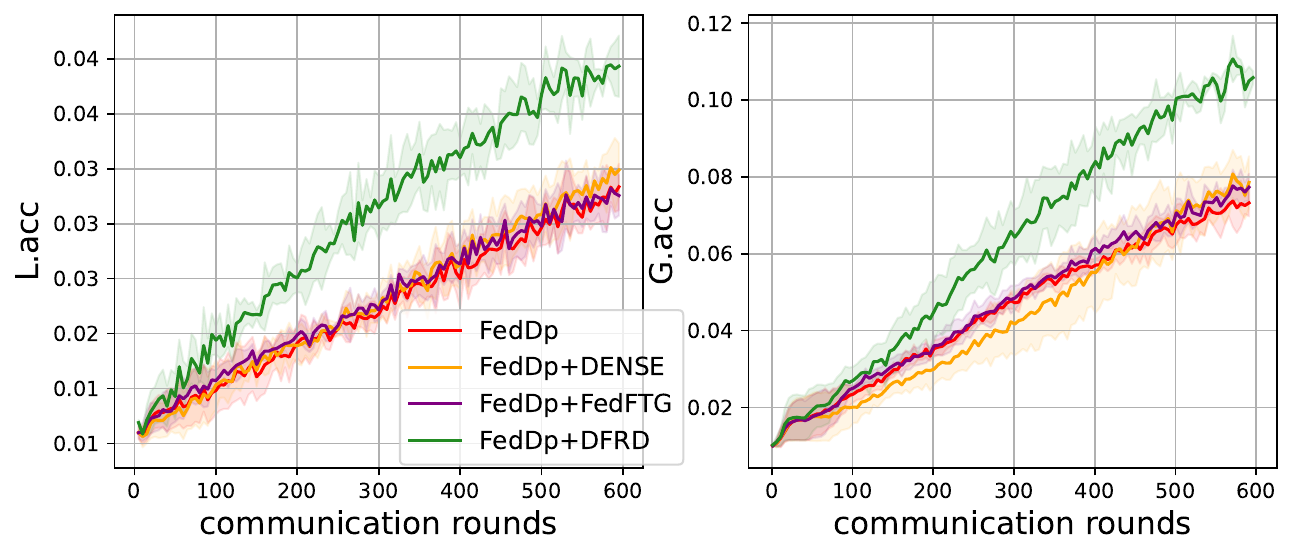}
        \caption{FOOD101, $\rho=5$}
        \label{chutian3}
    \end{subfigure}\\
    \centering
    \begin{subfigure}{0.45\linewidth}
        \centering
        \includegraphics[width=1.0\linewidth]{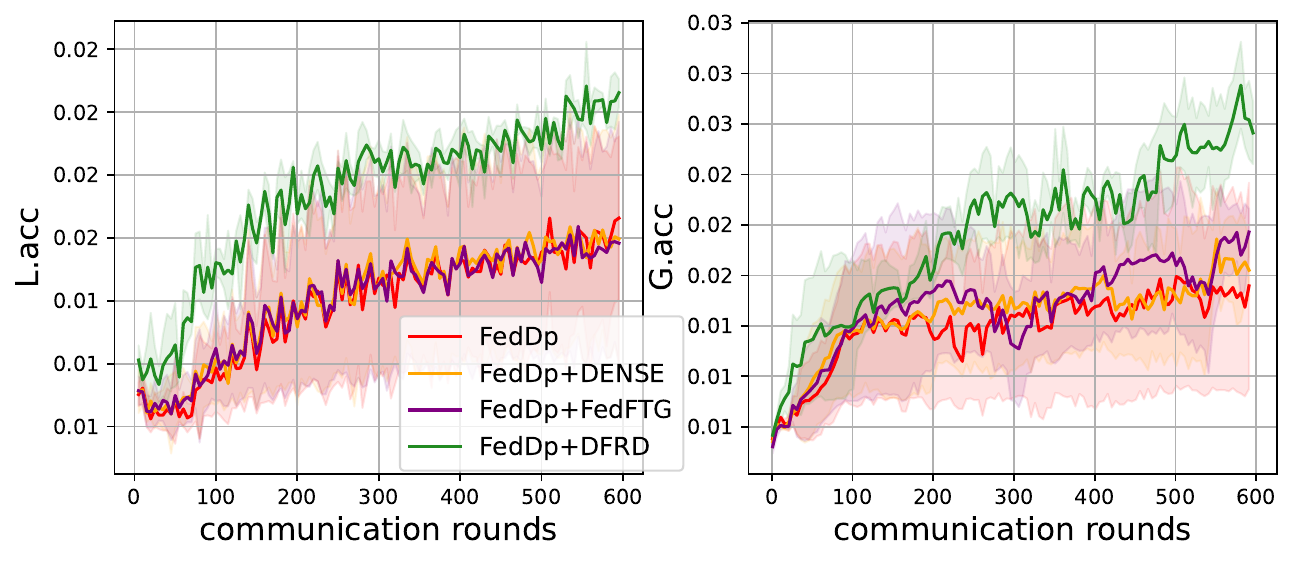}
        \caption{FOOD101, $\rho=10$}
        \label{chutian3}
    \end{subfigure}
    \centering
    \begin{subfigure}{0.45\linewidth}
        \centering
        \includegraphics[width=1.0\linewidth]{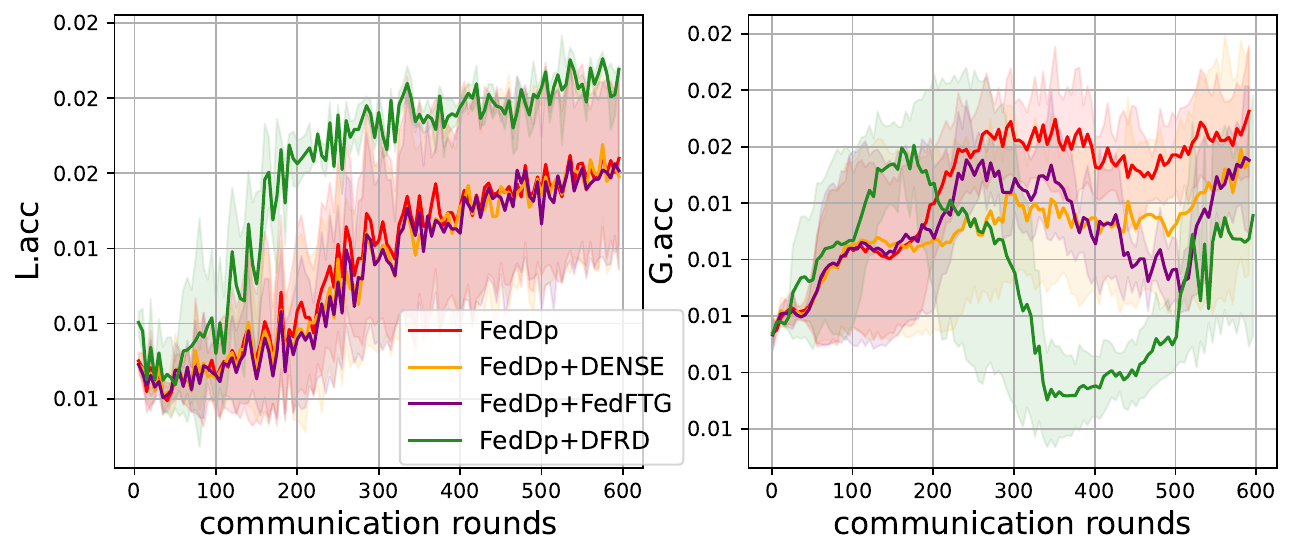}
        \caption{FOOD101, $\rho=40$}
        \label{chutian3}
    \end{subfigure}
    \caption{Learning curves of distinct fine-tuning methods based on FedDp across $\rho \in \{5, 10, 40\}$ on SVHN, CIFAR-10, Tiny-ImageNet and FOOD101.} 
    \label{model_random_heter:}
\end{figure*}

\begin{figure*}[h]\captionsetup[subfigure]{font=scriptsize}
    \centering
    \begin{subfigure}{0.45\linewidth}
        \centering
        \includegraphics[width=1.0\linewidth]{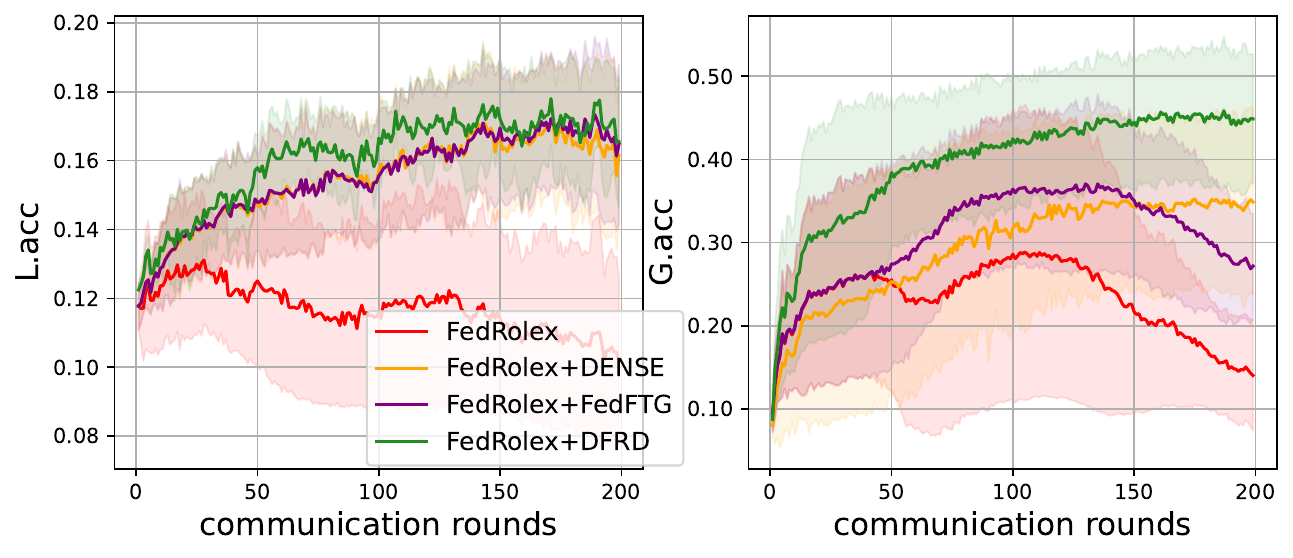}
        \caption{SVHN, $\rho=5$}
        \label{chutian3}
    \end{subfigure}
    \centering
    \begin{subfigure}{0.45\linewidth}
        \centering
        \includegraphics[width=1.0\linewidth]{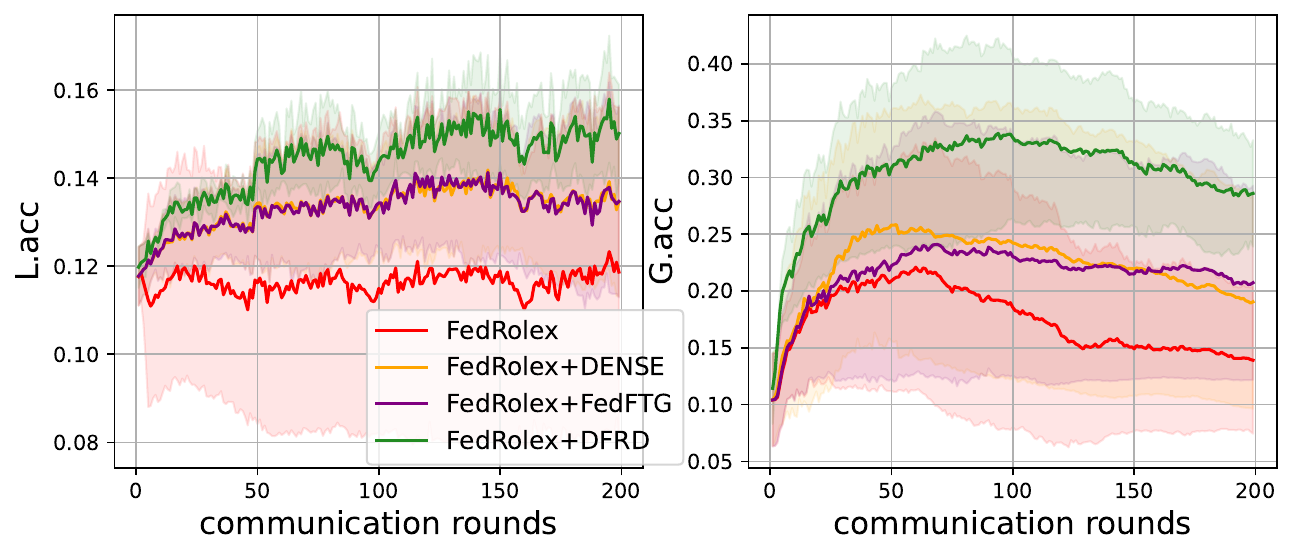}
        \caption{SVHN, $\rho=10$}
        \label{chutian3}
    \end{subfigure} \\
    \centering
    \begin{subfigure}{0.45\linewidth}
        \centering
        \includegraphics[width=1.0\linewidth]{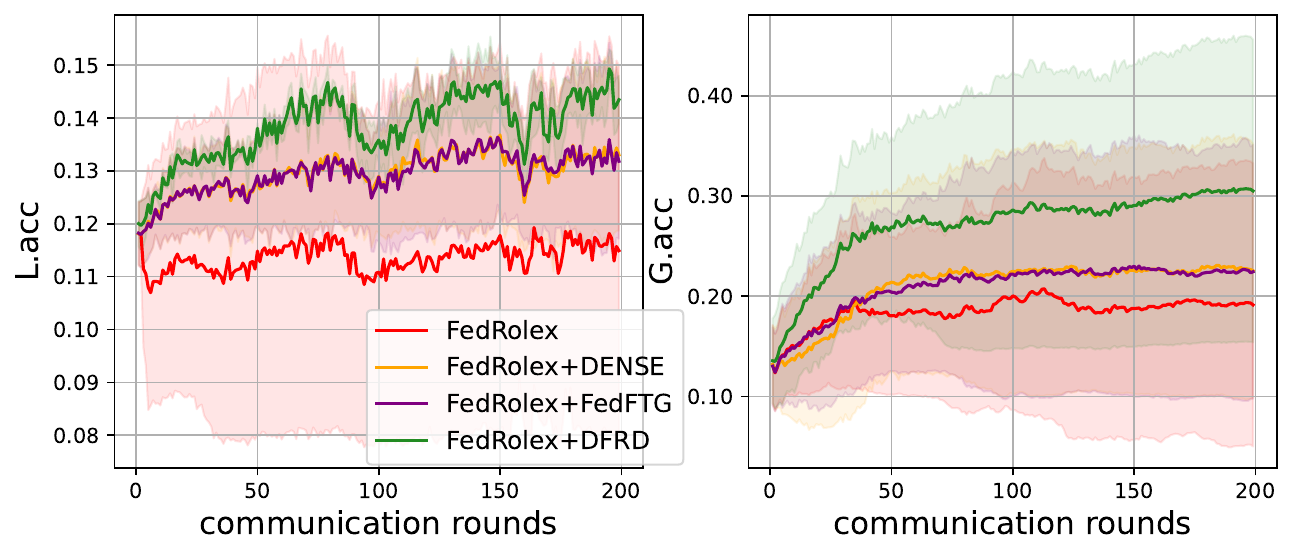}
        \caption{SVHN, $\rho=40$}
        \label{chutian3}
    \end{subfigure}
    \centering
    \begin{subfigure}{0.45\linewidth}
        \centering
        \includegraphics[width=1.0\linewidth]{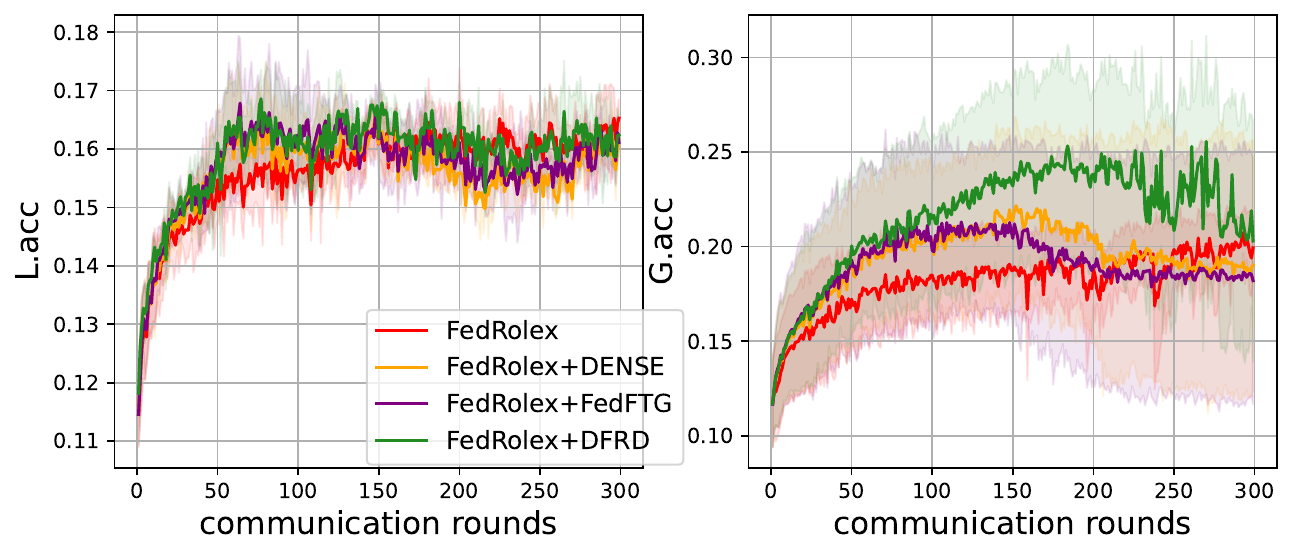}
        \caption{CIFAR-10, $\rho=5$}
        \label{chutian3}
    \end{subfigure} \\
    \centering
    \begin{subfigure}{0.45\linewidth}
        \centering
        \includegraphics[width=1.0\linewidth]{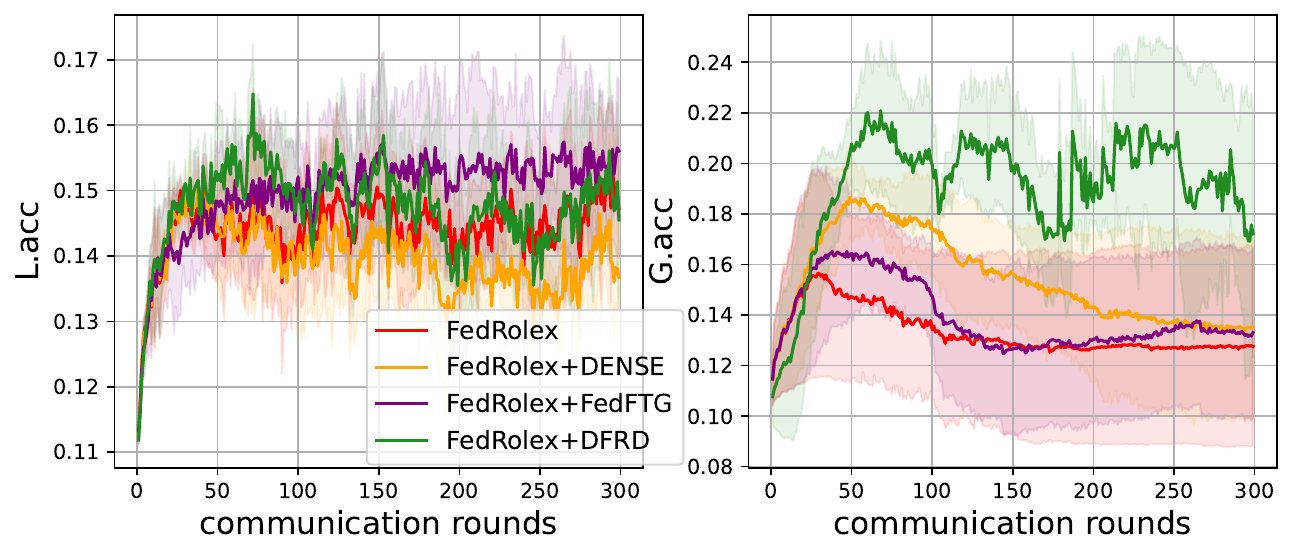}
        \caption{CIFAR-10, $\rho=10$}
        \label{chutian3}
    \end{subfigure}
    \centering
    \begin{subfigure}{0.45\linewidth}
        \centering
        \includegraphics[width=1.0\linewidth]{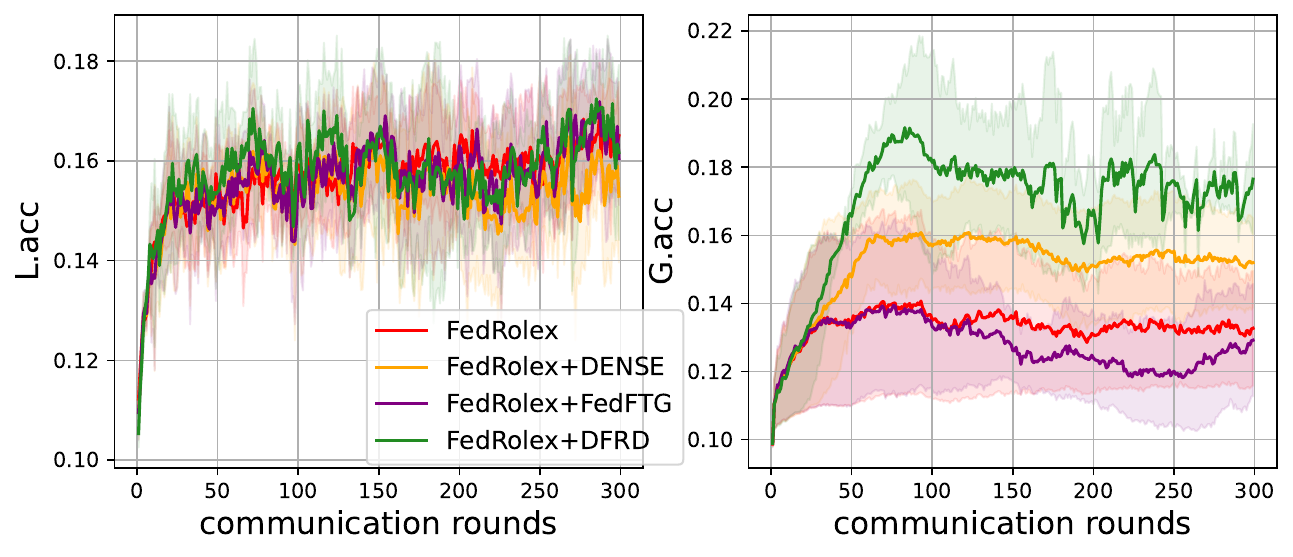}
        \caption{CIFAR-10, $\rho=40$}
        \label{chutian3}
    \end{subfigure}\\
    \centering
    \begin{subfigure}{0.45\linewidth}
        \centering
        \includegraphics[width=1.0\linewidth]{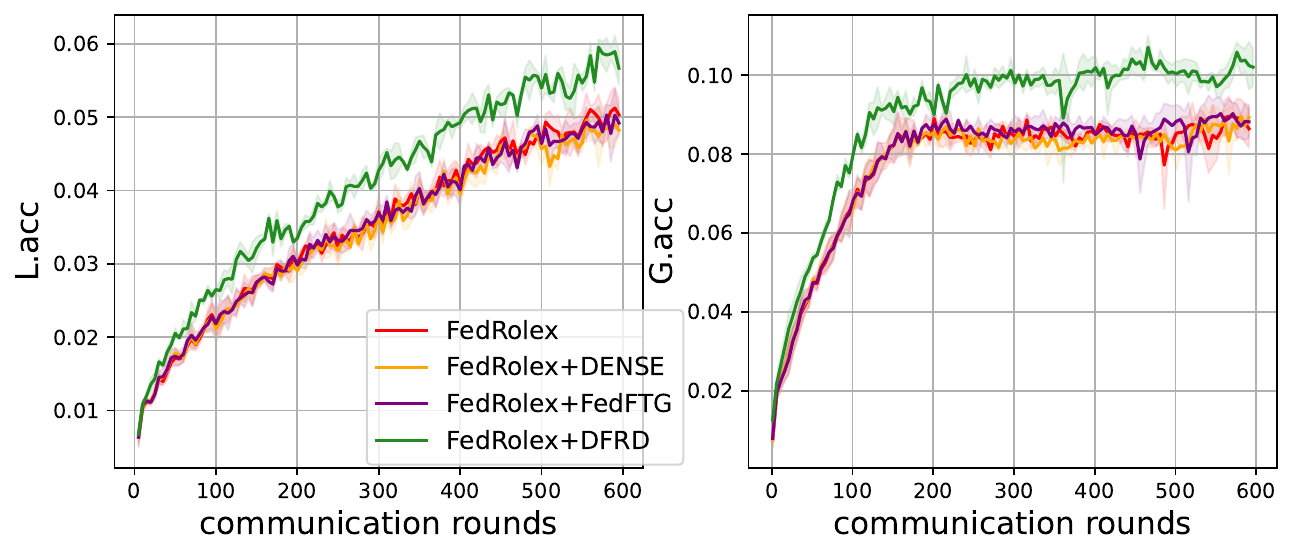}
        \caption{Tiny-ImageNet, $\rho=5$}
        \label{chutian3}
    \end{subfigure}
    \centering
    \begin{subfigure}{0.45\linewidth}
        \centering
        \includegraphics[width=1.0\linewidth]{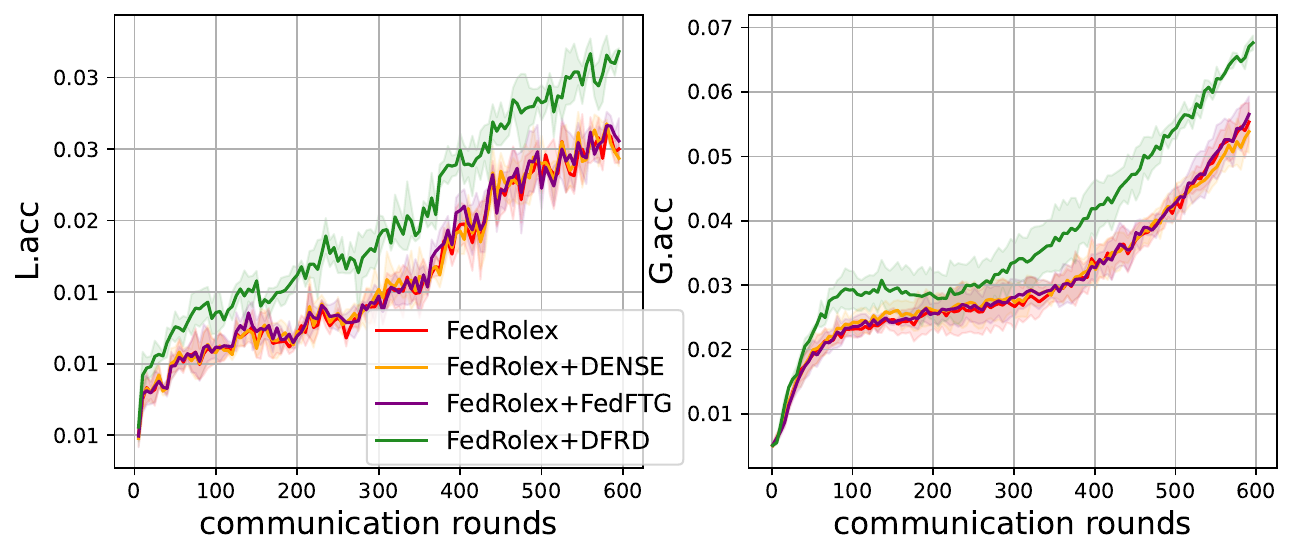}
        \caption{Tiny-ImageNet, $\rho=10$}
        \label{chutian3}
    \end{subfigure}\\
    \centering
    \begin{subfigure}{0.45\linewidth}
        \centering
        \includegraphics[width=1.0\linewidth]{Tiny-Imagenet_roll_level_10_comm_round.pdf}
        \caption{Tiny-ImageNet, $\rho=40$}
        \label{chutian3}
    \end{subfigure}
    \centering
    \begin{subfigure}{0.45\linewidth}
        \centering
        \includegraphics[width=1.0\linewidth]{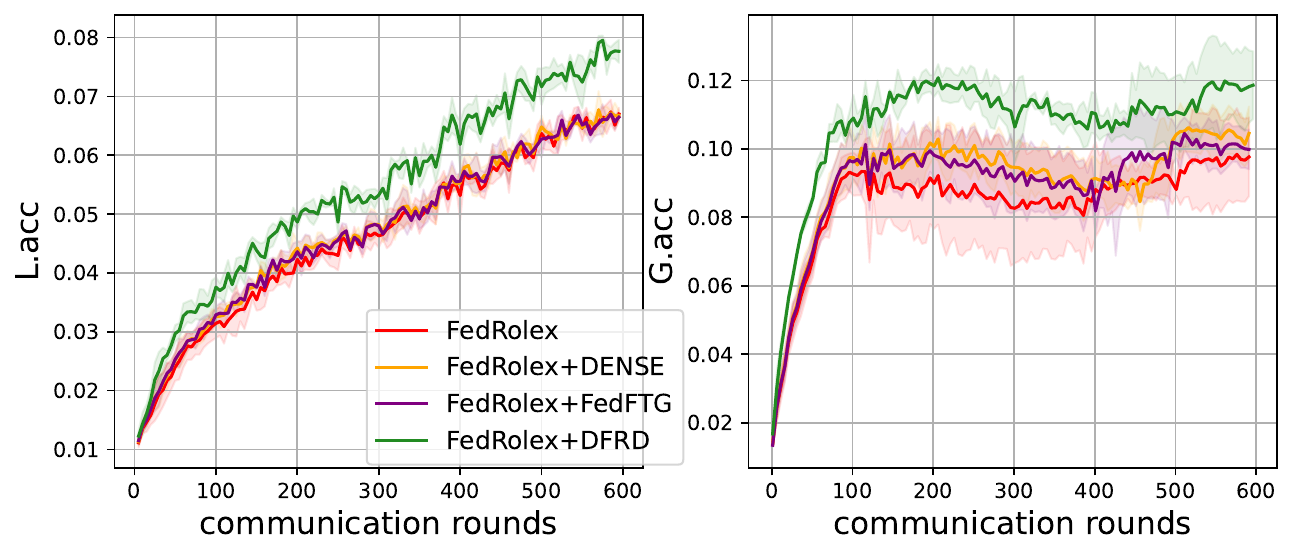}
        \caption{FOOD101, $\rho=5$}
        \label{chutian3}
    \end{subfigure}\\
    \centering
    \begin{subfigure}{0.45\linewidth}
        \centering
        \includegraphics[width=1.0\linewidth]{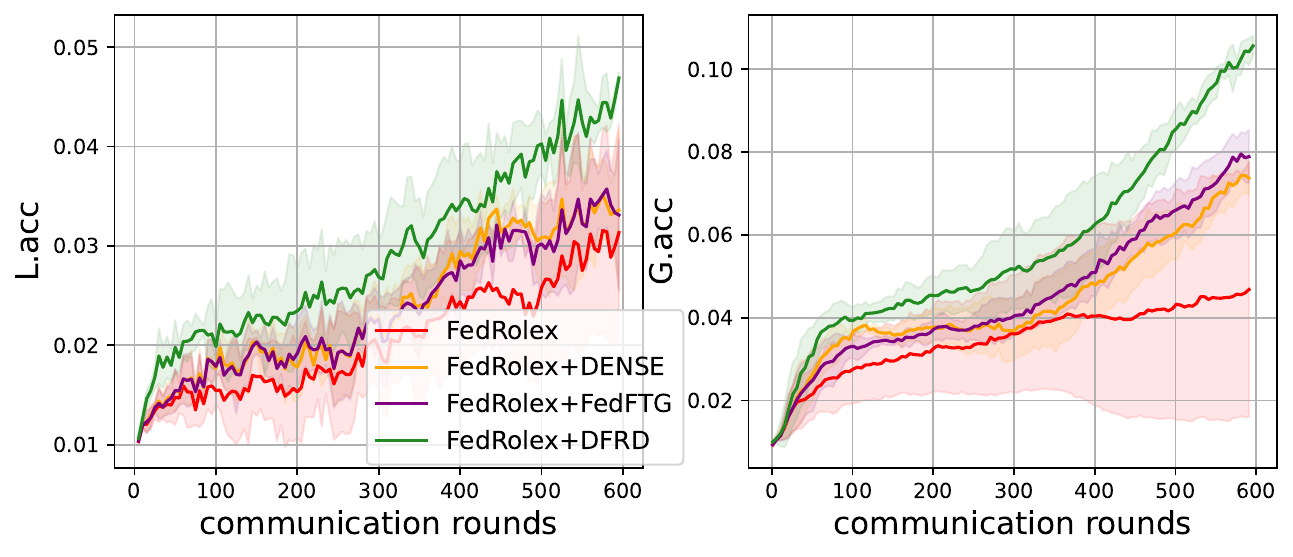}
        \caption{FOOD101, $\rho=10$}
        \label{chutian3}
    \end{subfigure}
    \centering
    \begin{subfigure}{0.45\linewidth}
        \centering
        \includegraphics[width=1.0\linewidth]{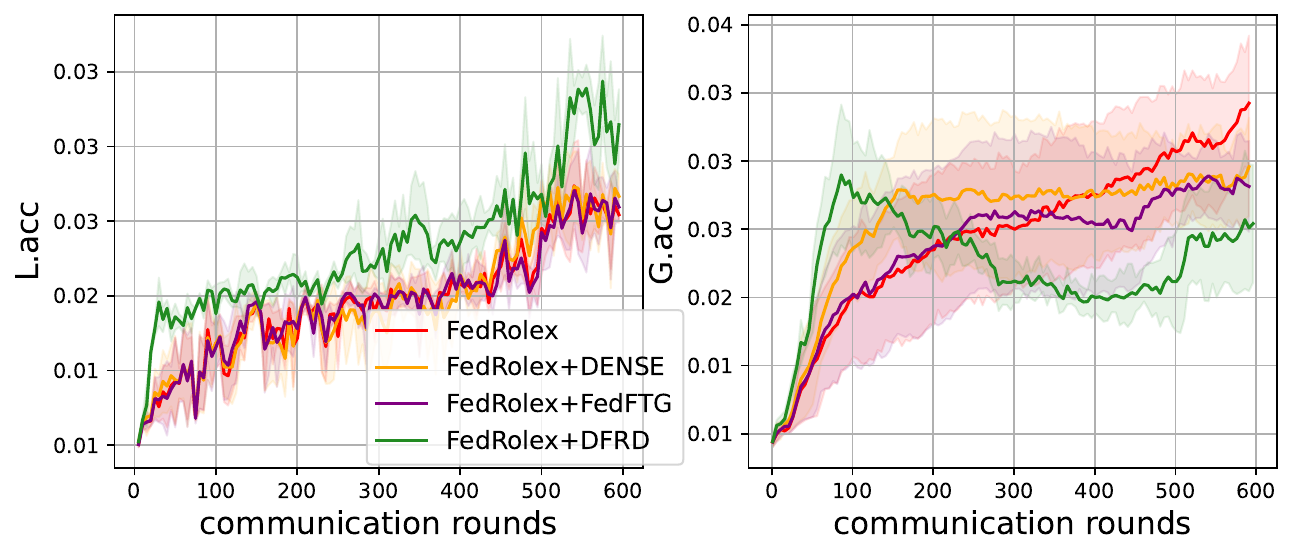}
        \caption{FOOD101, $\rho=40$}
        \label{chutian3}
    \end{subfigure}
    \caption{Learning curves of distinct fine-tuning methods based on FedRolex across $\rho \in \{5, 10, 40\}$ on SVHN, CIFAR-10, Tiny-ImageNet and FOOD101.} 
    \label{model_roll_heter:}
\end{figure*}

In this section, we report the complete learning curves corresponding to Tables~\ref{table_com_omega:} and~\ref{table_com_rho:} in the main paper, see Figures \ref{model_homo_data_free:} to \ref{model_roll_heter:}.

\section{Visualization of synthetic data}
\label{Vis_syn_samp:}
\begin{figure*}[h]\captionsetup[subfigure]{font=scriptsize}
    \centering
    \begin{subfigure}{0.325\linewidth}
        \centering
        \includegraphics[width=1.0\linewidth]{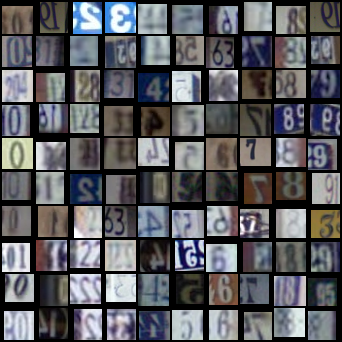}
        \caption{SVHN}
        \label{chutian3}
    \end{subfigure}
    \centering
    \begin{subfigure}{0.325\linewidth}
        \centering
        \includegraphics[width=1.0\linewidth]{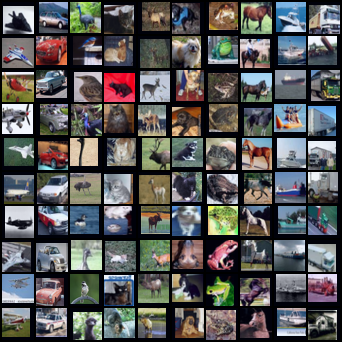}
        \caption{CIFAR-10}
        \label{chutian3}
    \end{subfigure} 
    \centering
    \begin{subfigure}{0.325\linewidth}
        \centering
        \includegraphics[width=1.0\linewidth]{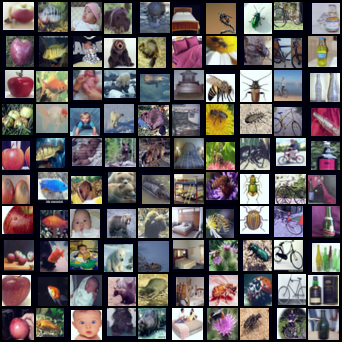}
        \caption{CIFAR-100}
        \label{chutian3}
    \end{subfigure} 
    \caption{Original data selected from SVHN, CIFAR-10 and CIFAR-100. We select $10$ images from each class in each dataset as a column. Note that we select the first 10 classes for display on CIFAR-100.} 
    \label{raw_data:}
\end{figure*}

\begin{figure*}[h]\captionsetup[subfigure]{font=scriptsize}
    \centering
    \begin{subfigure}{0.4875\linewidth}
        \centering
        \includegraphics[width=1.0\linewidth]{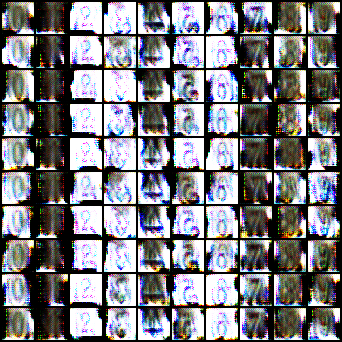}
        \caption{SVHN, generator $G$}
        \label{chutian3}
    \end{subfigure}
    \centering
    \begin{subfigure}{0.4875\linewidth}
        \centering
        \includegraphics[width=1.0\linewidth]{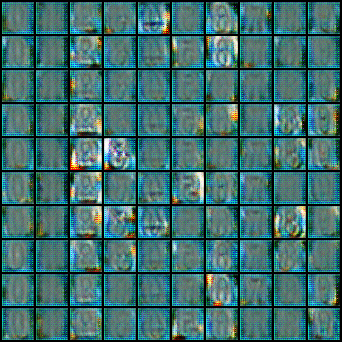}
        \caption{SVHN, generator $\Tilde{G}$}
        \label{chutian3}
    \end{subfigure} \\
    \begin{subfigure}{0.4875\linewidth}
        \centering
        \includegraphics[width=1.0\linewidth]{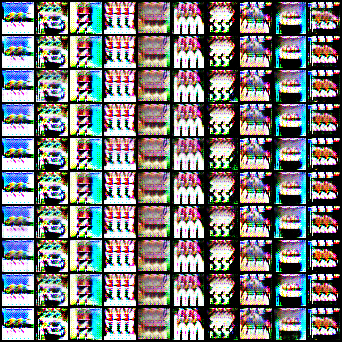}
        \caption{CIFAR-10, generator $G$}
        \label{chutian3}
    \end{subfigure}
    \centering
    \begin{subfigure}{0.4875\linewidth}
        \centering
        \includegraphics[width=1.0\linewidth]{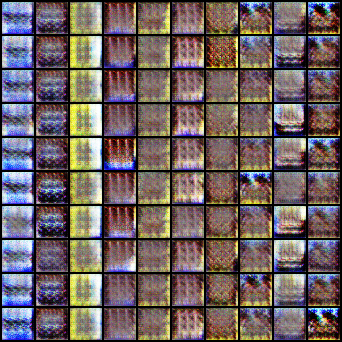}
        \caption{CIFAR-10, generator $\Tilde{G}$}
        \label{chutian3}
    \end{subfigure} \\
    \begin{subfigure}{0.4875\linewidth}
        \centering
        \includegraphics[width=1.0\linewidth]{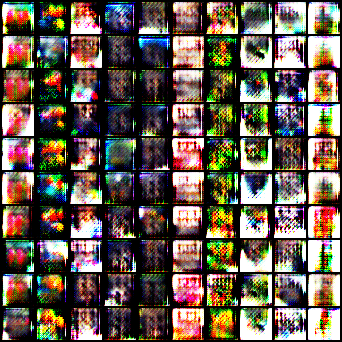}
        \caption{CIFAR-100, generator $G$}
        \label{chutian3}
    \end{subfigure}
    \centering
    \begin{subfigure}{0.4875\linewidth}
        \centering
        \includegraphics[width=1.0\linewidth]{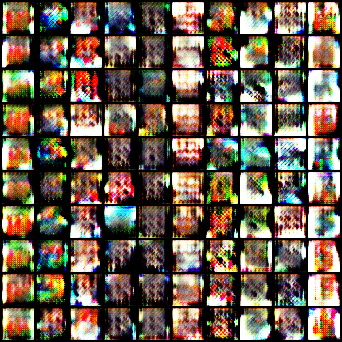}
        \caption{CIFAR-100, generator $\Tilde{G}$}
        \label{chutian3}
    \end{subfigure}
    \caption{Visualization of synthetic data on SVHN, CIFAR-10 and CIFAR-100 by using DFRD with diversity constraint based on \textit{mul}.} 
    \label{data_gen_mul:}
\end{figure*}

\begin{figure*}[h]\captionsetup[subfigure]{font=scriptsize}
    \centering
    \begin{subfigure}{0.4875\linewidth}
        \centering
        \includegraphics[width=1.0\linewidth]{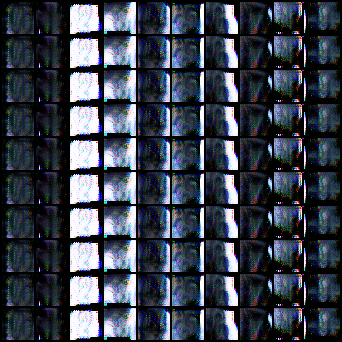}
        \caption{SVHN, generator $G$}
        \label{chutian3}
    \end{subfigure}
    \centering
    \begin{subfigure}{0.4875\linewidth}
        \centering
        \includegraphics[width=1.0\linewidth]{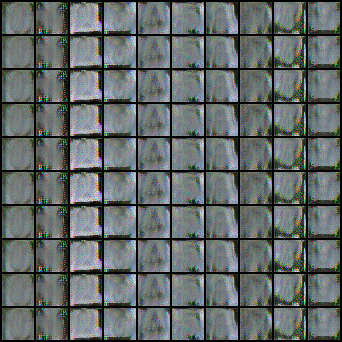}
        \caption{SVHN, generator $\Tilde{G}$}
        \label{chutian3}
    \end{subfigure} \\
    \begin{subfigure}{0.4875\linewidth}
        \centering
        \includegraphics[width=1.0\linewidth]{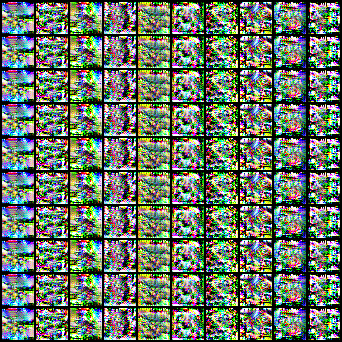}
        \caption{CIFAR-10, generator $G$}
        \label{chutian3}
    \end{subfigure}
    \centering
    \begin{subfigure}{0.4875\linewidth}
        \centering
        \includegraphics[width=1.0\linewidth]{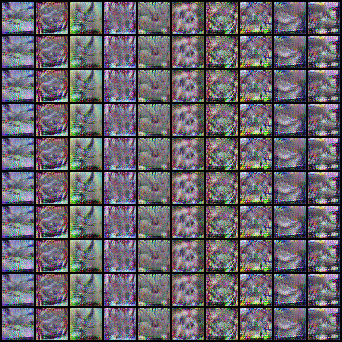}
        \caption{CIFAR-10, generator $\Tilde{G}$}
        \label{chutian3}
    \end{subfigure} \\
    \begin{subfigure}{0.4875\linewidth}
        \centering
        \includegraphics[width=1.0\linewidth]{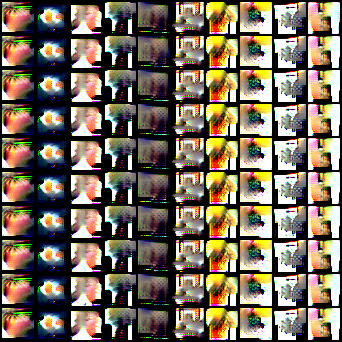}
        \caption{CIFAR-100, generator $G$}
        \label{chutian3}
    \end{subfigure}
    \centering
    \begin{subfigure}{0.4875\linewidth}
        \centering
        \includegraphics[width=1.0\linewidth]{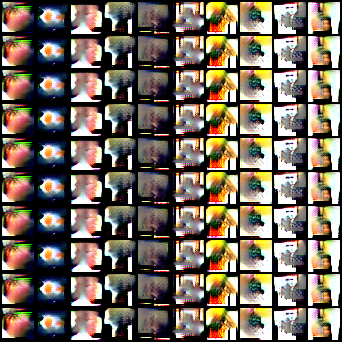}
        \caption{CIFAR-100, generator $\Tilde{G}$}
        \label{chutian3}
    \end{subfigure}
    \caption{Visualization of synthetic data on SVHN, CIFAR-10 and CIFAR-100 by using DFRD with diversity constraint based on \textit{add}.} 
    \label{data_gen_add:}
\end{figure*}

\begin{figure*}[h]\captionsetup[subfigure]{font=scriptsize}
    \centering
    \begin{subfigure}{0.4875\linewidth}
        \centering
        \includegraphics[width=1.0\linewidth]{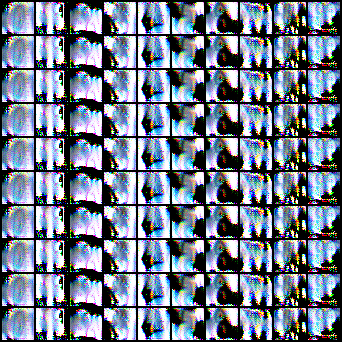}
        \caption{SVHN, generator $G$}
        \label{chutian3}
    \end{subfigure}
    \centering
    \begin{subfigure}{0.4875\linewidth}
        \centering
        \includegraphics[width=1.0\linewidth]{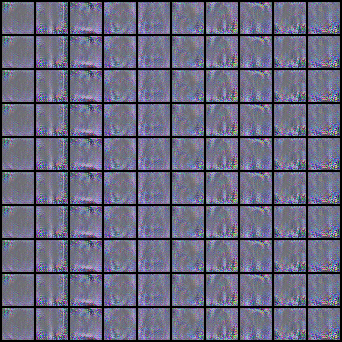}
        \caption{SVHN, generator $\Tilde{G}$}
        \label{chutian3}
    \end{subfigure} \\
    \begin{subfigure}{0.4875\linewidth}
        \centering
        \includegraphics[width=1.0\linewidth]{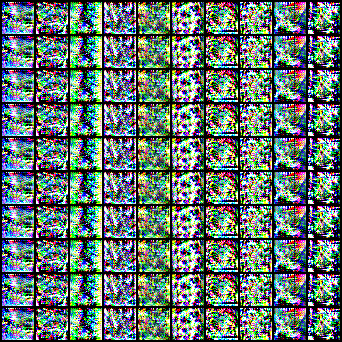}
        \caption{CIFAR-10, generator $G$}
        \label{chutian3}
    \end{subfigure}
    \centering
    \begin{subfigure}{0.4875\linewidth}
        \centering
        \includegraphics[width=1.0\linewidth]{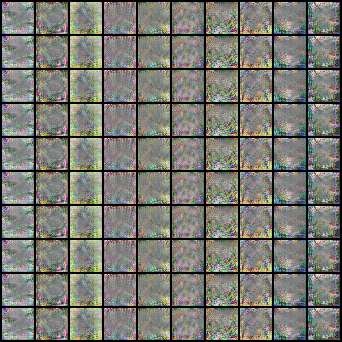}
        \caption{CIFAR-10, generator $\Tilde{G}$}
        \label{chutian3}
    \end{subfigure} \\
    \begin{subfigure}{0.4875\linewidth}
        \centering
        \includegraphics[width=1.0\linewidth]{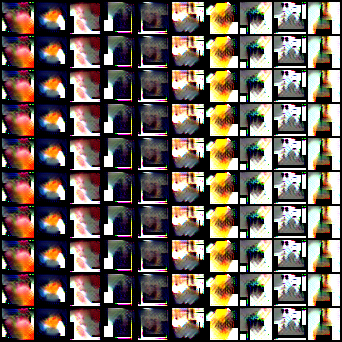}
        \caption{CIFAR-100, generator $G$}
        \label{chutian3}
    \end{subfigure}
    \centering
    \begin{subfigure}{0.4875\linewidth}
        \centering
        \includegraphics[width=1.0\linewidth]{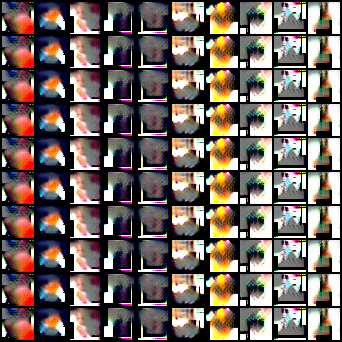}
        \caption{CIFAR-100, generator $\Tilde{G}$}
        \label{chutian3}
    \end{subfigure}
    \caption{Visualization of synthetic data on SVHN, CIFAR-10 and CIFAR-100 by using DFRD with diversity constraint based on \textit{cat}.} 
    \label{data_gen_cat:}
\end{figure*}

\begin{figure*}[h]\captionsetup[subfigure]{font=scriptsize}
    \centering
    \begin{subfigure}{0.4875\linewidth}
        \centering
        \includegraphics[width=1.0\linewidth]{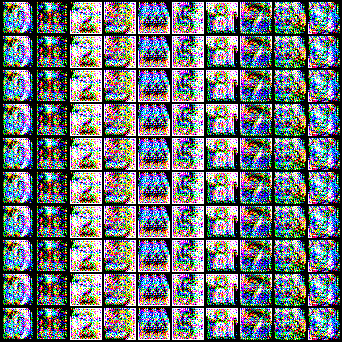}
        \caption{SVHN, generator $G$}
        \label{chutian3}
    \end{subfigure}
    \centering
    \begin{subfigure}{0.4875\linewidth}
        \centering
        \includegraphics[width=1.0\linewidth]{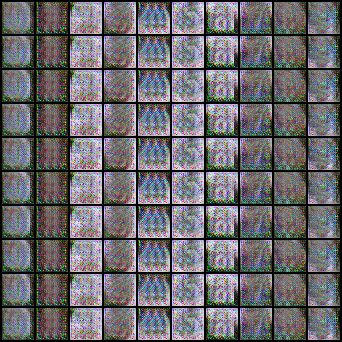}
        \caption{SVHN, generator $\Tilde{G}$}
        \label{chutian3}
    \end{subfigure} \\
    \begin{subfigure}{0.4875\linewidth}
        \centering
        \includegraphics[width=1.0\linewidth]{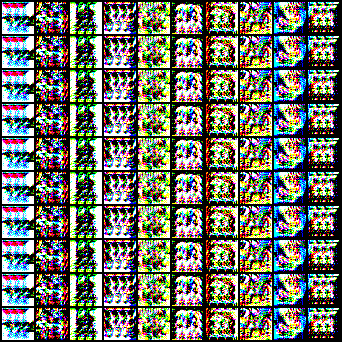}
        \caption{CIFAR-10, generator $G$}
        \label{chutian3}
    \end{subfigure}
    \centering
    \begin{subfigure}{0.4875\linewidth}
        \centering
        \includegraphics[width=1.0\linewidth]{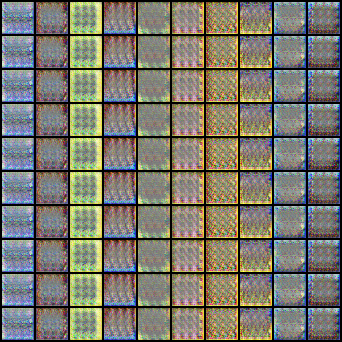}
        \caption{CIFAR-10, generator $\Tilde{G}$}
        \label{chutian3}
    \end{subfigure} \\
    \begin{subfigure}{0.4875\linewidth}
        \centering
        \includegraphics[width=1.0\linewidth]{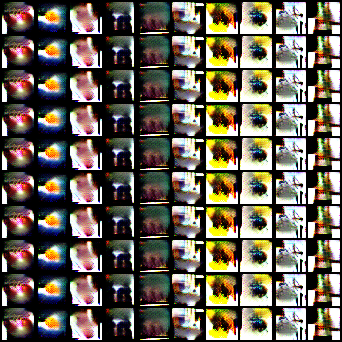}
        \caption{CIFAR-100, generator $G$}
        \label{chutian3}
    \end{subfigure}
    \centering
    \begin{subfigure}{0.4875\linewidth}
        \centering
        \includegraphics[width=1.0\linewidth]{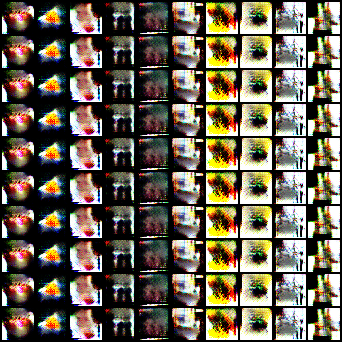}
        \caption{CIFAR-100, generator $\Tilde{G}$}
        \label{chutian3}
    \end{subfigure}
    \caption{Visualization of synthetic data on SVHN, CIFAR-10 and CIFAR-100 by using DFRD with diversity constraint based on \textit{n-cat}.} 
    \label{data_gen_n_cat:}
\end{figure*}

\begin{figure*}[h]\captionsetup[subfigure]{font=scriptsize}
    \centering
    \begin{subfigure}{0.4875\linewidth}
        \centering
        \includegraphics[width=1.0\linewidth]{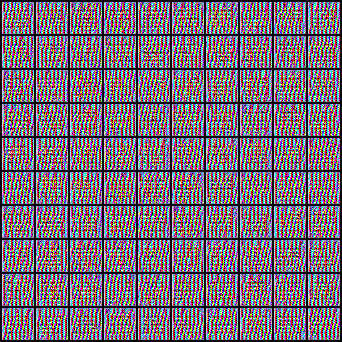}
        \caption{SVHN, generator $G$}
        \label{chutian3}
    \end{subfigure}
    \centering
    \begin{subfigure}{0.4875\linewidth}
        \centering
        \includegraphics[width=1.0\linewidth]{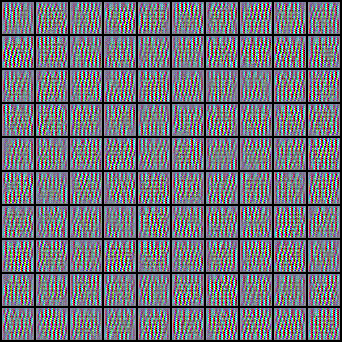}
        \caption{SVHN, generator $\Tilde{G}$}
        \label{chutian3}
    \end{subfigure} \\
    \begin{subfigure}{0.4875\linewidth}
        \centering
        \includegraphics[width=1.0\linewidth]{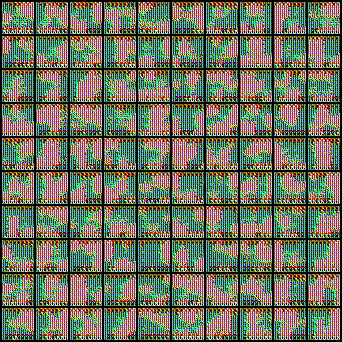}
        \caption{CIFAR-10, generator $G$}
        \label{chutian3}
    \end{subfigure}
    \centering
    \begin{subfigure}{0.4875\linewidth}
        \centering
        \includegraphics[width=1.0\linewidth]{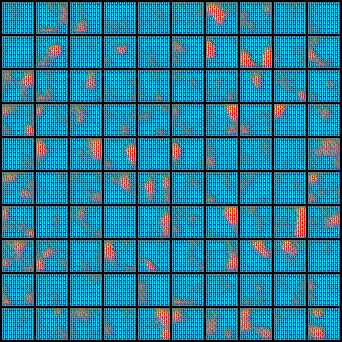}
        \caption{CIFAR-10, generator $\Tilde{G}$}
        \label{chutian3}
    \end{subfigure} \\
    \begin{subfigure}{0.4875\linewidth}
        \centering
        \includegraphics[width=1.0\linewidth]{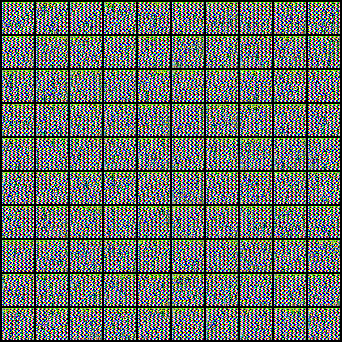}
        \caption{CIFAR-100, generator $G$}
        \label{chutian3}
    \end{subfigure}
    \centering
    \begin{subfigure}{0.4875\linewidth}
        \centering
        \includegraphics[width=1.0\linewidth]{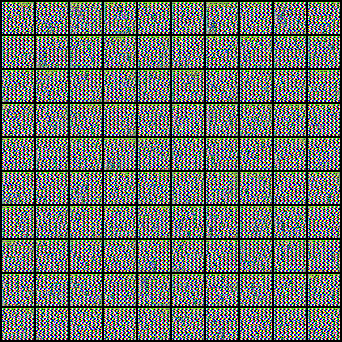}
        \caption{CIFAR-100, generator $\Tilde{G}$}
        \label{chutian3}
    \end{subfigure}
    \caption{Visualization of synthetic data on SVHN, CIFAR-10 and CIFAR-100 by using DFRD with diversity constraint based on \textit{none}.} 
    \label{data_gen_none:}
\end{figure*}

In this section, we visualize the synthetic data generated by $G(\cdot)$ and $\Tilde{G}(\cdot)$ on SVHN, CIFAR-10 and CIFAR-100 based on different merge operators, as shown in Figures~\ref{data_gen_mul:} to~\ref{data_gen_none:}.
Additionally, we visualize the partial original data of SVHN, CIFAR-10 and CIFAR-100, see Fig.~\ref{raw_data:} for details.
From Figures~\ref{raw_data:} to~\ref{data_gen_none:}, it can be observed that all the synthetic data are visually very dissimilar to the original data, which effectively avoids the leakage of the original data.
As we state in Section~\ref{Train_Gen:} of the main paper,  a well-trained generator should satisfy several key characteristics: \textit{fidelity}, \textit{transferability}, and \textit{diversity}.
Therefore, we analyze the performance of $G(\cdot)$ and $\Tilde{G}(\cdot)$ with different merge operators in terms of these three aspects.
Firstly, we can easily see from Figures~\ref{data_gen_mul:} to~\ref{data_gen_none:} that $G(\cdot)$ and $\Tilde{G}(\cdot)$ based on $mul$, $add$, $cat$ and $ncat$ can capture the shared information of the original data while ensuring data privacy.
However, synthetic data generated by $G(\cdot)$ and $\Tilde{G}(\cdot)$ based on $none$~(i.e. not conditional generator) is not visually discernible as to whether it extracts valid information from the original data.
From the test accuracies reported in Table~\ref{table_noise_label:} in the main paper, we can see that DFRD based on $none$ consistently underperforms other competitors in terms of global test accuracy, which indicates that the generator and EMA generator based on $none$ generate lower quality synthetic data.
Also, $mul$ consistently outperforms other merge operators w.r.t. global test accuracy, suggesting that the generator and EMA generator based on $mul$ can generate more effective synthetic data.
Furthermore, in terms of diversity, the synthetic data generated by $\Tilde{G}(\cdot)$ based on $mul$ is visually diverse both inter and intra classes, which suggests that no model collapse occurred during training of the EMA generator.
It is worth noting that the synthetic data generated by $\Tilde{G}(\cdot)$ is different from that generated by $G(\cdot)$, meaning that the EMA generator can be used as a complement to the generator in DFRD.

\section{Limitations}
\label{app_discussion:}
In the field of Federated Learning~(FL), there are many trade-offs, including utility, privacy protection, computational efficiency and communication cost, etc.
It is well known that trying to develop a universal FL method that can address all problems is extremely challenging. 
In this work, we mainly focus on improving the utility of the global model in FL.
Next, we discuss some of the limitations of DFRD.

\textbf{Privacy Protection.} 
We acknowledge that it is difficult for DFRD to strictly guarantee privacy without privacy protection.
Since DFRD generates synthetic data in the server that is similar to clients' training data, it may violate the privacy regulations in FL.
However, according to our observation, DFRD can only capture the shared attributes of the dataset without access to real data, and it is difficult to concretely and visually reveal the unique attributes of individual data~(see Figures~\ref{data_gen_mul:} to~\ref{data_gen_n_cat:}).
In addition, DFRD requires clients to upload the label statistics of the data, which also is at risk of compromising privacy.
However, based on our proposed dynamic weighting and label sampling, each client only needs to count the labels of the data involved in local training instead of counting that of the entire local training dataset, which mitigates the privacy leakage to some extent.
Although our work does not address the privacy issue, in tasks with high privacy protection requirements, DFRD can incorporate some privacy protection techniques to enhance the reliability of FL systems in practice, such as differential privacy~(DP)~\cite{Dwork2008In, Geyer2017Differentially, Cheng2022Differentially}.
Note that the combination of DFRD and encryption technologies~\cite{Ma2022Privacy, Zhang2022Homomorphic} may not be feasible, since {\it training generator} and {\it robust model distillation} in DFRD require knowledge of the structure of the local models.
In addition, we argue that the \textbf{theoretical guarantee} for privacy protection is crucial. 
However, It's worth noting that even in exist well-known efforts~\cite{Zhang2022Fine, Zhang2022DENSE, Do2022Momentum, Yoo2019Knowledge}, including those on handling data and model heterogeneity for FL with the help DFKD~\cite{Zhang2022Fine, Zhang2022DENSE}, comprehensive theoretical analysis concerning the boundaries of data generation is often absent. 
Given the lack of suitable theoretical frameworks, we concentrated on robust empirical validation, showcasing our method~(DFRD). 
Our results, we believe, robustly demonstrate our method's utility. 
We intend to delve deeper into theoretical aspects in future work.

\textbf{Computational Efficiency and Communication Cost.}
To achieve a robust global model, DFRD requires additional computation.
Specifically, compared with FedAvg or PT-based methods, the training time of DFRD as a fine-tuning method will be longer, as it needs to additionally train the generator and the global model on the server.
In our experiments, DFRD takes two to three times longer to run per communication round than they do.
Moreover, compared to FedAvg or PT-based methods, DFRD requires an additional vector of label statistics to be transmitted~(see line 11 in Algorithm~\ref{alg_all:}).
However, the communication cost of this vector is negligible compared to that of the model.

\section{Broader Impacts}
\label{Broader_Impacts:}
We work on how to train a robust global model with the help of data-free knowledge distillation~(DFKD) in the Federated Learning~(FL) scenario where data and models are simultaneously heterogeneous.
Our work points out that existing methods combining FL and DFKD do not thoroughly study the training of the generator, 
and neglect the catastrophic forgetting of the global model caused by large distribution shifts of the generator in the scenarios of coexisting data and model heterogeneity. 
Our proposed DFRD can deal with the said issues well.
DFRD exemplifies potential positive impacts on society, enabling a global model with superior performance while ensuring basic privacy protection in real-world FL applications.
Meanwhile, DFRD may have negative social impacts related to sensitive information and high resource consumption.
While DFRD does not compromise the private information of individual data, it may expose shared information with utility for the performance of the generator.
In addition, DFRD-based FL systems require more server-side power resources to train the global model.
DFRD does not involve social ethics.


\end{document}